%% file: main.tex
\documentclass[runningheads]{llncs}

\usepackage{accv}

\usepackage{accvabbrv}

\usepackage{graphicx}
\usepackage{booktabs}

\usepackage[accsupp]{axessibility}  

\usepackage{hyperref}
\hypersetup{
colorlinks=true,
linkcolor=black
}
\usepackage{orcidlink}
\usepackage{colortbl}
\usepackage{multirow}
\usepackage{array}
\begin{document}

\title{SpikeGS: Learning 3D Gaussian Fields from Continuous Spike Stream} 

\titlerunning{SpikeGS}
\author{Jinze Yu\inst{1} \and
Xin Peng\inst{2} \and
Zhengda Lu\inst{3}\orcidlink{0000-0002-9581-2268} \and  Laurent Kneip\inst{4}\orcidlink{0000-0001-6727-6608} \and Yiqun Wang\inst{1}\thanks{Corresponding author}\orcidlink{0000-0003-1942-5597}}

\authorrunning{Yu et al.}

\institute{College of Computer Science, Chongqing University, Chongqing, China \and
Motovis Co., Ltd., Shanghai, China
\and University of Chinese Academy of Sciences, Beijing, China 
\and
Mobile Perception Lab, ShanghaiTech University, Shanghai, China
}

\maketitle

\begin{abstract}
   A spike camera is a specialized high-speed visual sensor that offers advantages such as high temporal resolution and high dynamic range compared to conventional frame cameras. 
   These features provide the camera with significant advantages in many computer vision tasks.
    However, the tasks of novel view synthesis based on spike cameras remain underdeveloped. Although there are existing methods for learning neural radiance fields from spike stream, they either lack robustness in extremely noisy, low-quality lighting conditions or suffer from high computational complexity due to the deep fully connected neural networks and ray marching rendering strategies used in neural radiance fields, making it difficult to recover fine texture details. In contrast, the latest advancements in 3DGS have achieved high-quality real-time rendering by optimizing the point cloud representation into Gaussian ellipsoids. Building on this, we introduce SpikeGS, the method to learn 3D Gaussian fields solely from spike stream. We designed a differentiable spike stream rendering framework based on 3DGS, incorporating noise embedding and spiking neurons. By leveraging the multi-view consistency of 3DGS and the tile-based multi-threaded parallel rendering mechanism, we achieved high-quality real-time rendering results. Additionally, we introduced a spike rendering loss function that generalizes under varying illumination conditions. Our method can reconstruct view synthesis results with fine texture details from a continuous spike stream captured by a moving spike camera, while demonstrating high robustness in extremely noisy low-light scenarios. Experimental results on both real and synthetic datasets demonstrate that our method surpasses existing approaches in terms of rendering quality and speed.

  \keywords{Spike camera \and 3D Gaussian splatting \and Novel View Synthesis \and 3D reconstruction}
\end{abstract}

\section{Introduction}
\label{sec:intro}

In recent years, neuromorphic cameras have made significant advancements, most notably in the development of spike cameras \cite{huang20231000, dong2019efficient} and event cameras \cite{gallego2020event, rebecq2018esim}. These types of cameras excel in capturing intensity changes in high-speed scenes due to their high temporal resolution, and high dynamic range, offering significant potential for applications in the field of computer vision 
\cite{han2020neuromorphic, massa2020efficient, duwek2021image, zhao2023learning, hidalgo2022event, chen2024spikereveal, lin2023fast}. 
The output of a spike camera is a spike stream, fundamentally different from the data produced by traditional frame cameras. When the accumulated photons exceed the threshold, each pixel independently responds to the accumulation of photons by generating asynchronous spikes and then resets the accumulation. At each timestamp, the spike camera outputs a binary matrix, known as an event frame, indicating the presence of spikes at all pixels. This distinctive feature enables the spike camera to have complete texture sensing and high-speed sensing capabilities, and to record full visual details with an ultra-high temporal resolution of up to 40 kHz. Thanks to these capabilities, spike cameras excel in various fields of computer vision tasks \cite{han2020neuromorphic, hu2022optical, wang2022learning, dai2024spikenvs, zhao2021reconstructing}.

Meanwhile, the field of computer vision is increasingly inclined to explore neural radiance fields \cite{mildenhall2021nerf} and 3DGS\cite{kerbl20233d} as solutions for 3D scene reconstruction and novel view synthesis. However, due to the distinctive data modality of spike camera,  current algorithms for 3D reconstruction and novel view synthesis \cite{peng2021neural, niedermayr2024compressed, tretschk2021non, mildenhall2022nerf, srinivasan2021nerv, liu2024fast} primarily rely on high-quality images obtained by traditional frame-based cameras under optimal lighting conditions. This raises the question of whether we can reconstruct dense and realistic 3D scene representations solely from the spike stream captured by a moving spike camera, and whether such reconstructions can maintain robustness in extremely noisy and low-light scenarios (where, in real-world situations, the spike camera inevitably generates a large amount of noise due to its internal circuit structure and external ambient light).  

One of the most representative algorithms in the field of novel view synthesis today is NeRF \cite{mildenhall2021nerf}, which implicitly represents a scene as a neural radiance field. By combining implicit function representation of MLPs with differentiable rendering, NeRF has garnered widespread attention for its ability to recover high-quality 3D scene representations from 2D images. However, due to the use of deep fully connected neural networks and the need for per-pixel ray sampling during the rendering process, NeRF suffers from high sampling costs, potential noise generation, and considerable computational complexity. A recent advancement, 3D Gaussian Splatting (3DGS)~\cite{kerbl20233d}, explicitly represents scenes by optimizing Gaussian ellipsoids. Thanks to the tile-based multithreaded parallel rendering mechanism of 3D Gaussian, which achieves real-time rendering by splatting three-dimensional ellipsoids onto a two-dimensional plane, surpassing NeRF in both rendering quality and speed.

Although some studies have explored the application of 3DGS and NeRF in a unique type of neuromorphic camera known as an event camera, characterized by differential sampling. Representative examples include Ev-NeRF \cite{hwang2023ev}, EventNeRF \cite{rudnev2023eventnerf}, and EvGGS \cite{wang2024evggs}, all of which introduce neural radiance fields or Gaussian fields derived specifically from event streams. However, the inherent lack of texture detail in event data limits the effectiveness of these methods, resulting in suboptimal outcomes. The concurrent work \cite{guo2024spikegs} \cite{zhang2024spikegs} explored spike stream reconstruction under high-speed motion cameras. SpikeNeRF \cite{zhu2024spikenerf} and Spike-NeRF \cite{guo2024spike} have both explored methods for learning neural radiance fields from spike stream. However, the deep fully connected neural networks and ray marching-based rendering strategies used in neural radiance fields make it difficult to achieve high-quality real-time rendering. Additionally, Spike-NeRF \cite{guo2024spike} does not demonstrate robustness in extremely noisy, low-light scenarios.
 
\begin{figure}[h] 
\centering
\includegraphics[width=\linewidth]{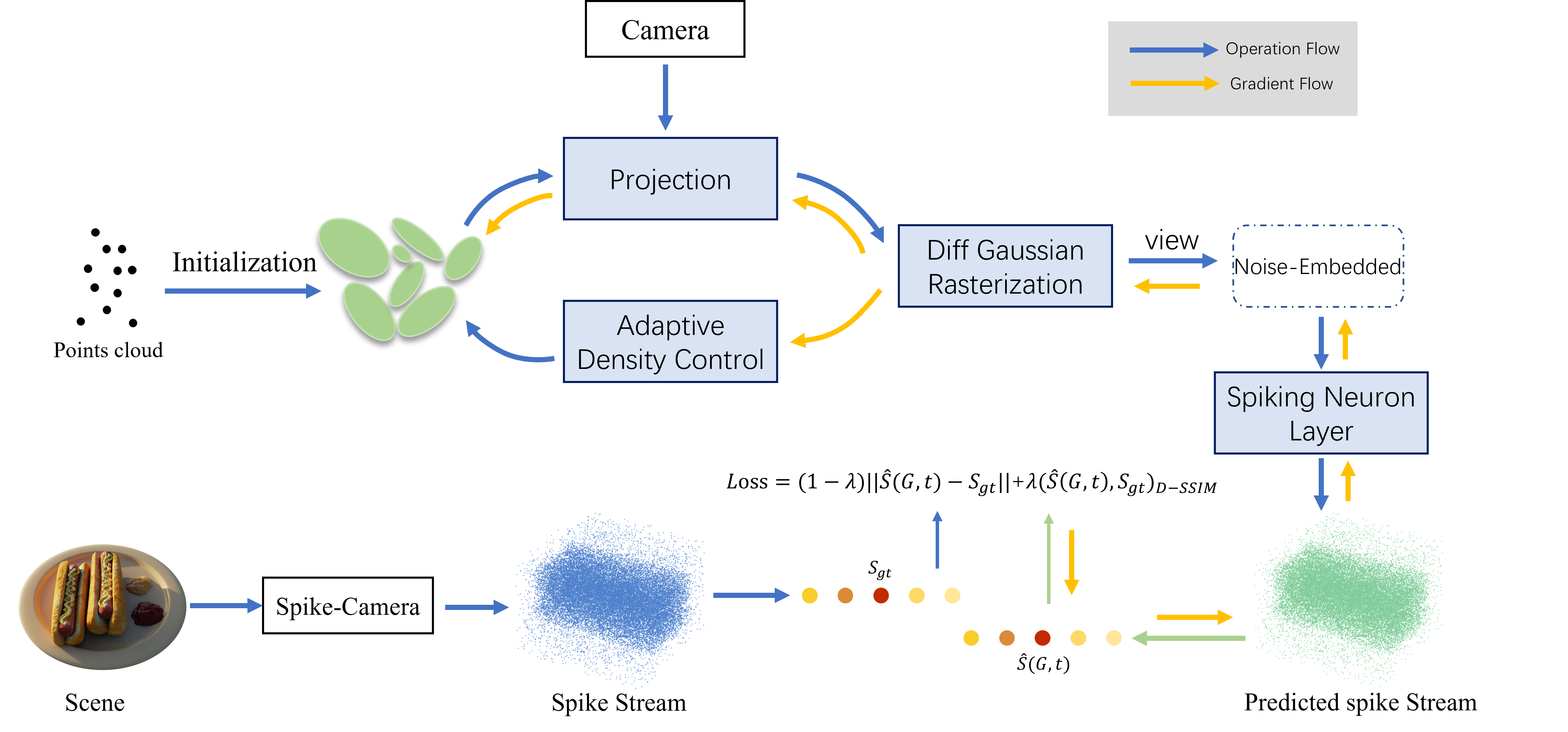}
\caption{The architecture of SpikeGS. we establish the connection between the 3D Gaussian fields and the real-world spike stream S. Firstly, the input is randomly initialized point clouds or sparse point clouds generated by structure-from-motion (SfM) \cite{schonberger2016structure}. The point clouds are then converted into Gaussian ellipsoids, and rasterization is used to render the pixel values from corresponding viewpoints. Next, the rendered pixels are converted into spike stream through a spike stream generation pipeline (Noise Embedded and Spiking Neurons), and a rendering loss is established by comparing them with real-world spike stream. Finally, the loss is backpropagated to update the learnable parameters (mean position $\boldsymbol{\mu}$, 3D covariance $\boldsymbol{\Sigma}$, opacity $\alpha$ and color $c$) of the Gaussian ellipsoids for optimization. In Section 3, we will model the event stream generation process based on 3D Gaussian distributions and derive the gradient backpropagation process for our model.} 
\label{fig:diagram}
\end{figure}

Based on the above, we leverage the multi-view consistency of 3D Gaussian and the tile-based multi-threaded parallel rendering mechanism in conjunction with spiking neurons to establish robust self-supervision, mitigating the impact of erroneous measurements under high noise levels and diverse illumination conditions. At the same time, we achieve high-quality real-time rendering. The main contributions of this paper are:

1) We proposed a novel differentiable rendering framework that learns 3D Gaussian fields solely from spike stream~(Fig.~\ref{fig:backpropagation}). SpikeGS~(Fig.~\ref{fig:diagram}) exhibits high robustness in extremely noisy, low-quality lighting scenarios. 

2) We proposed a novel spike stream rendering loss function based on 3D Gaussian splatting (3DGS) and spiking neurons capable of generalizing across varying illumination conditions~(Fig.~\ref{fig:qualitative_real}).

3) Experiments on synthetic and real datasets demonstrate that our method outperforms prior state-of-the-art implicit neural rendering methods in terms of rendering quality and speed.

\section{Related work}

\subsection{Neural Radiance Fields and 3DGS}
NeRF \cite{mildenhall2021nerf} employs MLPs to represent neural implicit fields and has garnered widespread attention for its excellent performance in synthesizing high-quality novel views and accurately representing 3D scenes. Improved works based on NeRF have also emerged subsequently. In terms of fast rendering, instant-ngp \cite{muller2022instant} replaces NeRF's fully connected neural networks with a smaller MLP and introduces multi-resolution hash encoding to enhance NeRF's rendering speed. In the realm of sparse view reconstruction, methods like PixelNeRF \cite{yu2021pixelnerf} and RegNeRF \cite{niemeyer2022regnerf} achieve high-quality novel view synthesis using minimal input images. In the domain of deblurring, approaches like Deblur-NeRF \cite{ma2022deblur} and DP-NeRF \cite{lee2023dp} aim to reconstruct clear scene representations from blurred input views by modeling the physical processes of motion blur. Enhancing rendering quality, Mip-NeRF \cite{barron2021mip} introduces a sampling strategy based on view frustum for NeRF-based anti-aliasing and addressing aliasing artifacts. Works such as Block-NeRF \cite{tancik2022block} and BungeeNeRF \cite{xiangli2022bungeenerf} extend NeRF's rendering scale from small to city-scale large scenes. 
A recent revolutionary 3D reconstruction method, 3D Gaussian Splatting (3DGS)~\cite{kerbl20233d}, represents the scene using optimizable Gaussian ellipsoids, which is fundamentally different from NeRF's MLP-based implicit representation. The novel representation of 3DGS makes it possible to render images in real-time, furthermore improving the training time. A plenty of 3DGS-based techniques  \cite{yu2024mip, lee2024deblurring, chen2024text, matsuki2024gaussian, charatan2024pixelsplat, jiang2024gaussianshader} have been proposed recently. Deformable-GS\cite{yang2024deformable} and 4DGS\cite{wu20244d} proposed dynamic scene reconstruction based on 3D Gaussians. VastGS\cite{lin2024vastgaussian} extended the reconstruction capabilities of 3D Gaussians to large-scale scenes. SuGaR\cite{guedon2024sugar} and 2DGS \cite{huang20242d} improve surface fitting by flattening Gaussian ellipsoids, allowing them to better conform to the scene's surface.

\subsection{Scene reconstruction based on event cameras and spike cameras}
Event cameras generate events asynchronously based on changes in scene brightness. Each event records the pixel position, timestamp of occurrence, and polarity change. Spike cameras capture the absolute brightness of each pixel and output spike stream. Specialized methods like Event-NeRF \cite{rudnev2023eventnerf} and Ev-NeRF \cite{hwang2023ev} have been proposed to derive neural radiance fields directly from event streams. E2NeRF \cite{qi2023e2nerf} integrates event data and blurred frames to guide the reconstruction of clear radiance fields from blurry inputs. Evdeblurnerf \cite{cannici2024mitigating} combines a series of previous works \cite{pan2019bringing, lee2023dp, ma2022deblur, qi2023e2nerf}, integrating blur kernels, adaptive weighting networks, and the EDI \cite{pan2019bringing} model with an event camera, achieving state-of-the-art (SOTA) results in deblurring reconstruction on NeRF. \cite{yu2024evagaussians} and \cite{xiong2024event3dgs} introduce event cameras into the framework of 3D Gaussian based on previous NeRF methods to assist in deblurring reconstruction. EvGGS \cite{wang2024evggs} introduces a generalized event-based Gaussian splatting learning framework. SpikeNVS \cite{dai2024spikenvs} combines spike stream and RGB images synergistically to recover clear neural radiance fields from blurry inputs. SpikeNeRF \cite{zhu2024spikenerf} and Spike-NeRF \cite{guo2024spike} both propose novel view synthesis methods based on neural radiance fields. However, constrained by the limitations of neural radiance fields, they struggle to recover fine texture details of scenes and and have extremely slow training speed, whereas \cite{guo2024spike} focuses only on simple synthetic datasets without significant noise. 


In general, although spike cameras offer advantages that traditional cameras lack, current NeRF-based methods with spike cameras cannot achieve high-quality real-time rendering. Therefore, we propose SpikeGS.

\section{Method}
\subsection{3D Gaussian Splatting}
The 3D Gaussian approach does not rely on neural radiance fields. instead, it represents the scene as a series of 3D Gaussian distributions. Based on the initialized sparse point cloud or randomly generated point cloud, a set of 3D Gaussians, defined as $\boldsymbol{G}$, is parameterized by its mean position $\boldsymbol{\mu} \in \mathbb{R}^{3}$, 3D covariance $\boldsymbol{\Sigma} \in \mathbb{R}^{3 \times 3}$, opacity $\alpha \in \mathbb{R}$ and color $c \in \mathbb{R}^{3}$. $c$ is represented by spherical harmonics for view-dependent appearance. The distribution of each Gaussian is defined as:
\begin{equation}
    \boldsymbol{G}\left( \boldsymbol{X} \right) = e^{- \frac{1}{2}{({\boldsymbol{x} - \boldsymbol{\mu}})}^{\boldsymbol{T}}\boldsymbol{\Sigma}^{- 1}(\boldsymbol{x} - \boldsymbol{\mu})}
\end{equation}
It is essential to note that directly optimizing the covariance
matrix $\boldsymbol{\Sigma}$ can result in a non-positive semi-definite matrix, which would not adhere to the physical interpretation typically associated with covariance matrices. To circumvent this issue, 3D-GS chooses to optimize a quaternion $\boldsymbol{q}$ and a 3D vector $\boldsymbol{s}$ represent rotation and scale, respectively. This approach allows the covariance matrix $\boldsymbol{\Sigma}$ to be reconstructed as follows:
\begin{equation}
    \boldsymbol{\Sigma} = \boldsymbol{R}\boldsymbol{S}\boldsymbol{S}^{\boldsymbol{T}}\boldsymbol{R}^{\boldsymbol{T}}
\end{equation}
where $\boldsymbol{R}$ and $\boldsymbol{S}$ denote the rotation and scaling matrix derived from $\boldsymbol{q}$ and $\boldsymbol{s}$, respectively. There is a complex computational graph to obtain the opacity $\alpha$: $\boldsymbol{q}$ and $\boldsymbol{s}\rightarrow\boldsymbol{\Sigma}$, $\boldsymbol{\Sigma} \rightarrow \boldsymbol{\Sigma}^{'}$, $\boldsymbol{\Sigma}^{'}\rightarrow\alpha$ \cite{chen2024survey}.


To enable differentiable Gaussian rasterization, 3D gaussians are projected into the 2D image space from a given camera pose $\boldsymbol{T}_{\boldsymbol{c}} = \left\{ \boldsymbol{R}_{\boldsymbol{c}} \in \mathbb{R}^{3 \times 3},\boldsymbol{~}\boldsymbol{t}_{\boldsymbol{c}} \in \mathbb{R}^{3} \right\}$ for rasterizing and rendering using the following equation, as described in \cite{zwicker2001ewa}. Given the viewing transformation $\boldsymbol{W}$ and 3D covariance matrix $\boldsymbol{\Sigma}$, the projected 2D covariance matrix $\boldsymbol{\Sigma}^{\boldsymbol{'}}$ is computed using:
\begin{equation}
    \boldsymbol{\Sigma}^{'} = \boldsymbol{J}\boldsymbol{W}\boldsymbol{\Sigma}\boldsymbol{W}^{T}\boldsymbol{J}^{T}
\end{equation}
where $\boldsymbol{J}$ is the Jacobian of the affine approximation of the projective transformation.

Subsequently, the transformed Gaussian ellipsoids are sorted based on their depth and the sorted Gaussian ellipsoids are rasterized to render pixel values using the following volume rendering equation:
\begin{equation}
    C = {\sum\limits_{i \in N}c_{i}}\alpha^{'}_{i}{\prod\limits_{j = 1}^{i - 1}\left( 1 - \alpha^{'}_{j} \right)}
\end{equation}
where $c_{i}$ is the learned color and the final opacity $\alpha^{'}_{i}$ is
the multiplication result of the learned opacity $\alpha_{i}$ and the Gaussian:
\begin{equation}
{\alpha^{'}}_{i} = \alpha_{i} \times exp\left( - \frac{1}{2}\left( {\boldsymbol{x}^{'} - \boldsymbol{\mu}_{\boldsymbol{i}}^{'}} \right)^{T}{\boldsymbol{\Sigma}_{\boldsymbol{i}}^{'}}^{- 1}\left( {\boldsymbol{x}^{'} - \boldsymbol{\mu}_{\boldsymbol{i}}^{'}} \right) \right)
\end{equation}
where $\boldsymbol{x}^{'}$ and $\boldsymbol{\mu}_{\boldsymbol{i}}^{'}$ are coordinates in the projected space.

\subsection{Spike Signal Model}
A spike camera reflects the light intensity of a scene through discharge events that occur when the voltage of a photodiode is released to the reference voltage (the received incoming photons will be transferred to voltage). The accumulator at each pixel accumulates the light intensity. For a pixel at position $(x,y)$, if the accumulated intensity reaches the scheduling threshold $\Omega$, a spike is emitted. Simultaneously, the corresponding accumulator is reset by subtracting the scheduling threshold from its own intensity. As shown in Equation (6) below, $A_{t_{i}}$ and $A_{t_{i-1}}$  represent the values of the accumulator at time $t_{i}$ and $t_{i-1}$, respectively. $I_{t_{i}}$ represents the input value at time $t_{i}$.
\begin{equation}
    A_{t_{i}} = \left( {A_{t_{i-1}} + I_{t_{i}}} \right)\boldsymbol{~}mod\boldsymbol{~}\Omega
\end{equation}
The integral form of the accumulator voltage can be expressed as:
\begin{equation}
    A{({x,y,T})} = {\int_{0}^{T}\eta} \cdot I{({x,y,t})}dt~mod~\Omega
\end{equation}
where $I{({x,y,t})}$ represents the light intensity at pixel $(x,y)$ at time $t$, and $\eta$ is the photoelectric conversion rate.
We will directly use $I\left( t \right)$ to represent the luminance intensity to simplify our presentation. Due to the limitations of circuits, each spike is read out at discrete time $t$, $t \in T$ ($T=Nt$, where t represents the unit time step and N is the size of the time window). Thus, the output of the spike camera is a spatial-temporal binary stream S with $H \times W \times N$ size. Here, $H$ and $W$ are the height and width resolution of the sensor, respectively, and N is the temporal window size of the spike stream. In our experiments, we set N to 256. The process of spike emission can be represented by the following equation:
\begin{equation}
S_{t_{i}} = 
\begin{cases}
1, & \textit{if  } A_{t_{i-1}} + I_{t_{i}} \geq \Omega \\
0, & \textit{otherwise}
\end{cases}
\end{equation}
where, 0 indicates no spike, while 1 indicates a spike is sent. 

\subsection{Spike Noise Model}
Inspired by \cite{zhu2023recurrent}, we understand that due to significant circuit differences, the noise characteristics of spike cameras differ greatly from traditional cameras. The photodetectors in spike cameras receive photons from the scene, but even under uniform illumination, the photons striking the diodes are not constant. The difference between the number of photons at a given moment and the ideal number of photons is referred to as shot noise $N_p$. The dark current noise generated by the thermal diffusion of charge carriers and defects within or on the surface of a PN junction is denoted as $N_d$. Additionally, differences in photodiode characteristics and capacitance contribute to variations in pixel sensitivity to incident light intensity, resulting in photo-response non-uniformity noise $N_{rnu}$. Moreover, the temporal delay between the generation of the reset signal and the subsequent release of the spike signal introduces quantization noise $N_q$. If the temporal length of the spike stream is not long enough, truncation noise will appear in the process from the spike to the image. we define truncation noise as $N_c$. Therefore, based on the above, we can define the equation for pulse noise as:
\begin{equation}
    I + N = \frac{1}{\frac{Q_{r}}{L + N_{p} + N_{d}} + N_{rnu} + N_{q}} + N_{c}
\end{equation}
Where $I$ is the ideal image without noise, $N$ is the total noise, $L$ represents the scene light intensity, $Q_r$ is the relative quantity matrix of electric charge. $N_p$, $N_d$, $N_{rnu}$, $N_q$, and $N_c$ represent shot noise, dark current noise, response nonuniformity noise, quantization noise, and truncation noise, respectively.
\begin{figure}[tb] 
\centering
\includegraphics[width=0.55\linewidth]{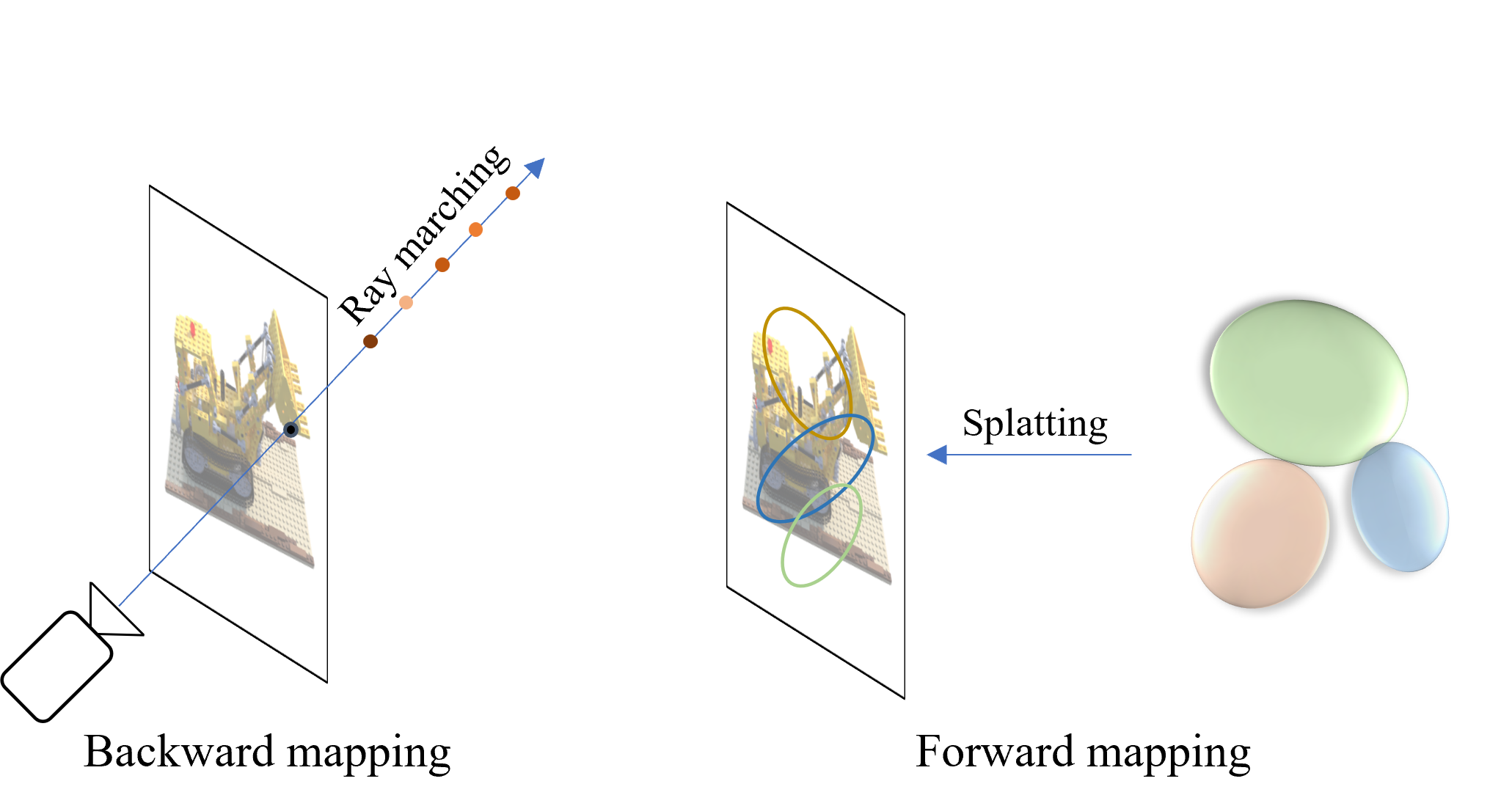}
\caption{Rendering Diagrams for 3D Gaussians and NeRF.}
\label{fig:mapping}
\end{figure}
\subsection{SpikeGS}
The first spike-camera-based 3D Gaussian Splatting framework, SpikeGS, is introduced in this section. An integrate-and-fire mechanism is utilized to simulate the generation of spikes from intensity, which is estimated by splatting 3D Gaussians into an image plane. The noise generation mechanism is considered to imitate how a spike camera works under real scenarios, especially low-illumination scenarios. Simulated spike stream are compared with real spike stream to indicate 3D Gaussians representing 3D scenes correctly. Details are demonstrated as follows. 

\textbf{Establish the relationship between rendered pixel values and the real-world light intensity values $I$.} Before establishing the relationship between rendering spike stream and real spike stream, we first establish their relationship in terms of light intensity. The goal is to estimate the intensity values corresponding to the real light of the scene. Unlike NeRF, which is a backward mapping process that operates on a per-pixel basis, emitting rays from 2D pixels and integrating along sampled points on the ray via volume rendering to synthesize the corresponding pixel values shown as Fig.~\ref{fig:mapping}. On the other hand, 3D Gaussians utilize forward mapping. In this forward mapping, 3D Gaussian ellipsoids are splatted onto the 2D image plane through a rasterization pipeline (Each ellipsoid corresponds to multiple two-dimensional pixels). 
Therefore, in models based on 3D Gaussians, we cannot directly establish the relationship between the pixel ray $r$ and the real-world light intensity values $I$. Therefore, we chose to establish the relationship between the view(pose) and the real-world light intensity values $I$. We denote the estimated intensity value corresponding to the real light of the scene as:
\begin{equation}
    \begin{aligned}
    \hat{I}\left( R_{G}(P),t \right), \quad R_{G}(P) = \left\{ C(x) \middle| x \in X \right\}  
    \end{aligned}
\end{equation}
where $R_G(P)$ represents the rendering image $R_{G}$ of Gaussian Splatting $G$ from the pose $P$. $C(x)$ represents the pixel value at the rendered pixel position $x$. $X$ denotes the set of all pixels in the image space.

\noindent \textbf{Noise embedding.} To supervise $\hat{I}\left( R_{G}(P),t \right)$ using real noise spike stream, we need to consider noise in multiple scenarios. As stated in Section 3.3, we can establish the following relationship:
\begin{equation}
    \hat{I}\left( R_{G}(P),t \right) + I\left(N,t\right) = \frac{1}{\frac{Q_{r}}{L + N_{p} + N_{d}} + N_{rnu} + N_{q}} + N_{c}
\end{equation}
The above equation can be written as: $\hat{I}\left( R_{G}(P),t \right) + I\left(N,t\right) = I\left(S,t\right)$, $N$ represents the total noise, $I\left(N,t\right)$ represents the intensity variation caused by noise, and $I\left(S,t\right)$ represents the intensity variation capturing the real spike stream. In fact, the deviation matrix $R\left(x,y\right)$\cite{zhu2024spikenerf} corresponding to the response nonuniformity noise can be obtained by capturing a uniform light scene and recording the intensity. So we use the matrix  $R\left(x,y\right)$ to simulate noise embedding. Choosing the pixel $(x_{m}, y_{m})$ which is closest to the average response value as the reference pixel. $R\left( {x,y} \right)$ is obtained by calculating the ratio of the response value of a reference pixel to the response values of other pixels. Finally, we can rewrite Equation (11) as $\hat{I}\left( R_{G}(P),t \right) \cdot R\left( {x,y} \right) = I\left(S,t\right)$.

\begin{figure}[tb] 
\centering
\includegraphics[width=\linewidth]{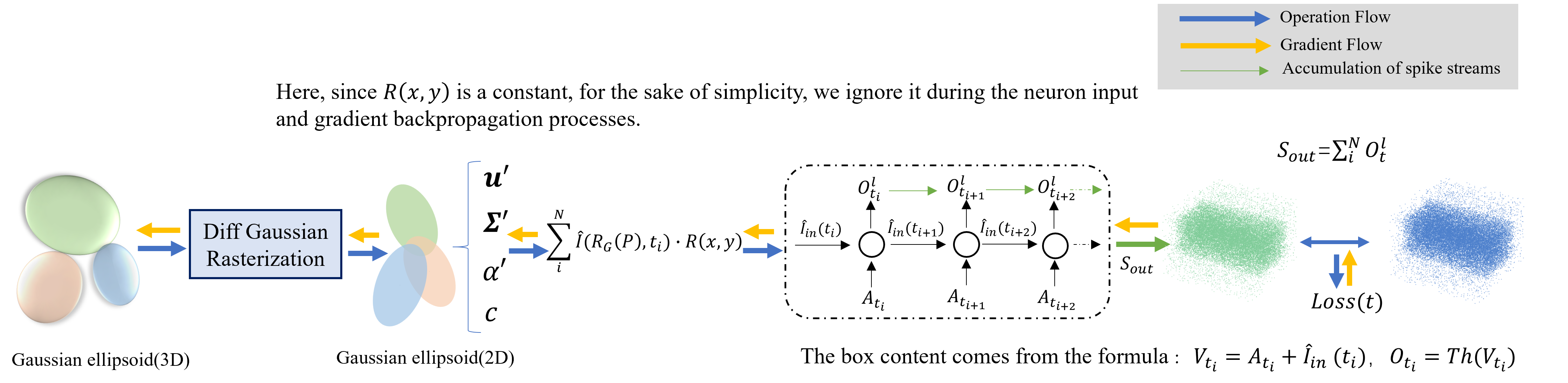}
\caption{Diagram of the backpropagation process in spiking neurons based on 3D Gaussians. After passing through the rasterization pipeline, three-dimensional Gaussian ellipsoids are projected onto a two-dimensional plane, resulting in two-dimensional Gaussian ellipsoids (rendered pixels) with corresponding learnable parameters $\theta$ ($\boldsymbol{u}^{'}$, $\boldsymbol{\Sigma}^{'}$, $c$, $\alpha$). Using $N=T$ as the step size (window size) and discrete time steps as units, the predicted light intensity values are input into the spiking neurons. Each spiking neuron receives the accumulated value $A_{t}$ from the previous moment and the current input $\hat{I}(R_G\left(P\right),t)$ (summed as $V_{t}$). The output consists of a spike stream $S$ and the residual voltage in the accumulator after reset, which is used as the input for the next neuron.} 
\label{fig:backpropagation}
\end{figure}

\noindent \textbf{Spike neuron layer.} Based on Section 3.2, we construct the process of converting scene light intensity into spike stream using an integrate-and-fire mechanism \cite{gerstner2014neuronal} in the spike camera as a spike neuron layer (Fig.~\ref{fig:backpropagation}). We denote it by $SNL$ and represent its discrete form as:
\begin{equation}
    \begin{aligned}
    A_{t} = A_{t - 1} + I_{in}(t), \quad S_{t} = Thr\left( A_{t} \right)
    \end{aligned}
\end{equation}
Where $A_{t}$ represents the value of the accumulator at time $t$, $A_{t-1}$ represents the value of the accumulator at the previous time step, and $I_{in}(t)$ represents the input value at time $t$. Function $Thr$ is represented by formula (8). When $A_{t}$ is greater than or equal to the threshold, it outputs 1; otherwise, it outputs 0. Note that if the accumulator's value exceeds the threshold, it releases a potential and then resets itself by subtracting the threshold from its own value.

\noindent \textbf{Spike stream rendering loss.} To measure the difference between the generated spike stream and the input spike stream, we propose a spike stream rendering loss function based on 3D Gaussians:
\begin{equation}
\left. L = \left( {1 - \lambda} \right) || \hat{S}(G,t) - S_{gt}{||}_{1}~ + ~\lambda\left( {\hat{S}\left( {G,t} \right),S_{gt}} \right)_{D - SSIM} \right.
\end{equation}
Where $\hat{S}\left( {G,t} \right) = SNL \left(\hat{I}\left( R_{G}(P),t \right) \cdot R\left( {x,y} \right)\right) $. Note that for the Structural Similarity assessment, we transposed the shape of the spike stream to $N\times H \times W$, where $N$ is the spike stream window size.

\noindent \textbf{Gradient Derivation.}
In this section, we will derive the backpropagation process of our SpikeGS model. Based on Section 3.1, we will uniformly represent the learnable parameters of the 2D Gaussian ellipsoid using $\theta$. where $\theta=\left[\boldsymbol{u}^{'} \hspace{3pt} \boldsymbol{\Sigma}^{'} \hspace{3pt} {c} \hspace{3pt} \alpha\right]$. We set the time step of the spiking camera to N (in our experiments, N is set to 256), corresponding to the total time T. As shown in Fig.~\ref{fig:backpropagation}, after rasterizing the Gaussian ellipsoid, we use $\sum_{t}^{T}{\hat{I}(R_G\left(P\right),t)}$ to represent the total light intensity over the entire time step T. The light intensity at each step is processed by the spiking neurons to generate the corresponding spike flow at that moment. The accumulated spike flows from all time steps result in the final output spike stream $S_{out}$.

In Fig.~\ref{fig:backpropagation} (To distinguish the input and output values of the spiking neurons, for convenient we denote $V_{t_{i}} = A_{t_{i-1}} + I_{t_{i}}$. )
we can see that each spiking neuron has two output values: the accumulated value of the accumulator at the next time step $A_{t}$ (including the reset case) and the corresponding transmit signal (emit spike) $O^{l}_{t}$. The accumulated value $A_{t}$ of the accumulator continues to be combined with the light intensity $\hat{I}_{in}(t)$ at the next step as input for the next spiking neuron. 

The network is unrolled for all discrete steps, and the total gradient for all discrete steps is calculated. The gradient transmission formula is shown below:
\begin{equation}
    \frac{\partial L_{total}}{\partial{\hat{I}}_{in}}=\sum_{i}^{N}\frac{\partial L_{total}}{\partial O_{t_{i}}^l}\frac{\partial O_{t_{i}}^l}{\partial V_{t_{i}}}\frac{\partial V_{t_{i}}}{\partial{\hat{I}}_{in}(t_{i})} , \quad \frac{\partial O_{t_{i}}^l}{\partial V_{t_{i}}}={Thr}^\prime\left(V_{t_{i}}\right)
\end{equation}
Since $\frac{\partial O_{t_{i}}^l}{\partial V_{t_{i}}}$ itself is non-differentiable, we refer to \cite{neftci2019surrogate} and use the surrogate gradient method to compute it. The calculation of the learnable parameters for a 2D Gaussian ellipsoid is as follows:
\begin{equation}
    \frac{\partial L_{total}}{\partial\boldsymbol{\theta}}=\left[\frac{\partial L_{total}}{\partial{\hat{I}}_{in}}\frac{\partial{\hat{I}}_{in}}{\partial\boldsymbol{u}^\prime}\ \ \ \ \frac{\partial L_{total}}{\partial{\hat{I}}_{in}}\frac{\partial{\hat{I}}_{in}}{\partial\boldsymbol{\Sigma}^\prime}\ \ \ \frac{\partial L_{total}}{\partial{\hat{I}}_{in}}\frac{\partial{\hat{I}}_{in}}{\partial\alpha^\prime}\ \ \ \frac{\partial L_{total}}{\partial{\hat{I}}_{in}}\frac{\partial{\hat{I}}_{in}}{\partial{c}}\right]
\end{equation}
Subsequently, we can derive the gradients of the learnable parameters for the 3D Gaussian ellipsoids. For a detailed explanation of the gradient propagation process from 2D to 3D, please refer to the original 3D Gaussian splatting paper \cite{kerbl20233d}. 

In this section, we propose the first learnable spike stream generation pipeline based on 3D Gaussians. Next, we will present the advanced results of our model on both synthetic and real-world datasets in the experimental section.

\section{Experiments}
\begin{figure*}[!htb]
\centering
\begin{minipage}{\textwidth}
\begin{tabular*}{\textwidth}{ccccccc}
\includegraphics[width=0.14\textwidth, height=0.14\textwidth]{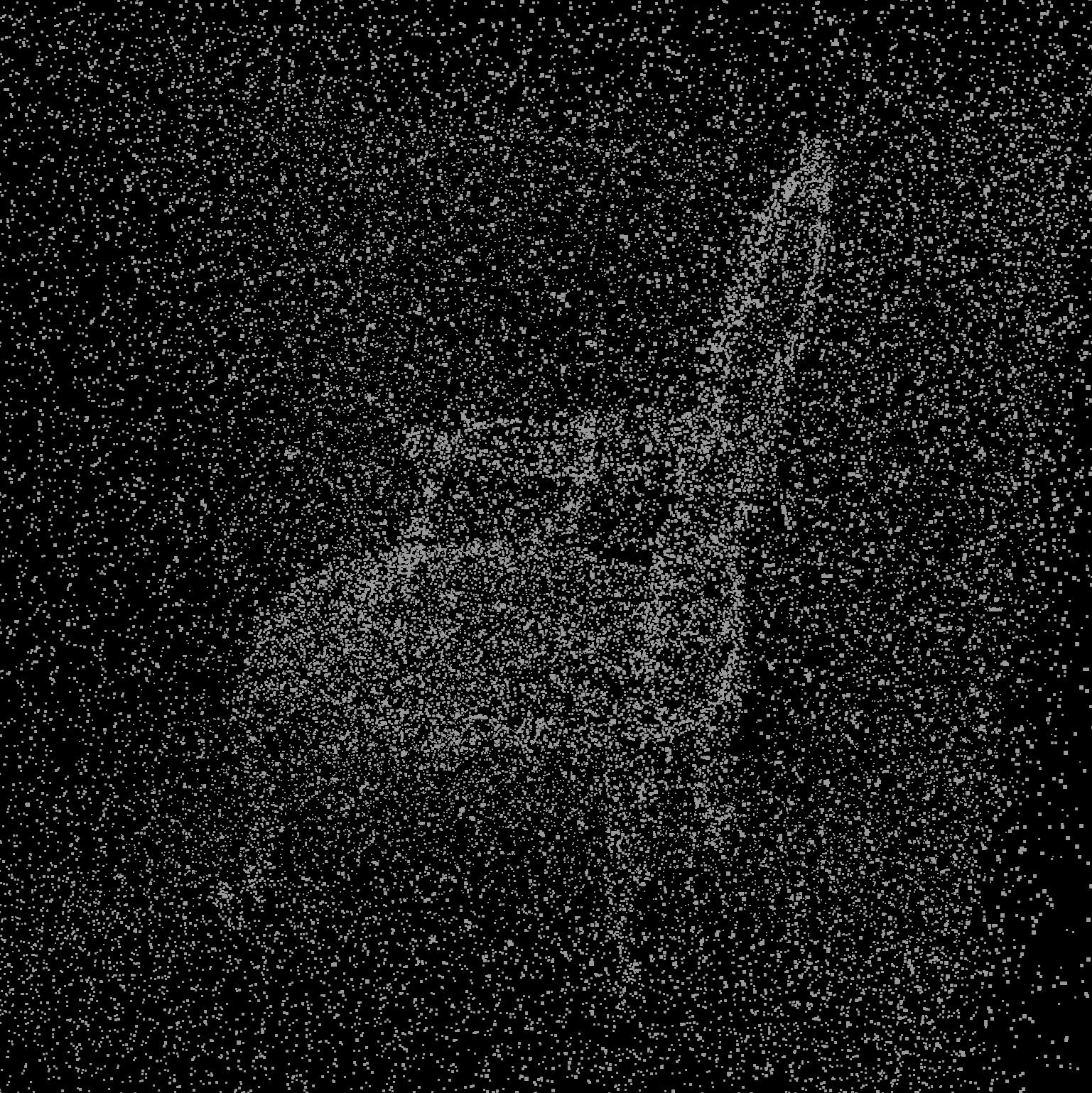} 
&\includegraphics[width=0.14\textwidth, height=0.14\textwidth]{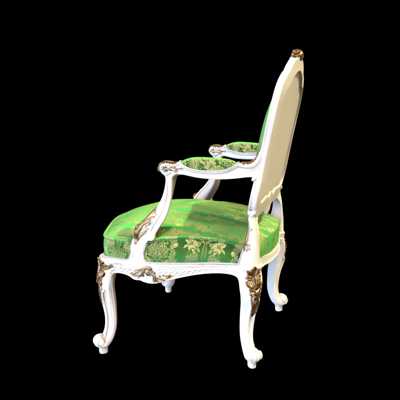} 
&\includegraphics[width=0.14\textwidth, height=0.14\textwidth]{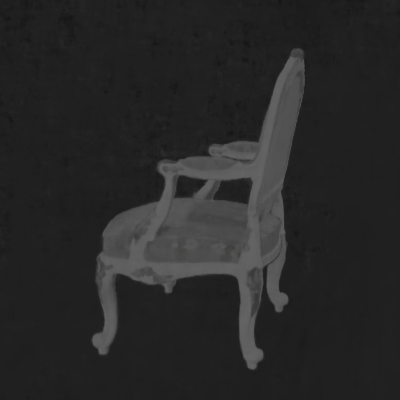} 
&\includegraphics[width=0.14\textwidth, height=0.14\textwidth]{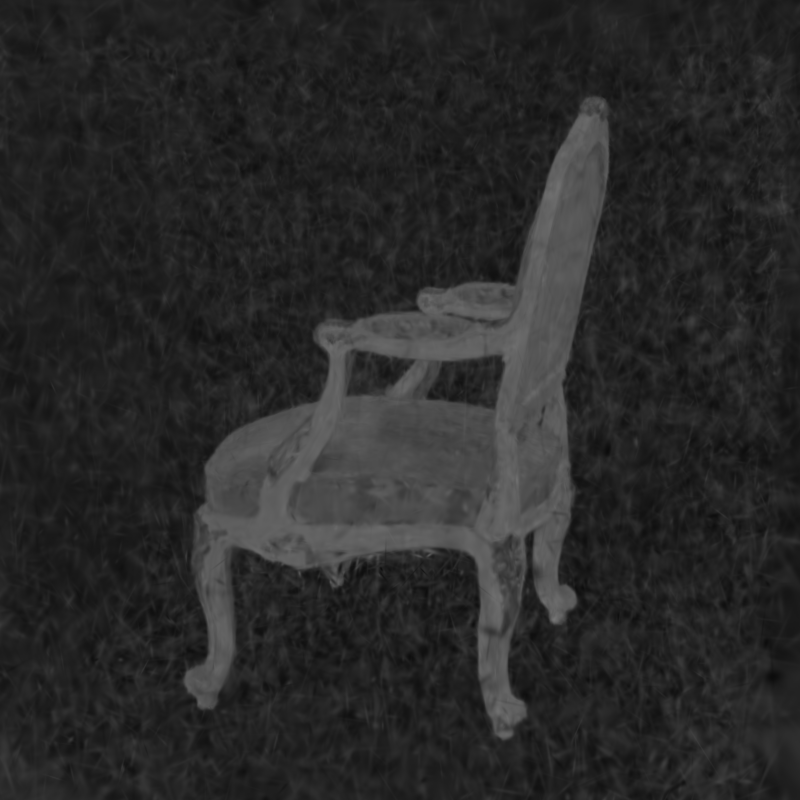}
&\includegraphics[width=0.14\textwidth, height=0.14\textwidth]{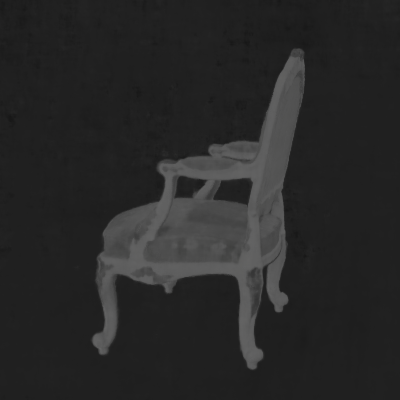}
&\includegraphics[width=0.14\textwidth, height=0.14\textwidth]{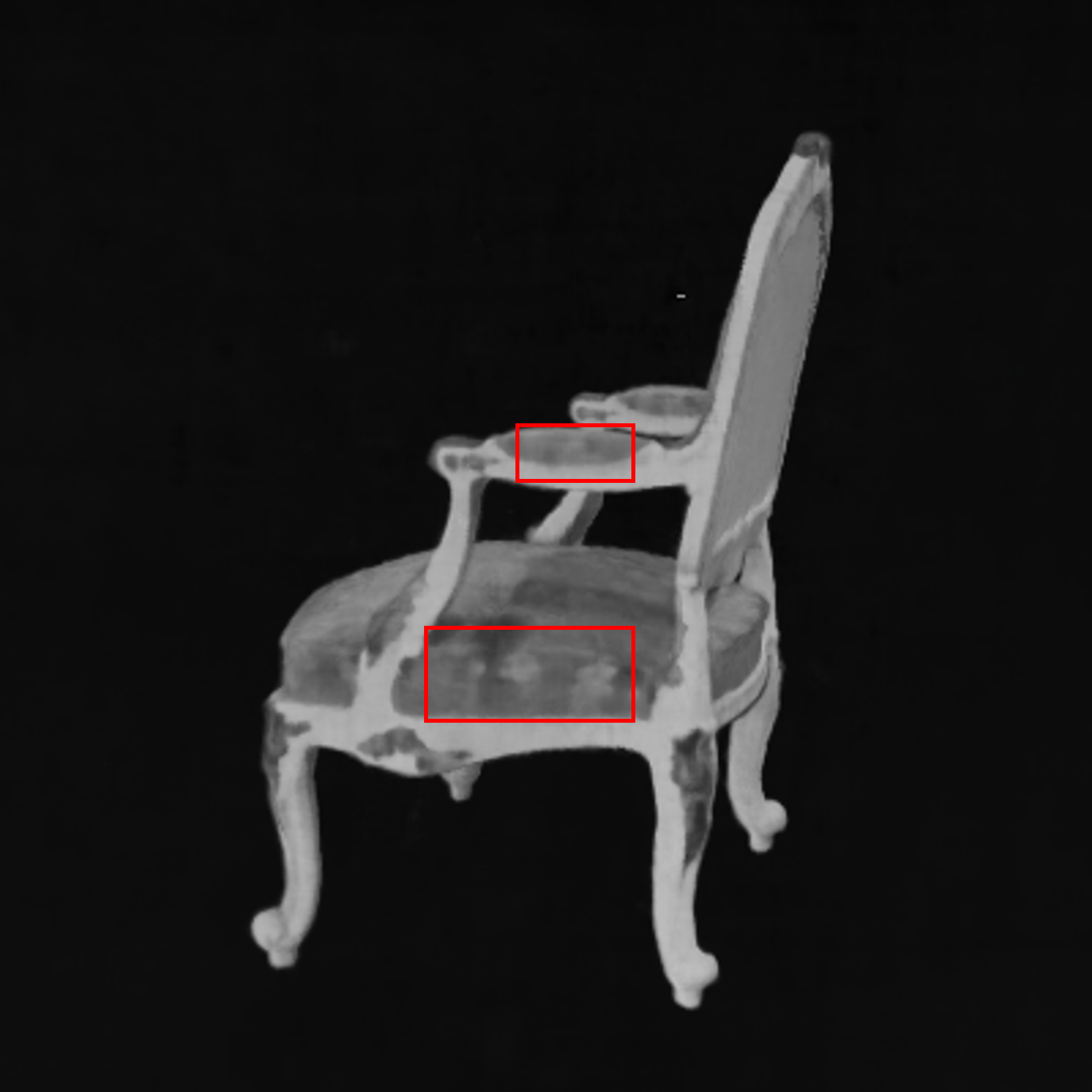} 
&\includegraphics[width=0.14\textwidth, height=0.14\textwidth]{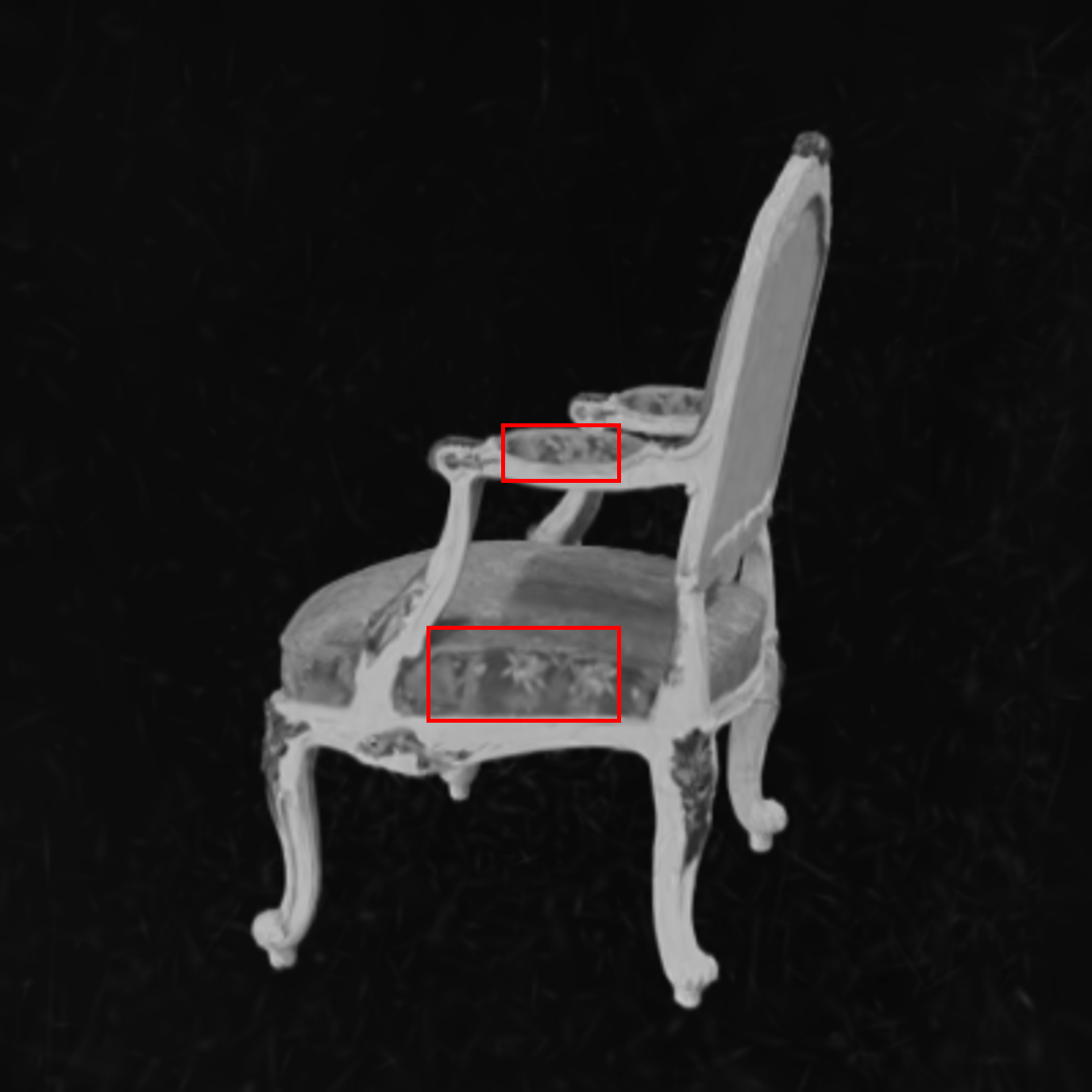}\\

\includegraphics[width=0.14\textwidth, height=0.14\textwidth]{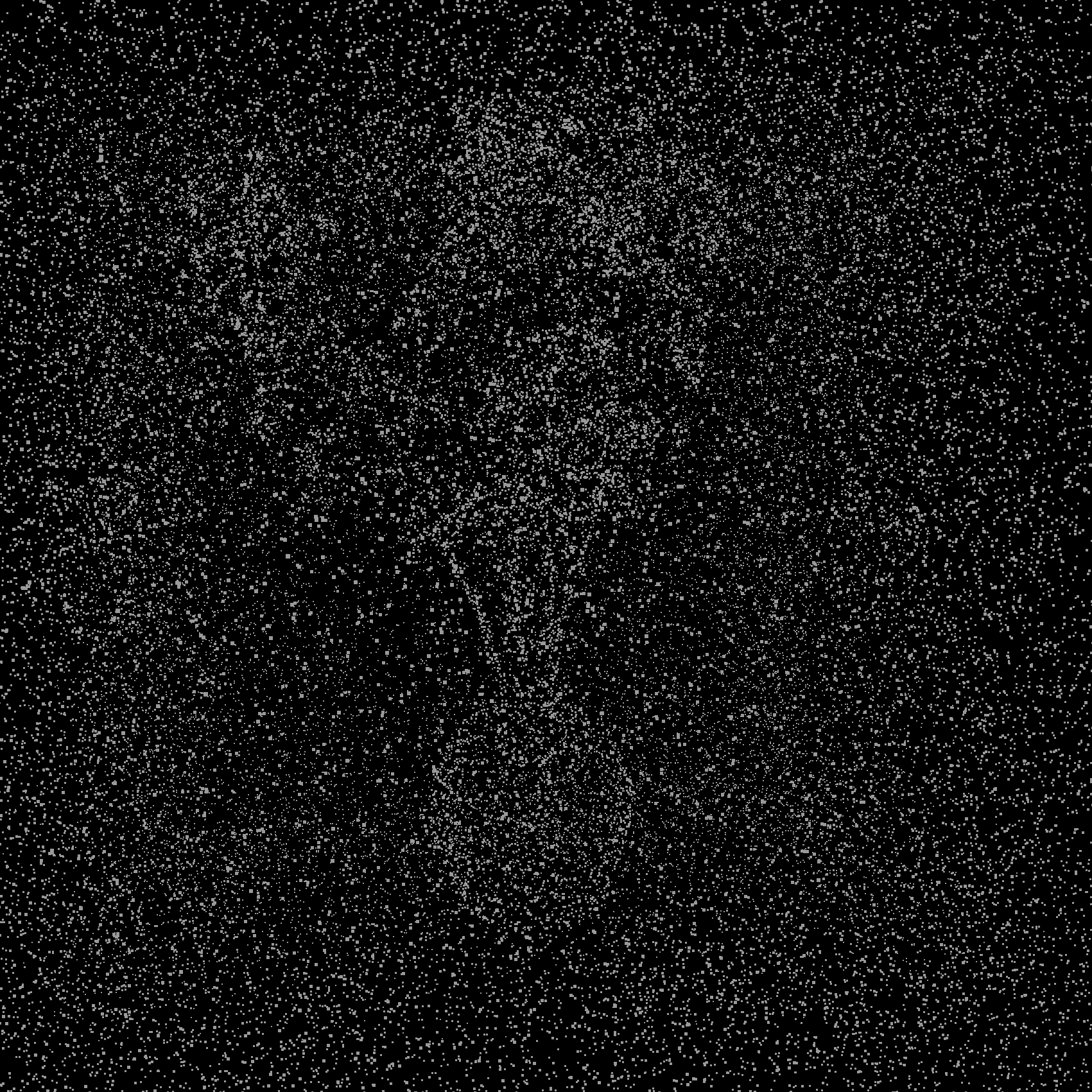} &
\includegraphics[width=0.14\textwidth, height=0.14\textwidth]{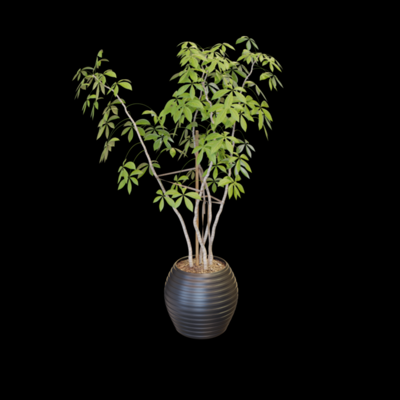} &
\includegraphics[width=0.14\textwidth, height=0.14\textwidth]{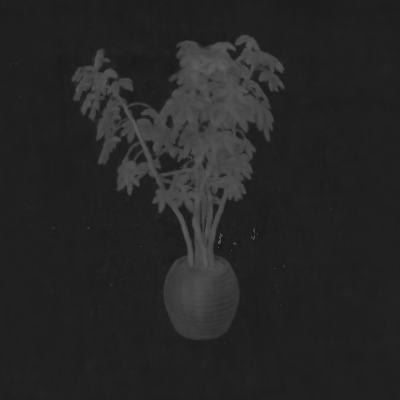} &
\includegraphics[width=0.14\textwidth, height=0.14\textwidth]{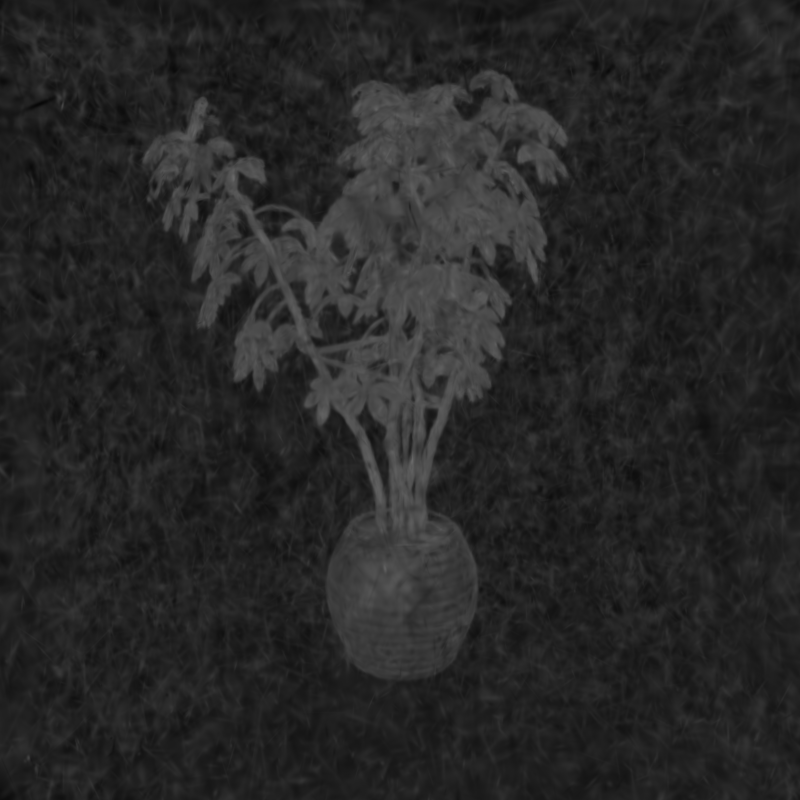} &
\includegraphics[width=0.14\textwidth, height=0.14\textwidth]{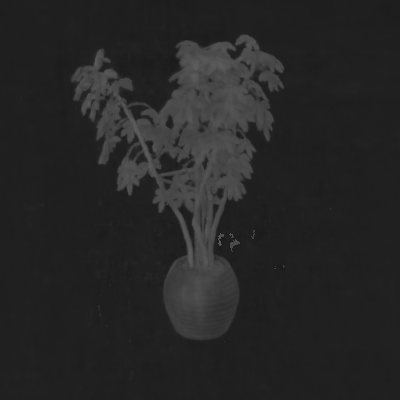} &
\includegraphics[width=0.14\textwidth, height=0.14\textwidth]{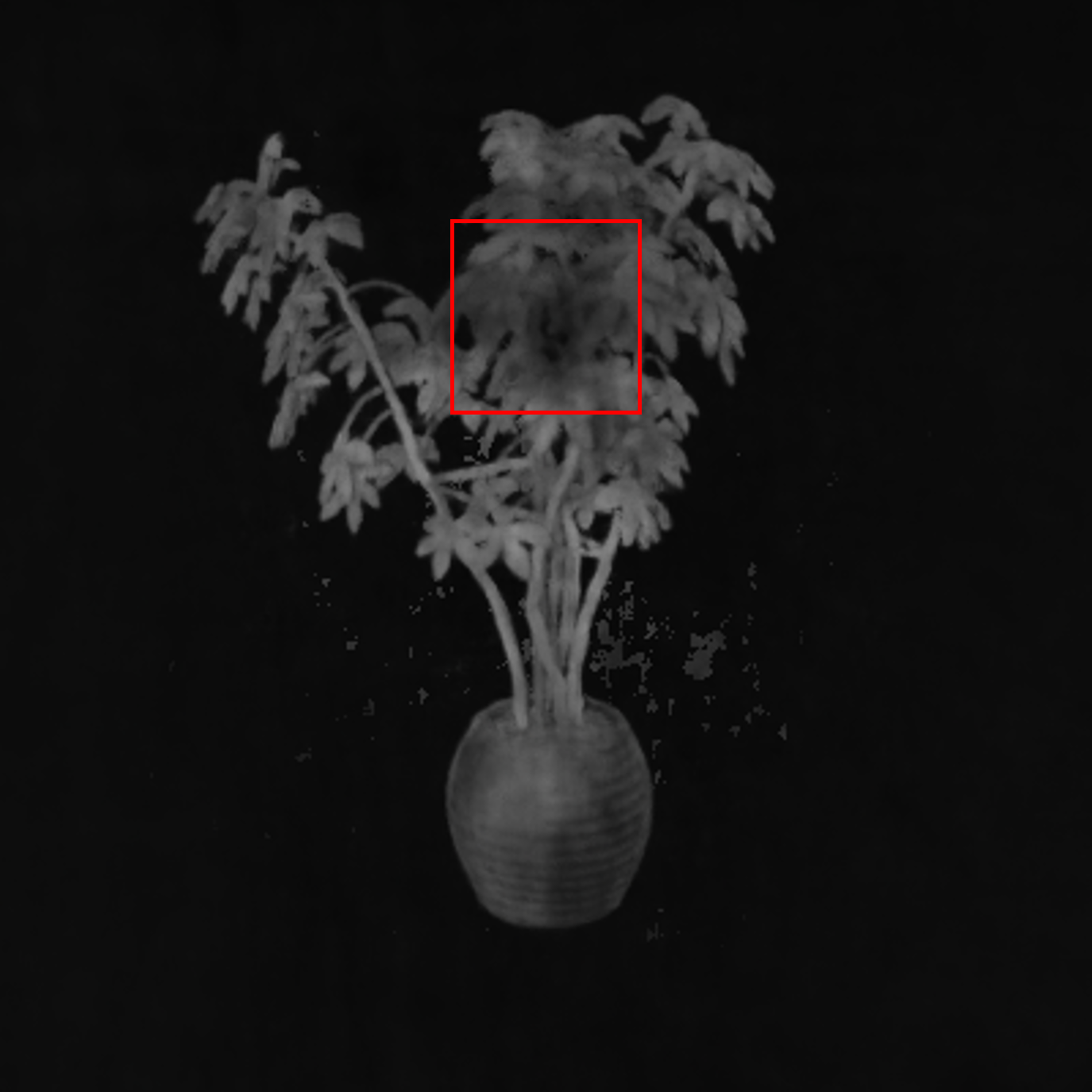} &
\includegraphics[width=0.14\textwidth, height=0.14\textwidth]{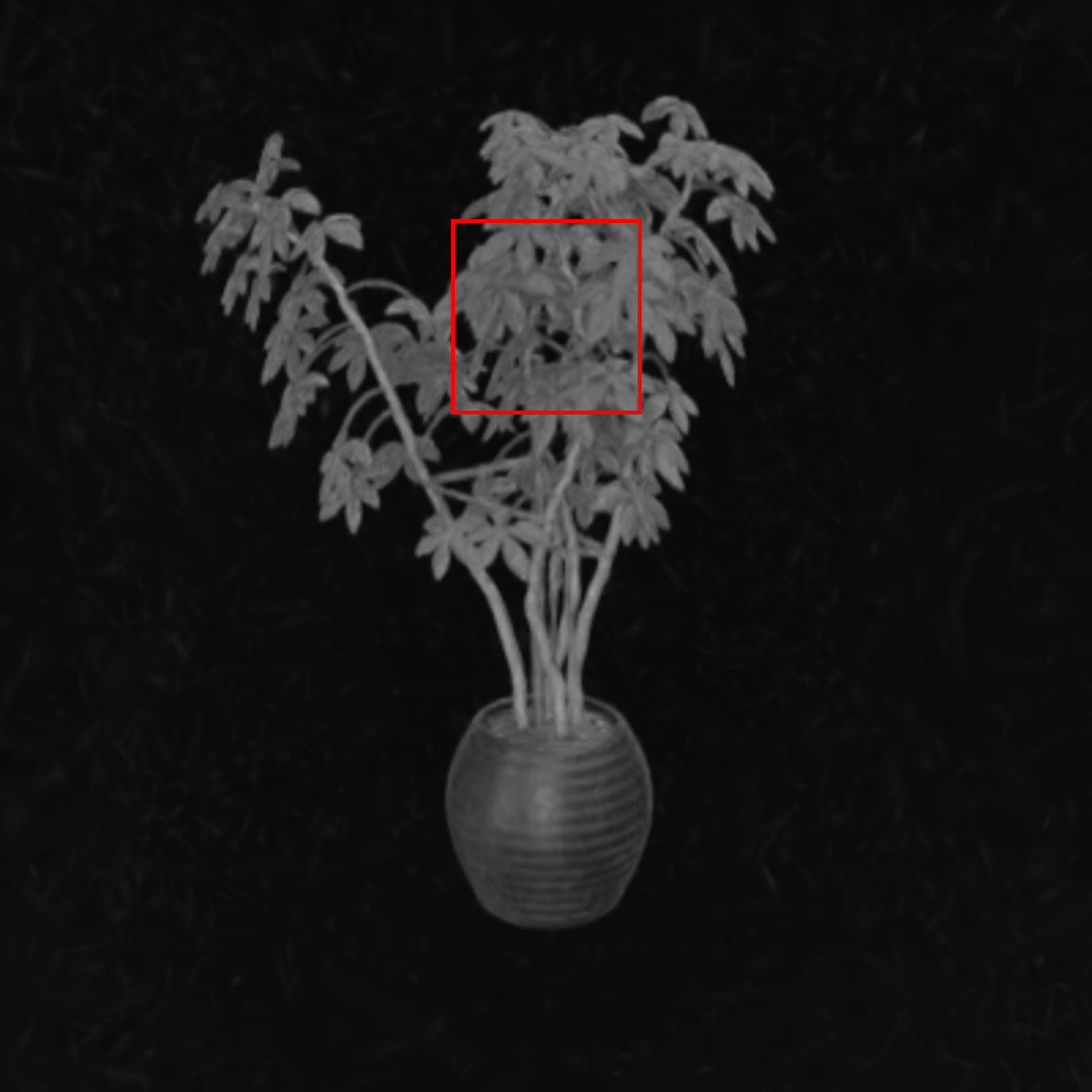}\\

\includegraphics[width=0.14\textwidth, height=0.14\textwidth]{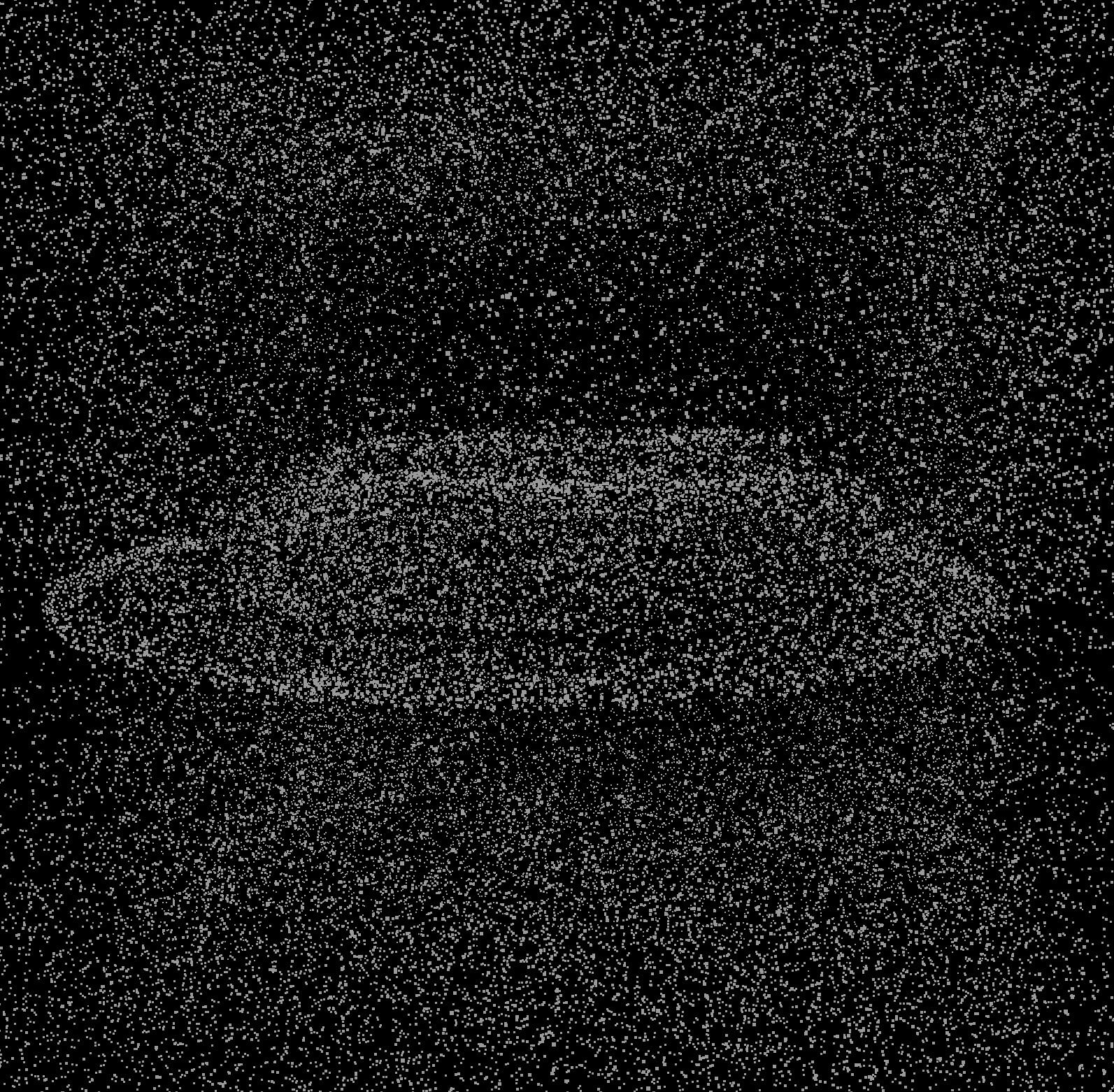} &
\includegraphics[width=0.14\textwidth, height=0.14\textwidth]{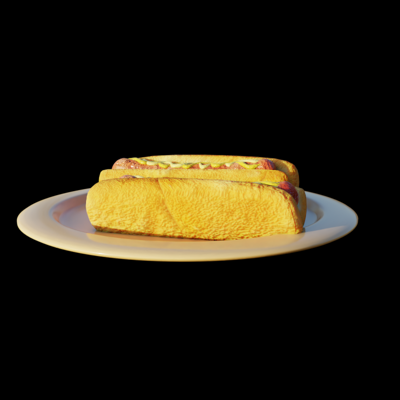} &
\includegraphics[width=0.14\textwidth, height=0.14\textwidth]{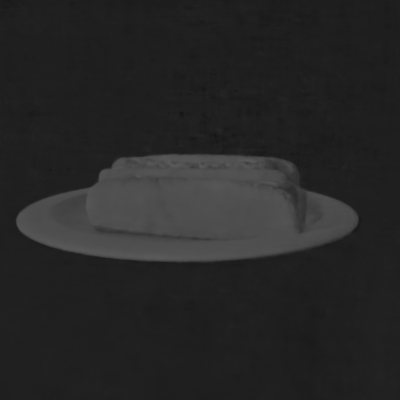} &
\includegraphics[width=0.14\textwidth, height=0.14\textwidth]{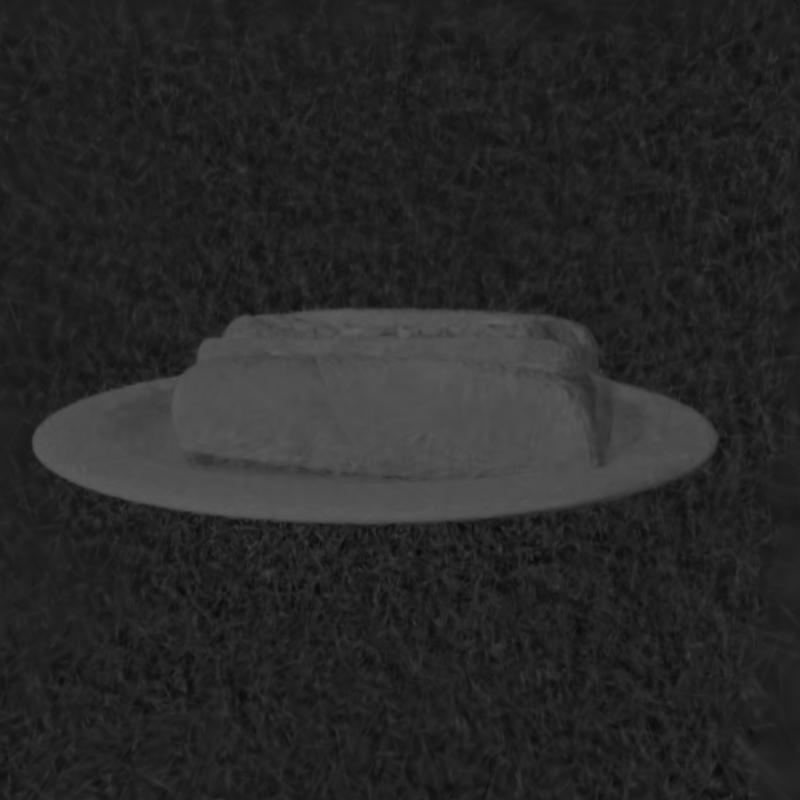} &
\includegraphics[width=0.14\textwidth, height=0.14\textwidth]{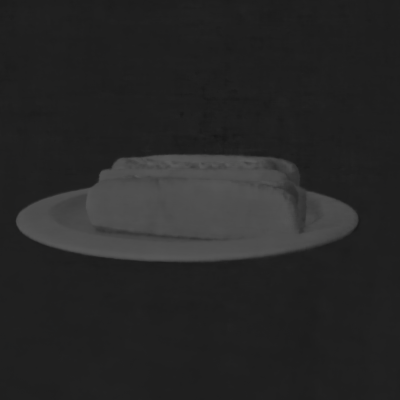} &
\includegraphics[width=0.14\textwidth, height=0.14\textwidth]{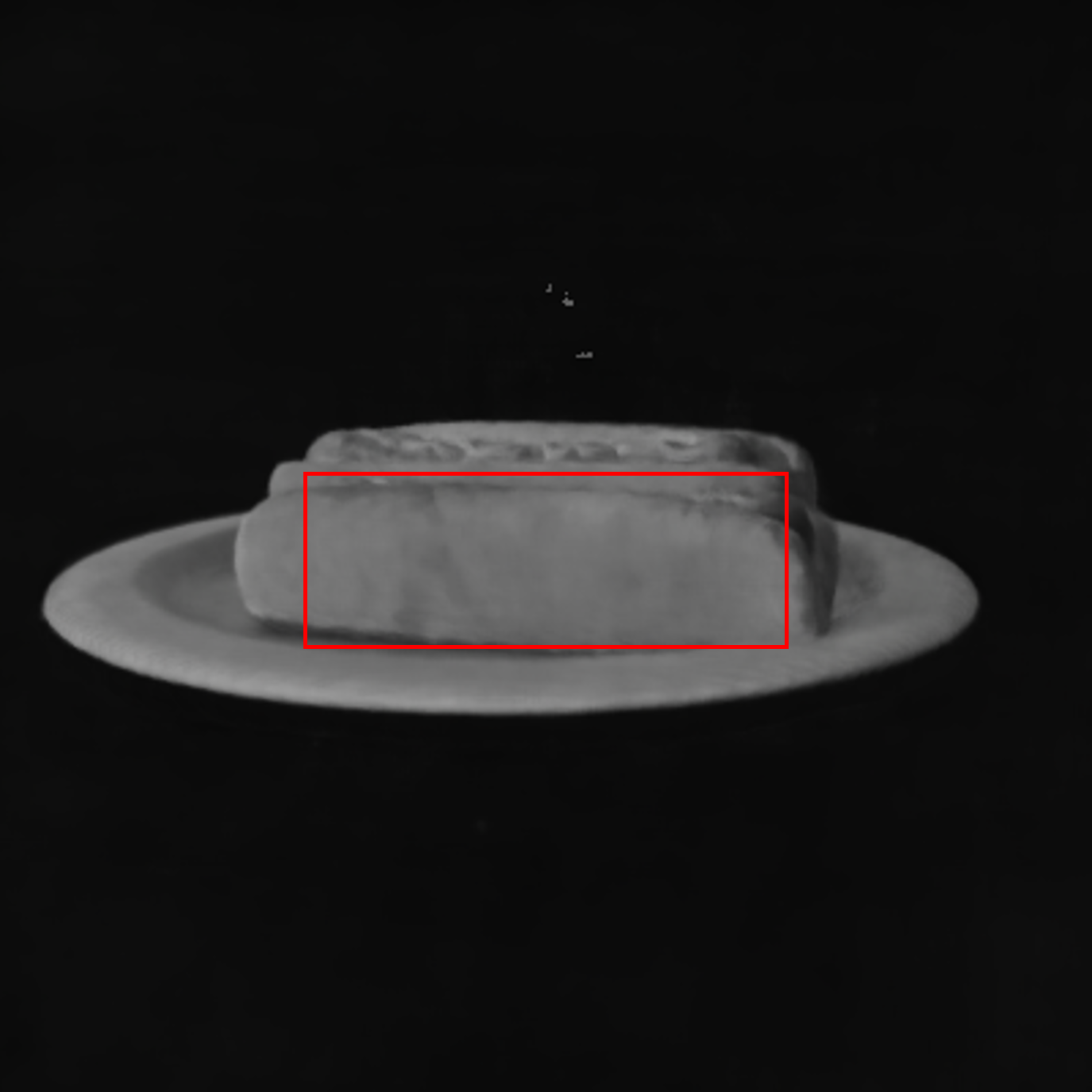} &
\includegraphics[width=0.14\textwidth, height=0.14\textwidth]{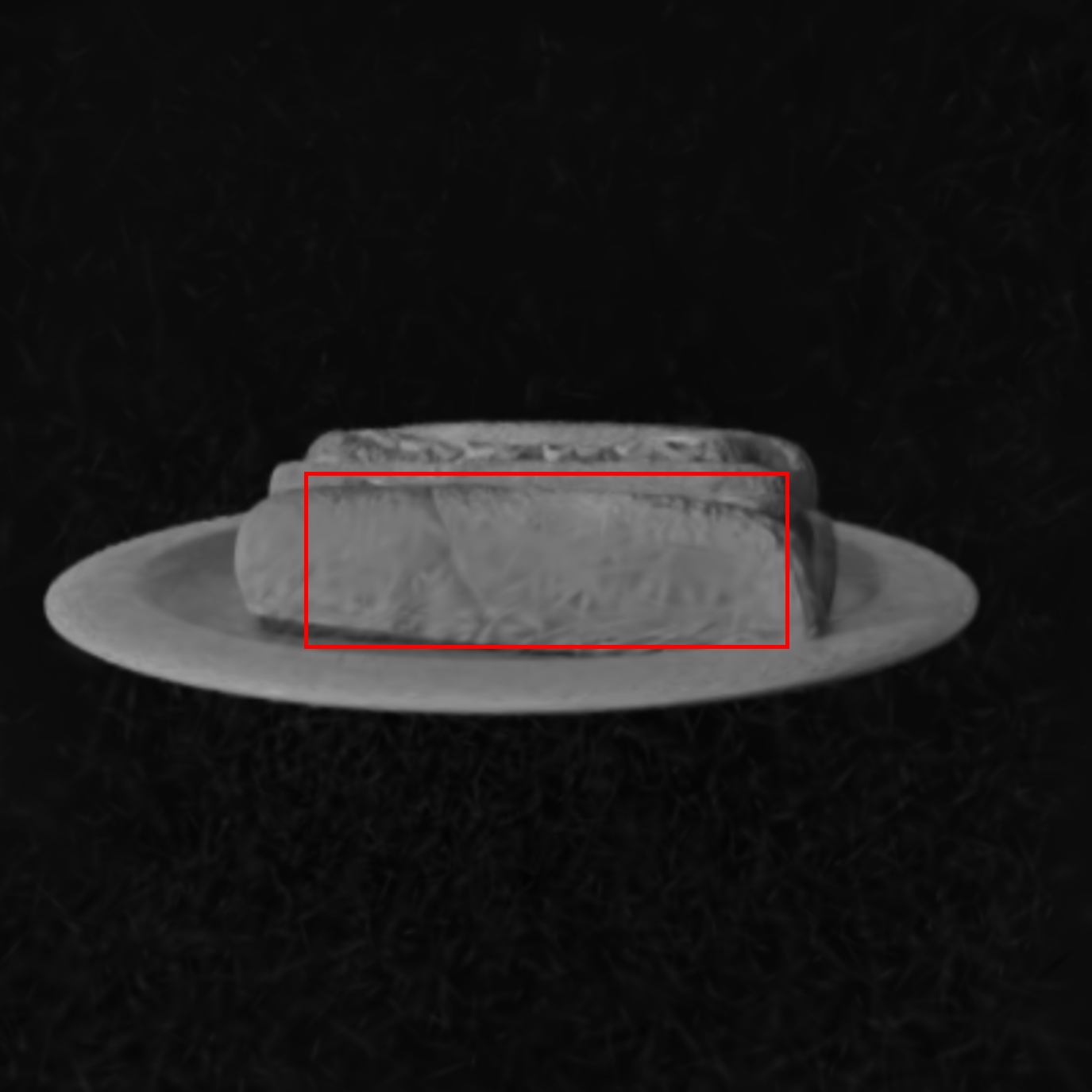}\\

\includegraphics[width=0.14\textwidth, height=0.14\textwidth]{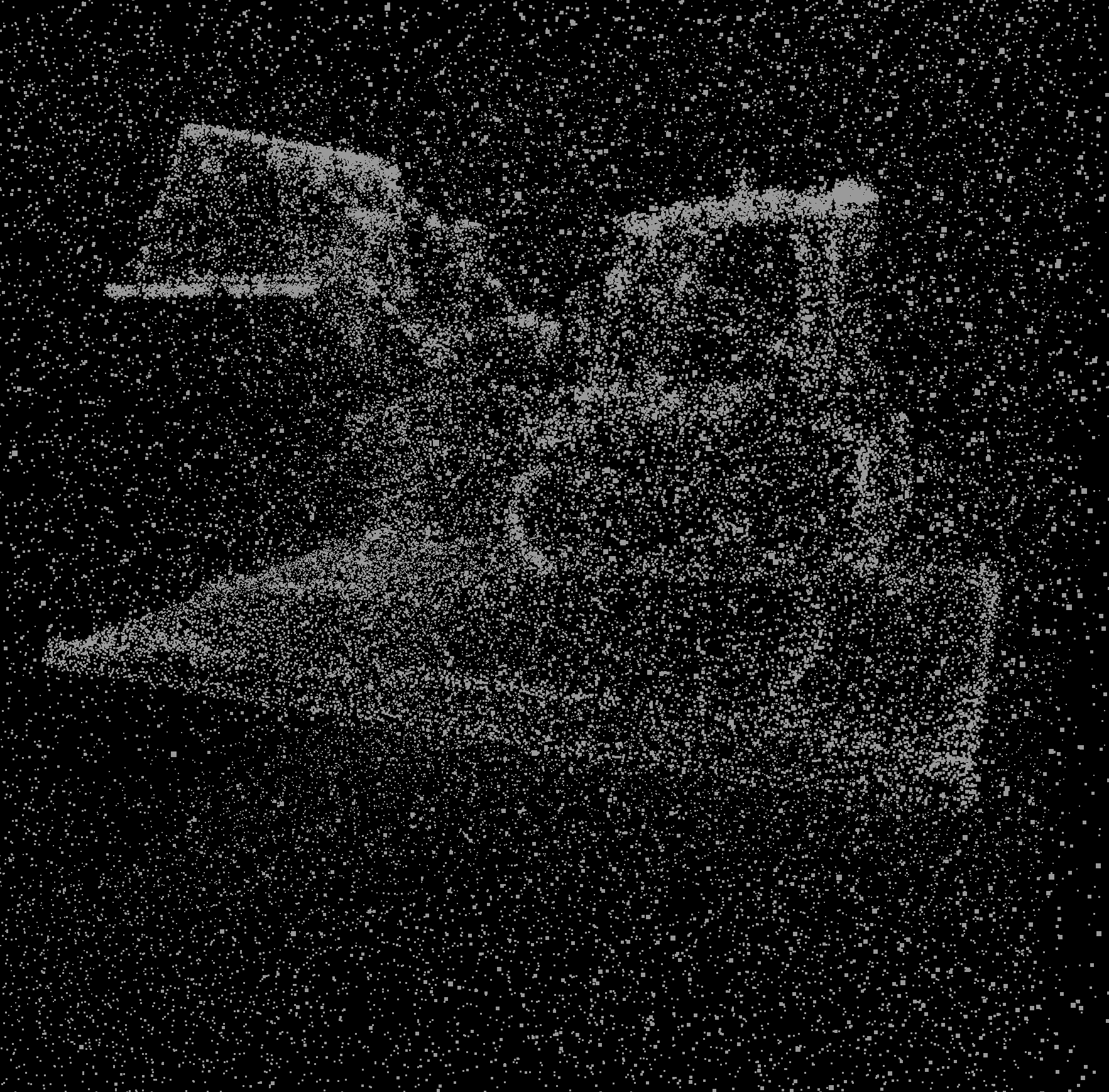} &
\includegraphics[width=0.14\textwidth, height=0.14\textwidth]{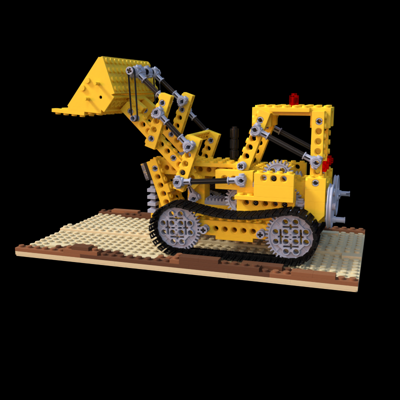} &
\includegraphics[width=0.14\textwidth, height=0.14\textwidth]{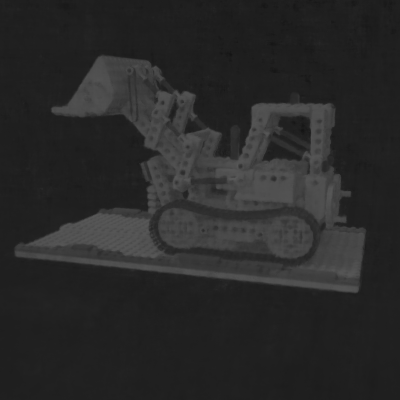} &
\includegraphics[width=0.14\textwidth, height=0.14\textwidth]{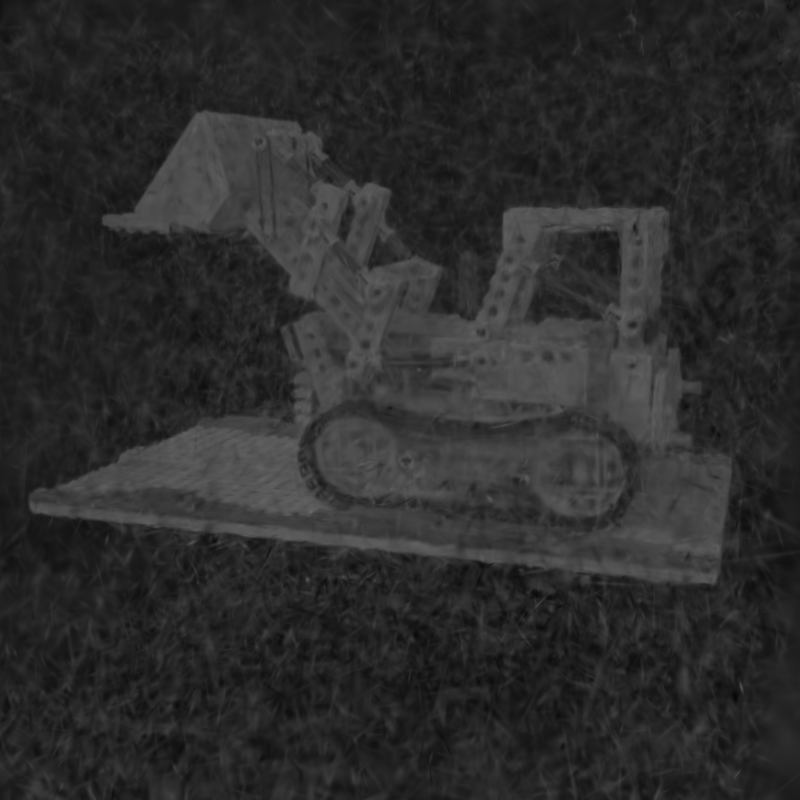} &
\includegraphics[width=0.14\textwidth, height=0.14\textwidth]{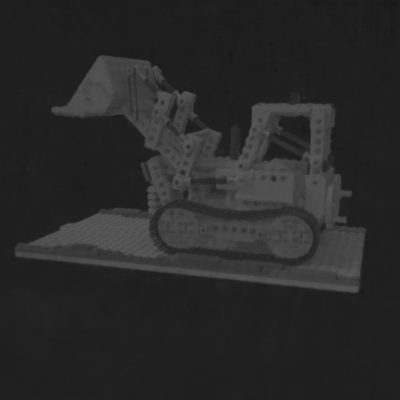} &
\includegraphics[width=0.14\textwidth, height=0.14\textwidth]{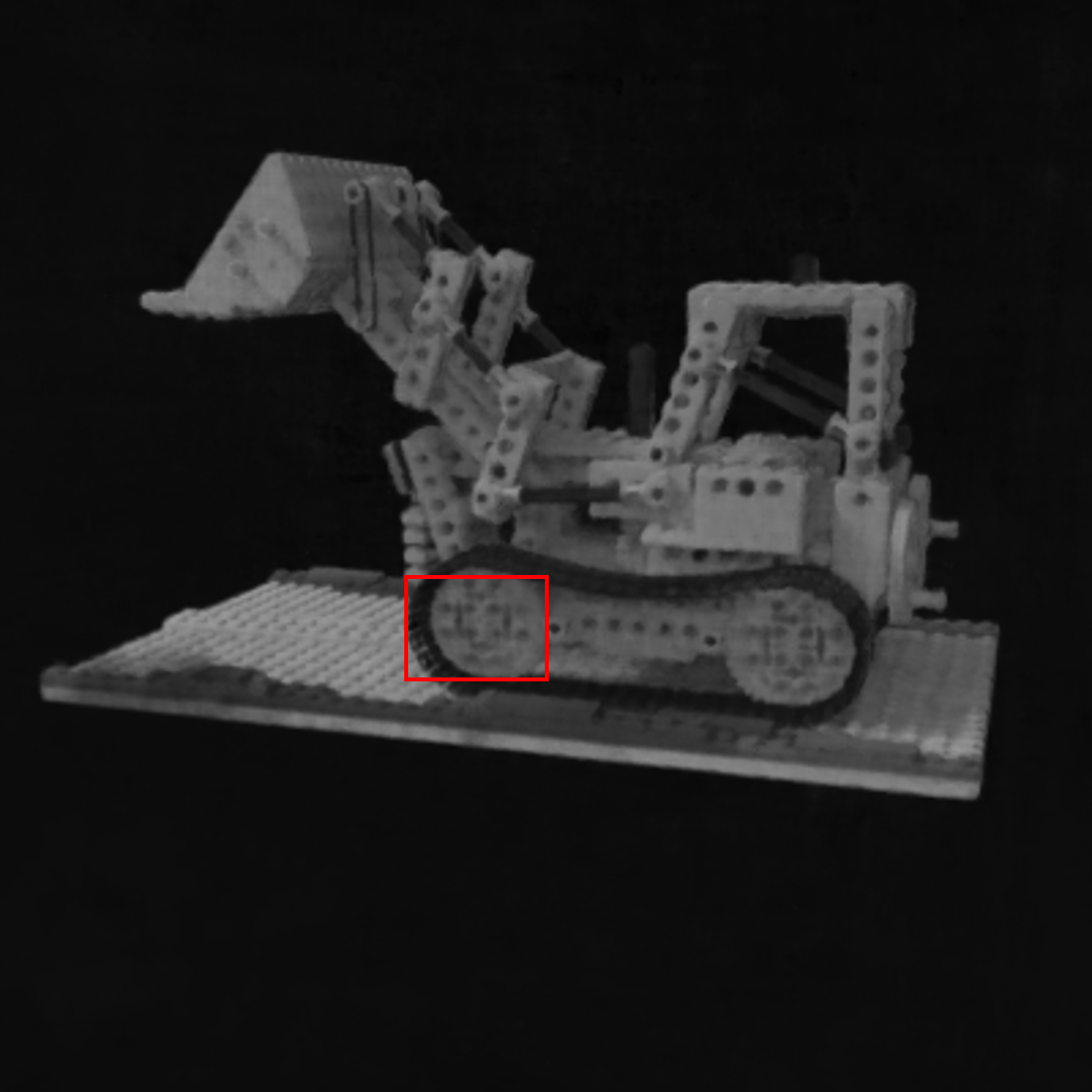} &
\includegraphics[width=0.14\textwidth, height=0.14\textwidth]{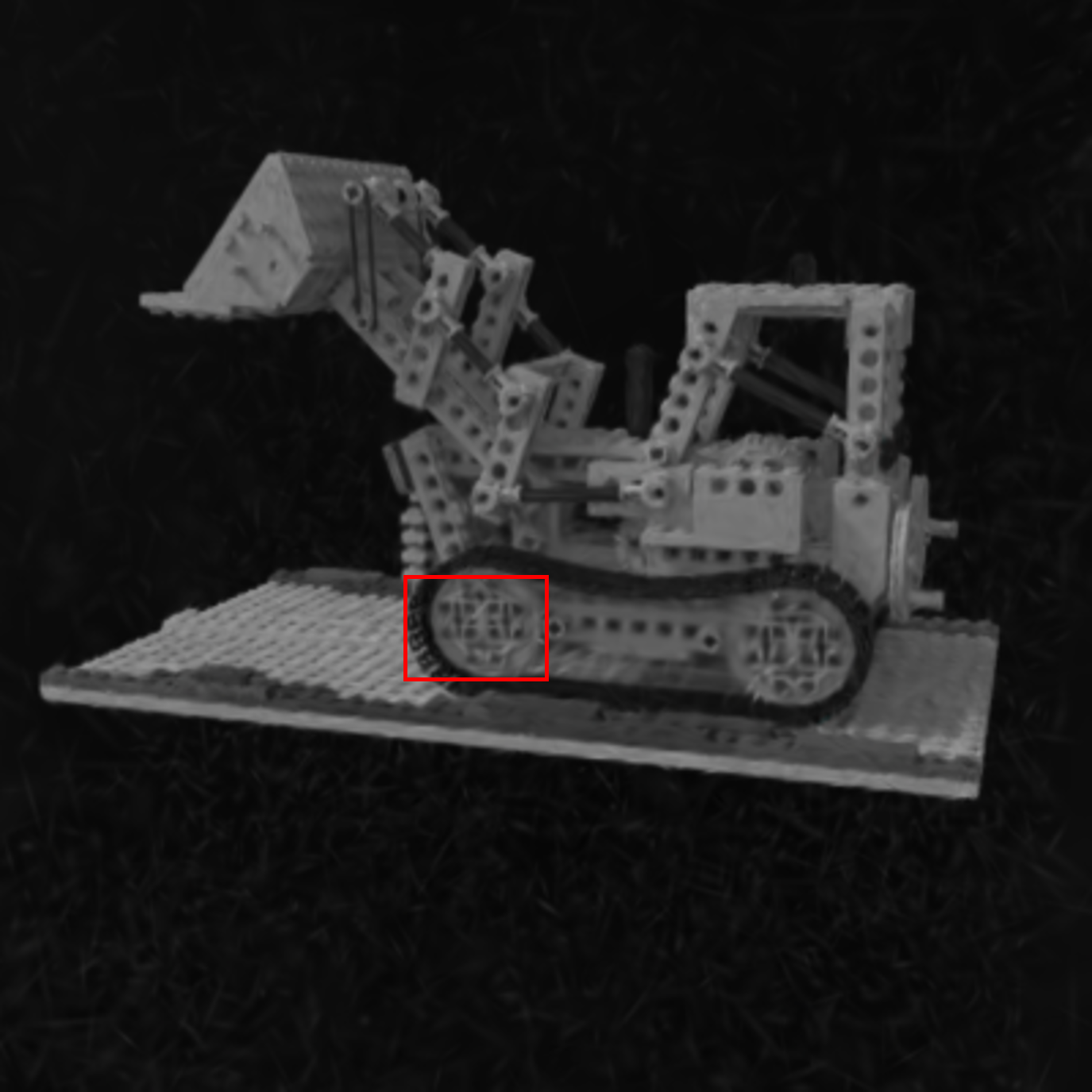}\\

\includegraphics[width=0.14\textwidth, height=0.14\textwidth]{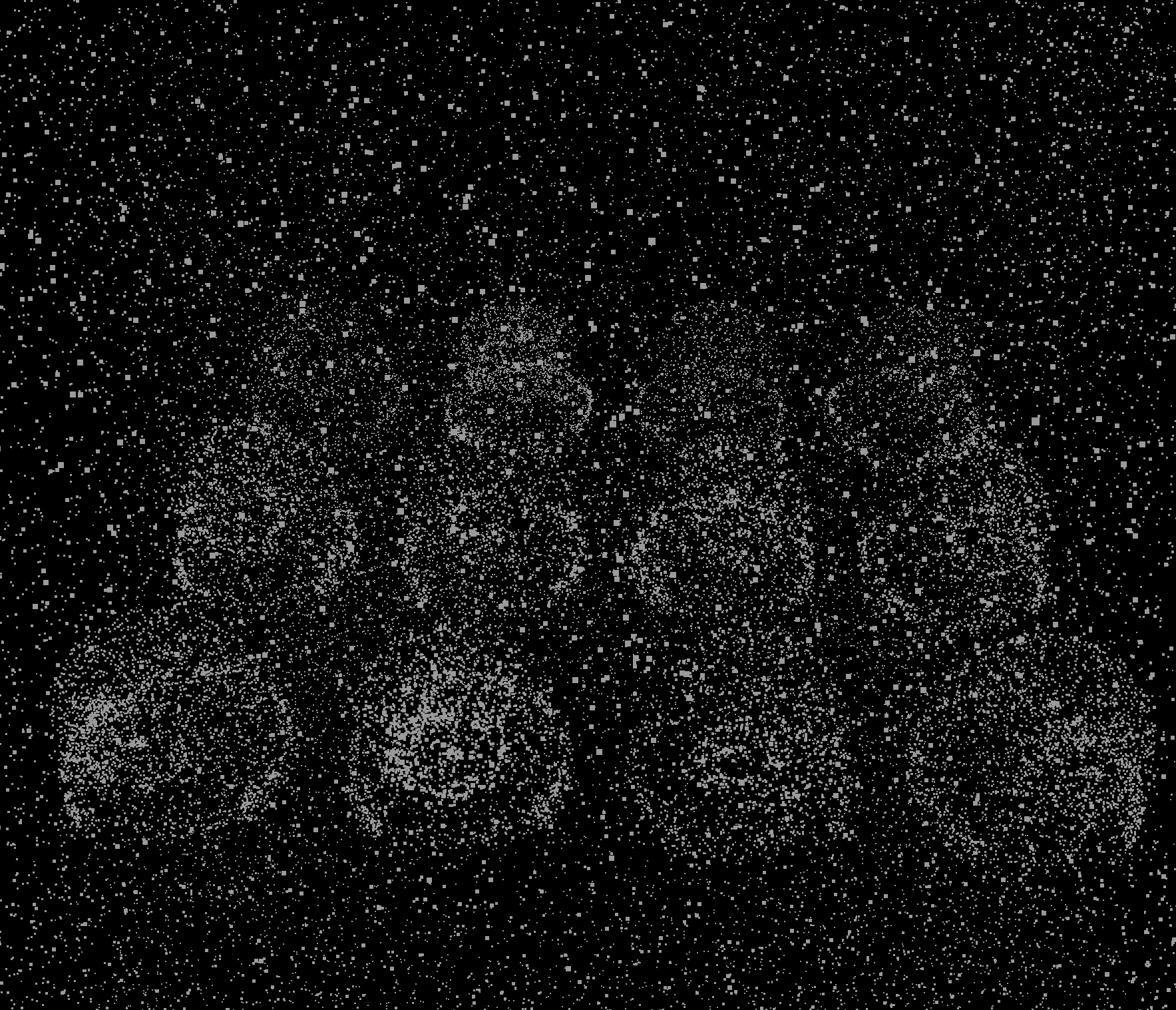} &
\includegraphics[width=0.14\textwidth, height=0.14\textwidth]{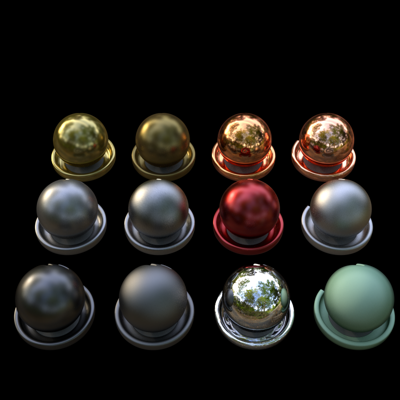} &
\includegraphics[width=0.14\textwidth, height=0.14\textwidth]{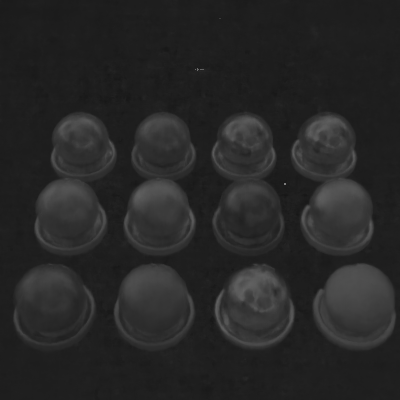} &
\includegraphics[width=0.14\textwidth, height=0.14\textwidth]{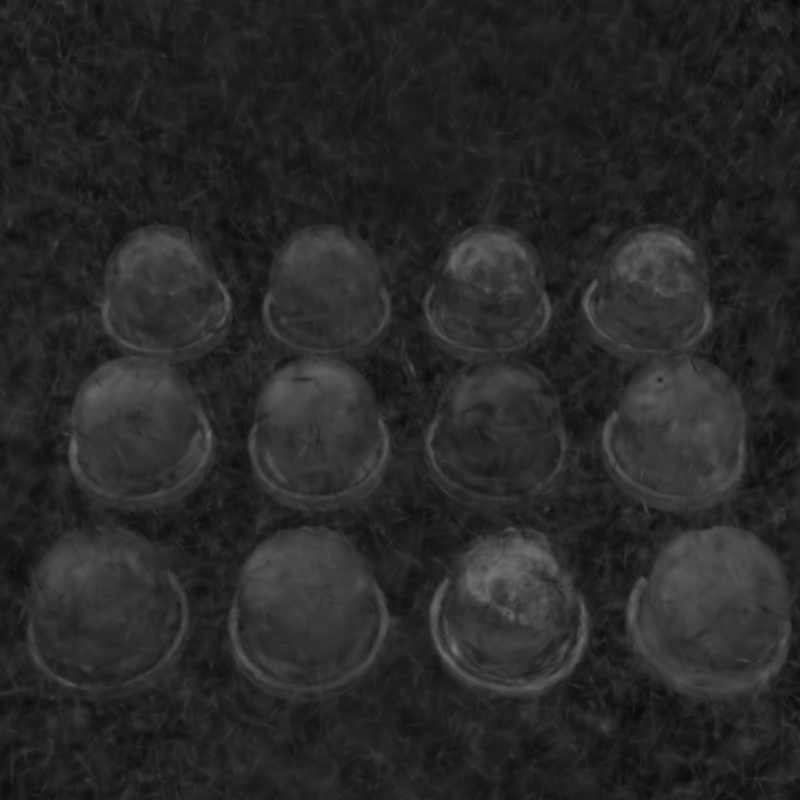} &
\includegraphics[width=0.14\textwidth, height=0.14\textwidth]{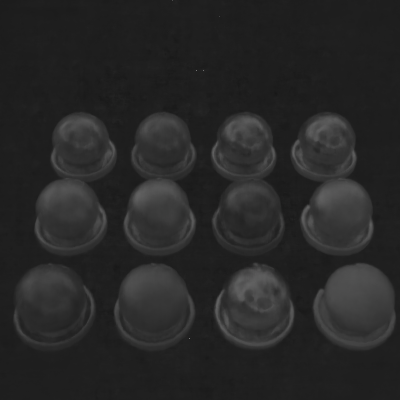} &
\includegraphics[width=0.14\textwidth, height=0.14\textwidth]{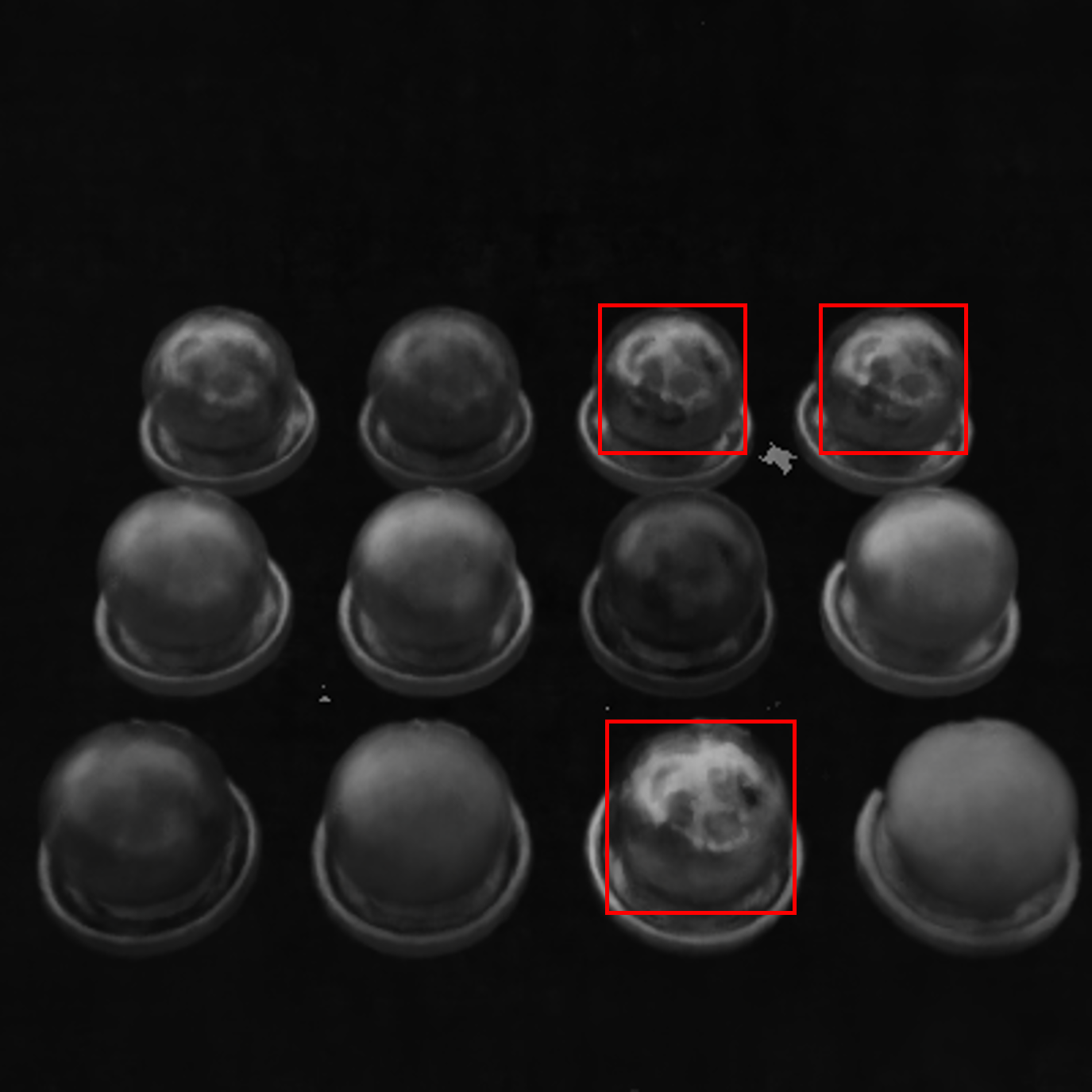} &
\includegraphics[width=0.14\textwidth, height=0.14\textwidth]{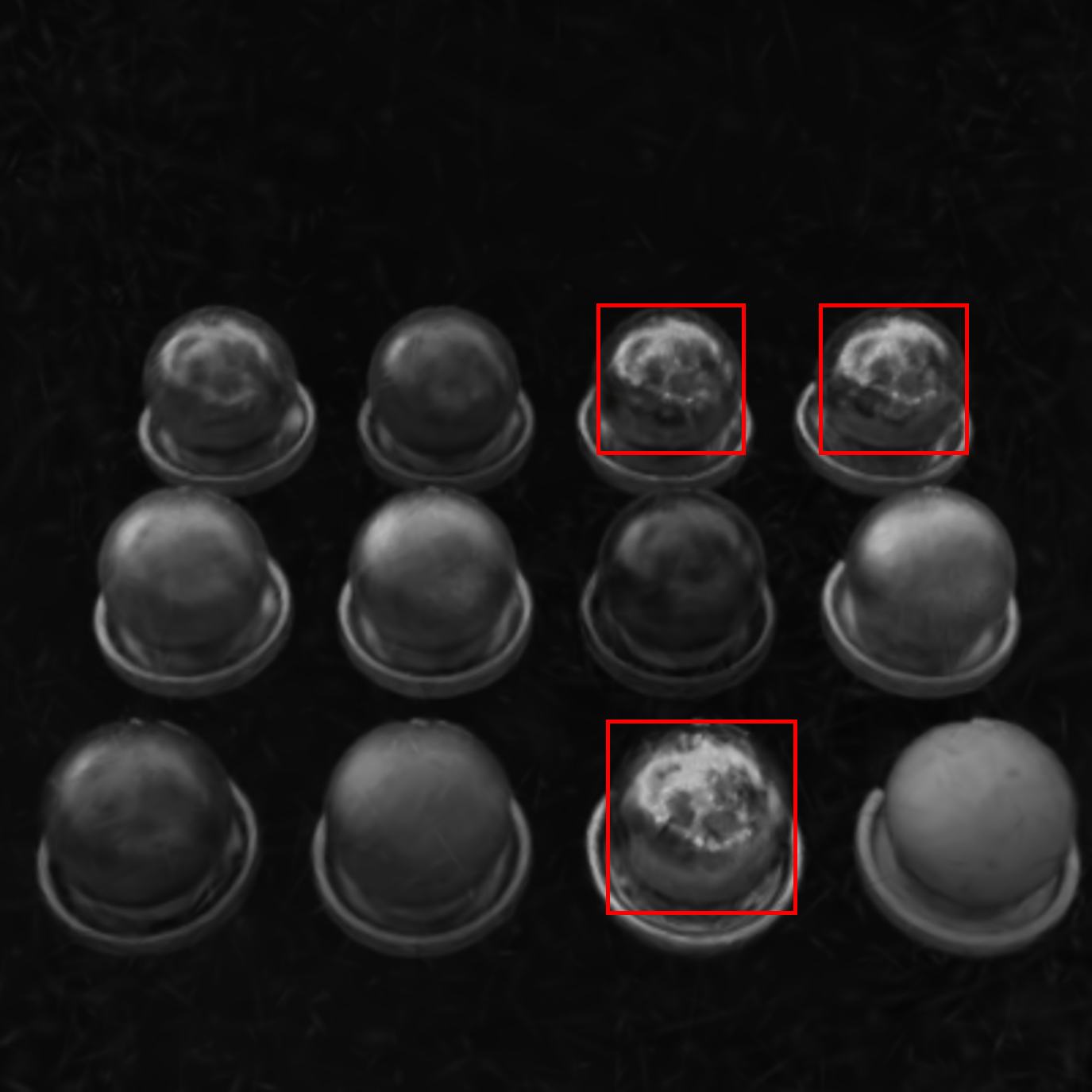}\\

\includegraphics[width=0.14\textwidth, height=0.14\textwidth]{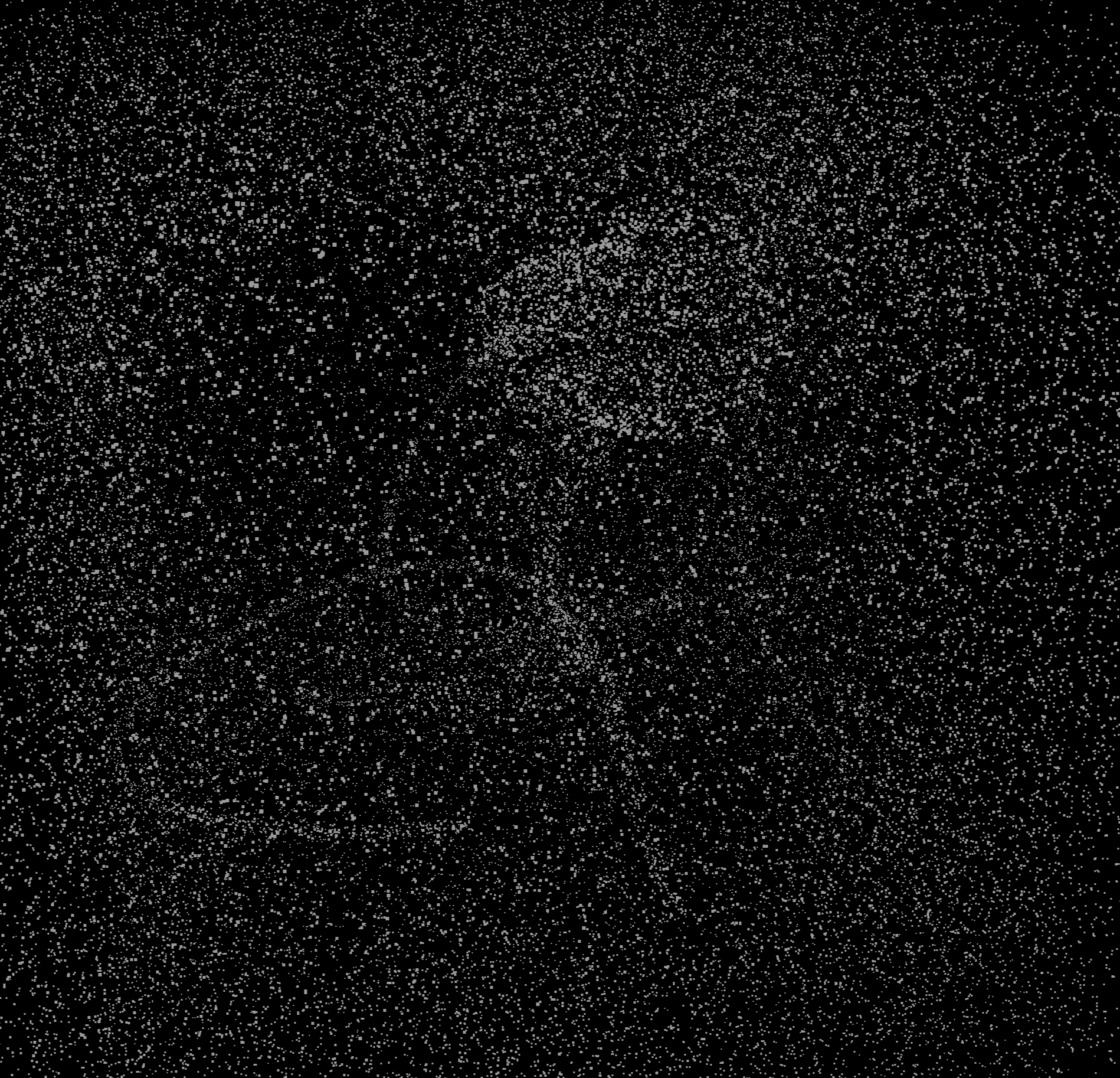} &
\includegraphics[width=0.14\textwidth, height=0.14\textwidth]{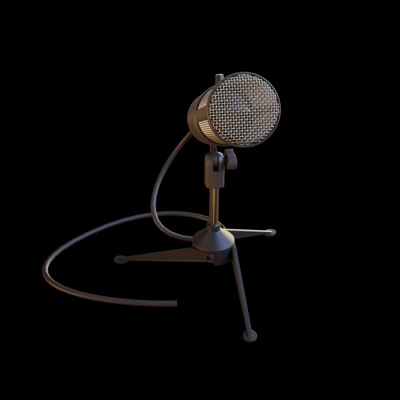} &
\includegraphics[width=0.14\textwidth, height=0.14\textwidth]{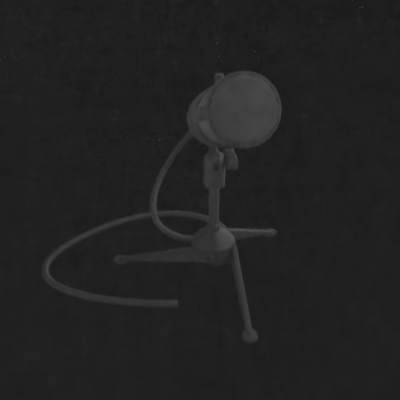} &
\includegraphics[width=0.14\textwidth, height=0.14\textwidth]{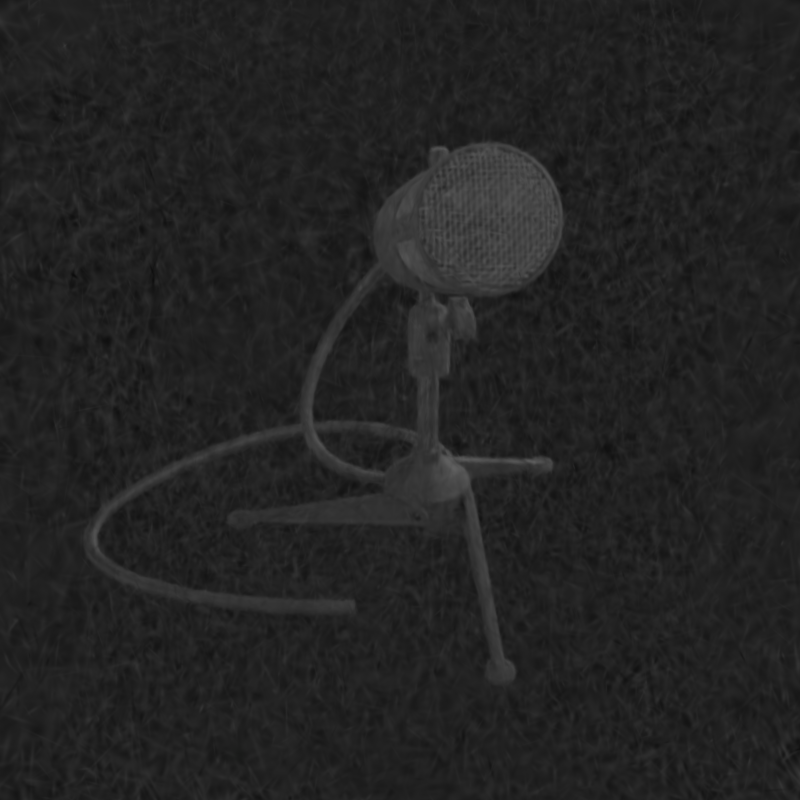} &
\includegraphics[width=0.14\textwidth, height=0.14\textwidth]{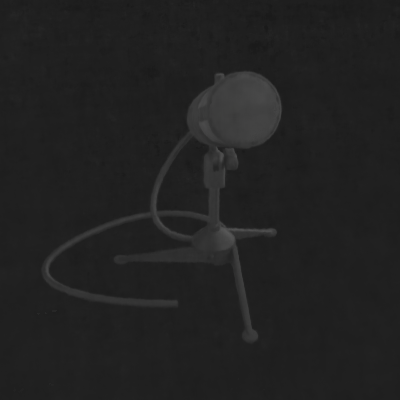} &
\includegraphics[width=0.14\textwidth, height=0.14\textwidth]{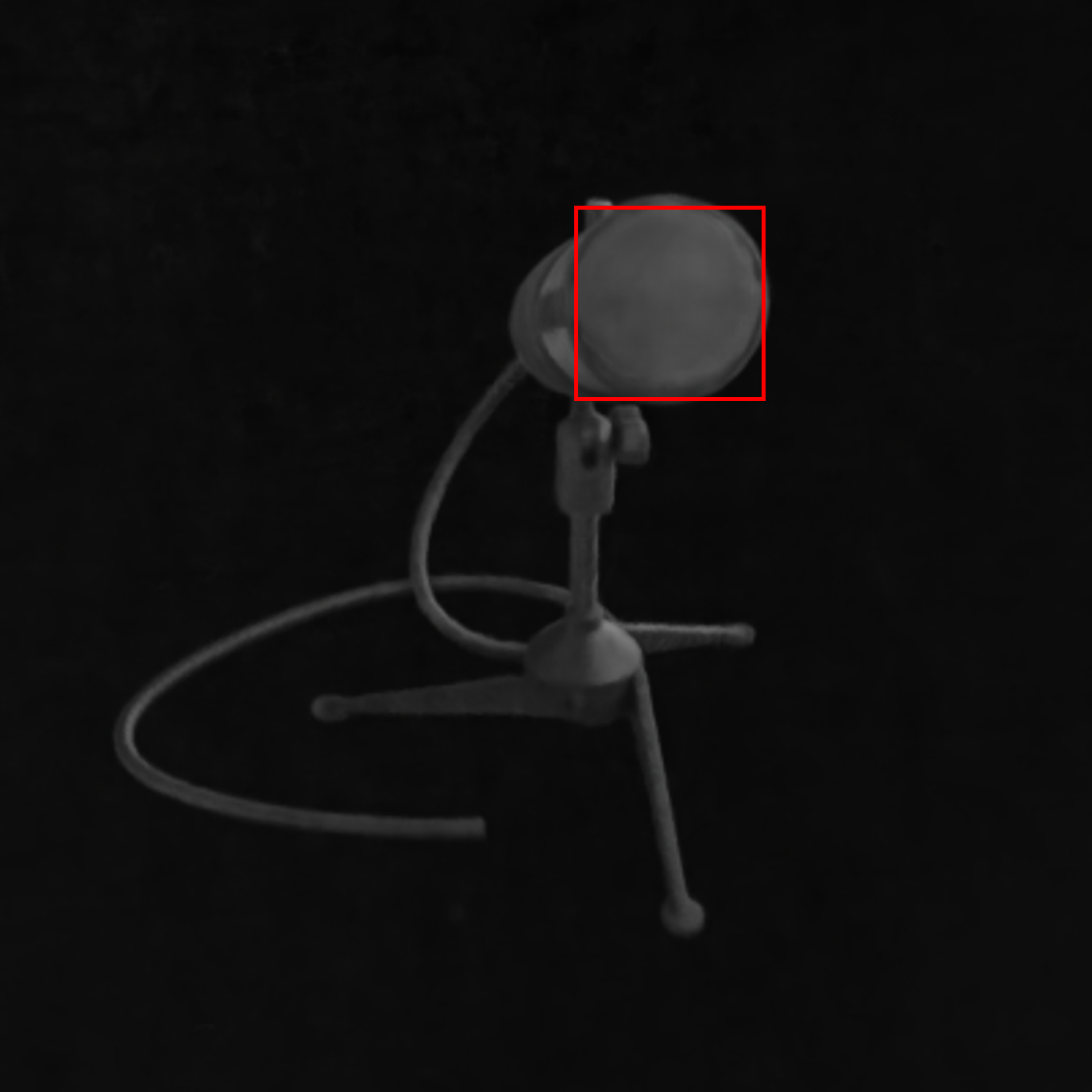} &
\includegraphics[width=0.14\textwidth, height=0.14\textwidth]{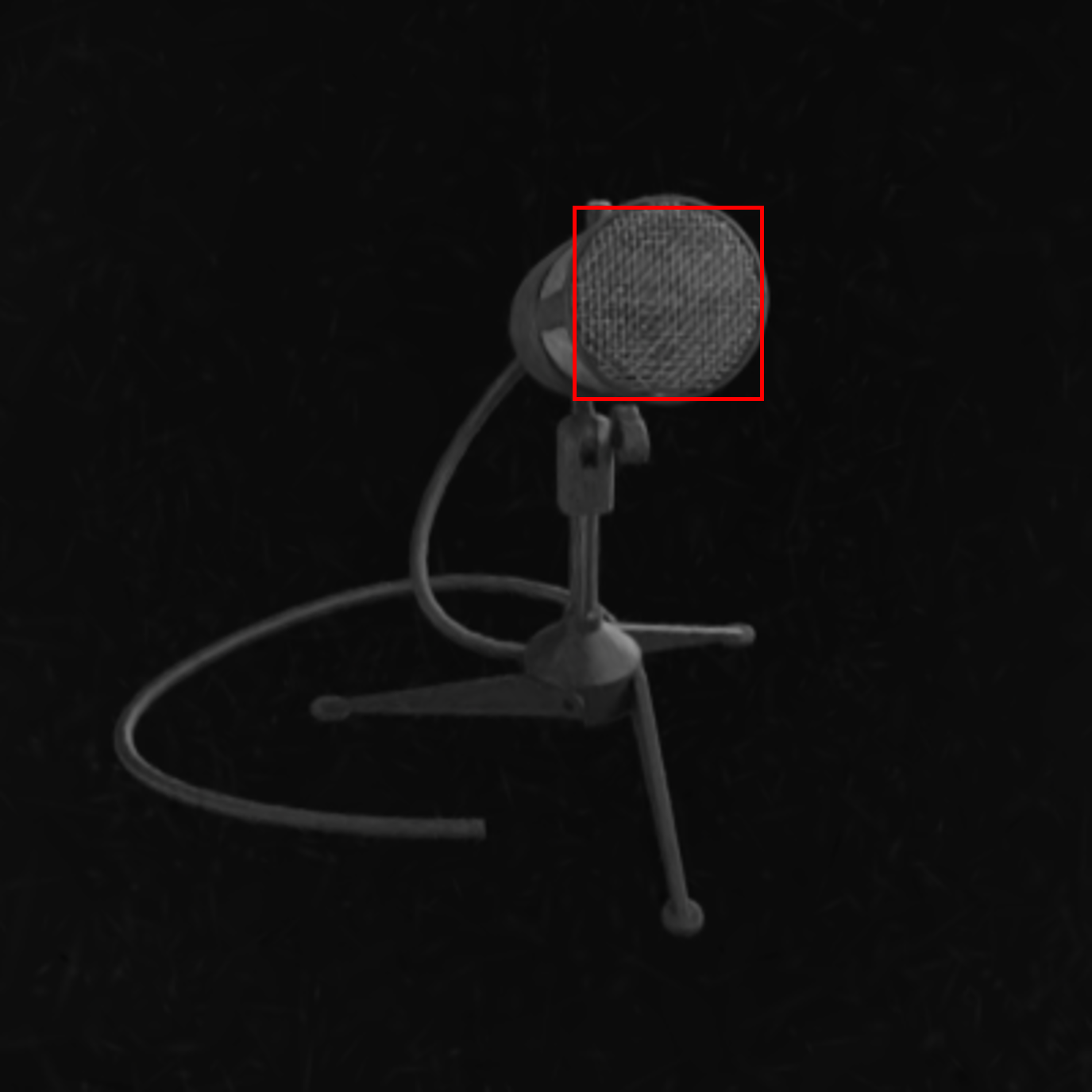}\\

\tiny {Spike} & \tiny {GT} & \tiny {Spk2img+NeRF} & \tiny {Spk2img+GS}  & \tiny {Spike-NeRF}\cite{guo2024spike} & \tiny {SpikeNeRF}\cite{zhu2024spikenerf} & \tiny {Ours}\\
\end{tabular*}
\end{minipage}
\caption{Qualitative comparison of our model with other methods on the synthetic dataset. The scenes depicted above, from top to bottom, are "chair", "ficus", "hotdog", "lego", "materials" and "mic".}
\label{fig:qualitative_synthetic}
\end{figure*}
\begin{figure}[!htb]
\centering
\begin{minipage}{\textwidth}
\begin{tabular*}{\textwidth}{cccccc}
  \includegraphics[width=\textwidth]{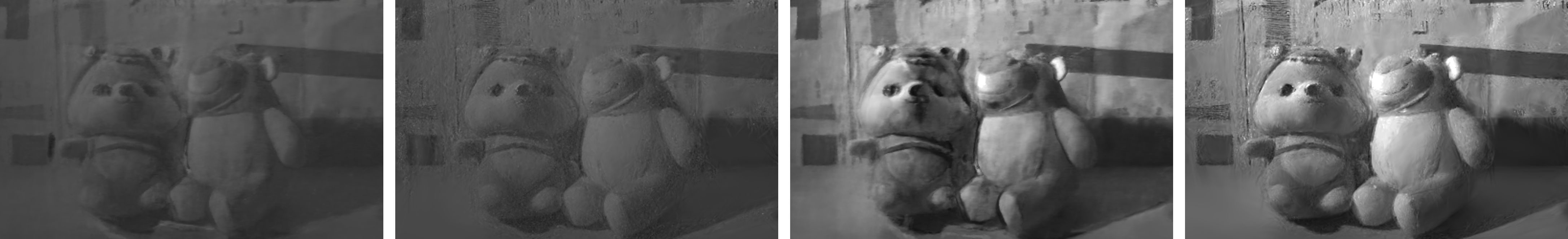}\\
  \includegraphics[width=\textwidth]{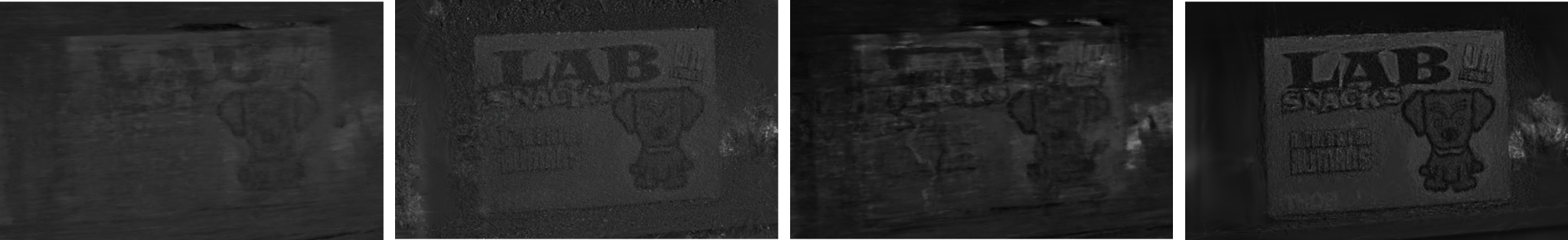}\\
  \includegraphics[width=\textwidth]{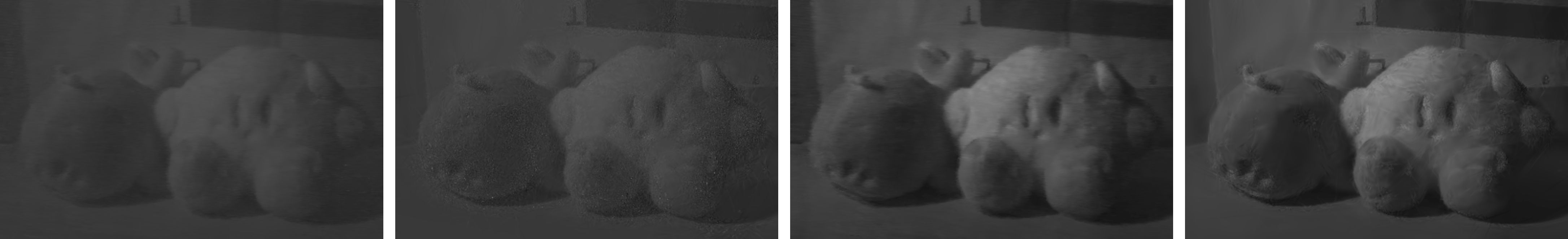}\\
  \includegraphics[width=\textwidth]{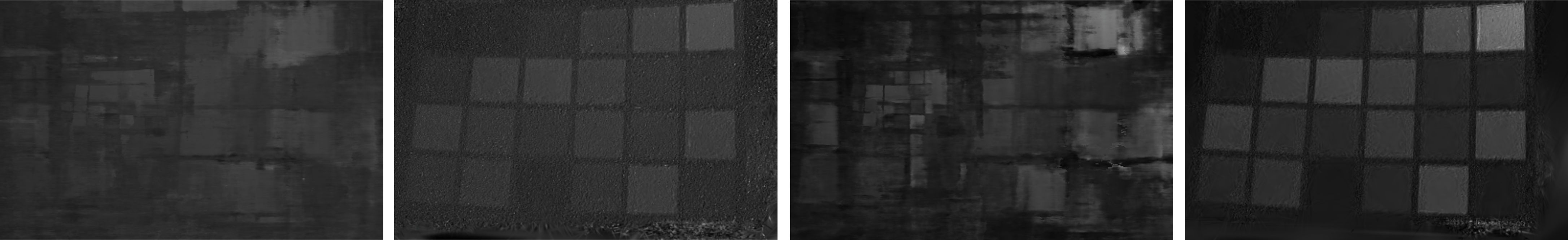}\\
  \tiny {\hspace{2em}Spk2img+NeRF\hspace{6.5em}Spk2img+GS\hspace{7em}SpikeNeRF\cite{zhu2024spikenerf}\hspace{7.5em}SpikeGS\hspace{4em}}
  \end{tabular*}
\end{minipage}
  \caption{Qualitative comparison of our model with other methods on the real datasets under different lighting conditions. The scenes depicted above, from top to bottom, are "dolls(high light intensity)", "box"(lowest illumination), "toys"(moderate illumination), and "grid"(moderate illumination) .}
  \label{fig:qualitative_real}
\end{figure}
\subsection{Experimental Settings}
\textbf{Benchmark Datasets.}
We used the spike dataset provided by SpikeNeRF\cite{zhu2024spikenerf} to evaluate our model. The synthetic dataset includes six scenes (chair, ficus, hotdog, lego, materials and mic), with each scene comprising 100 images from different viewpoints. This synthetic dataset is generated based on NeRF's synthetic dataset using the spike generator provided in \cite{zhu2023recurrent}. The original size of the synthetic dataset is 800x800.
The real-world dataset includes four scenes (dolls, box, toys, grid), each with 35 images from different viewpoints. This dataset is recorded using a handheld spike camera, capable of capturing spike stream at a spatial resolution of 250x400 and a temporal resolution of 20 KHz.

\noindent \textbf{Baselines.}
Due to the relative lack of methods for novel view synthesis based on spike camera, we compared our approach with four spike cameras image reconstruction methods: Spk2img+NeRF, Spk2img+GS, SpikeNeRF \cite{zhu2024spikenerf} and Spike-NeRF \cite{guo2024spike} (This paper is not open-source, therefore, we manually reproduced the method. Since the \cite{guo2024spike} paper does not include experiments on real-world datasets, we do not test \cite{guo2024spike} on real datasets for the sake of fairness). Spk2img (Spk2imgNet) \cite{zhao2021spk2imgnet} is a neural network-based learning method capable of recovering images from event streams. We combined Spk2imgNet with NeRF and 3DGS model. Spk2imgNet first recovers multi-view images from the event stream, and then these images are input into the NeRF model and the 3DGS model for novel view synthesis and comparison.

\noindent \textbf{Training Details.}
For both synthetic and real datasets, the inputs consist of the corresponding spike stream and poses. Since 3D Gaussian inputs require an initial point cloud, we randomly generate the initial point cloud for the synthetic dataset. For the real dataset, we use the sparse point cloud (points3D) provided by SpikeNeRF and exported by COLMAP. Our experiments were conducted on a single NVIDIA 4090, with 20K iterations for Spike2img+GS and our model, and 200K iterations for Spk2img+NeRF, SpikeNeRF and Spike-NeRF. Notably, our framework is approximately 15 times faster than SpikeNeRF, and its memory consumption is only half that of NeRF-based methods.

\noindent \textbf{Experimental Evaluation Metrics.}
In the synthetic dataset, we use PSNR, SSIM \cite{wang2004image}, and LPIPS \cite{zhang2018unreasonable} as our experimental evaluation metrics to quantify the distance between the synthesized novel views and the ground truth RGB images. For the real dataset, which lacks corresponding ground truth images, we employ NIQE \cite{niqe-python} \cite{mittal2012making}  and BRISQUE \cite{BRISQUE} \cite{mittal2012no} as no-reference image quality evaluation metrics. Below, we present the quantitative and qualitative comparisons of our model on both synthetic and real datasets. In the table, each color shading indicates the \colorbox{red!25}{best} and \colorbox{orange!25}{second-best} result.

\subsection{Quantitative and Qualitative Results}
\begin{table}[t]
	\centering
	\caption{Quantitative comparison for novel view synthesis on the real dataset. 
 }
	\setlength\tabcolsep{6pt}{
		\resizebox{0.65\linewidth}{!}{
		\begin{tabular}{c|c|c|c}
			\hline
			Method & \multicolumn{1}{c|}{Brisque$\downarrow$} & \multicolumn{1}{c|}{NIQE$\downarrow$} & \multicolumn{1}{c}{Time$\downarrow$}\\
			
			\specialrule{0.05em}{1pt}{1pt}
			Spk2img+NeRF(200K) & 47.45 & 25.08  & >3 hours \\
			\hline
			Spk2img+GS(30K) & \cellcolor{orange!25}19.59 & \cellcolor{orange!25}16.91 & $\sim$~3 mins \\
			\hline
			SpikeNeRF\cite{zhu2024spikenerf}(200K) & 39.91 & 27.22 & >8 hours\\
			\hline
			Ours N = 128(30K) & 23.39 & \cellcolor{red!25}16.15 & $\sim$~20 mins  \\

            Ours N = 512(30K) & 20.40 &  20.41 & >1 hours \\

            Ours N = 256(30K) & \cellcolor{red!25}13.11 & 19.04 & $\sim$~30mins  \\
			\hline
		\end{tabular}
	}
	}
 \label{tab:spike_llff}
\end{table}
Table~\ref{tab:spike_llff} and ~\ref{tab:spike_synthetic} present the quantitative comparison results for the synthetic and real-world data, respectively (Note that since Spike-NeRF \cite{guo2024spike} is not suitable for extremely noisy spike data and does not incorporate noise embedding or spiking neurons, its final performance is only comparable to Spike2img+NeRF). We tested our model's performance with different window sizes~(128, 256, 512). Our model outperforms existing methods across almost all metrics in each scene, meanwhile also being faster compared to NeRF-based methods. As illustrated in the last column of the two tables, our method requires only around 40 minutes(synthetic dataset) or 30 minutes(real dataset), whereas NeRF-based methods typically take over ten hours(synthetic dataset) or eight hours(real dataset). Furthermore, our method consumes only half the memory of NeRF-based methods, and our rendering speed can reach 100 FPS, whereas NeRF achieves less than 10 FPS. It is important to note that the computation time for Spk2img+NeRF and Spk2img+GS does not include the time spent on image reconstruction by the Spk2imgNet network, only the training time for NeRF and GS is considered.
\begin{table*}[t]
	\centering
	\caption{Quantitative comparison for novel view synthesis on the synthetic dataset. 
 }
	\resizebox{\linewidth}{!}{
 \renewcommand\arraystretch{2}
        \large
		\begin{tabular}{c|ccc|ccc|ccc|ccc|ccc|ccc|c}
			\hline
			\multicolumn{1}{c|}{\Large \textbf{\multirow{2}{*}{Method}}} & \multicolumn{3}{c|}{\Large \textbf{Chair}}& \multicolumn{3}{c|}{\Large \textbf{Ficus}} &  \multicolumn{3}{c|}{\Large \textbf{Hotdog}} & \multicolumn{3}{c|}{\Large \textbf{Lego}} & \multicolumn{3}{c|}{\Large \textbf{Materials}} & \multicolumn{3}{c|}{\Large \textbf{Mic}} & \multicolumn{1}{c}{\Large \textbf{\multirow{2}{*}{Time$\downarrow$}}}\\
			& { \textbf{PSNR$\uparrow$}} & { \textbf{SSIM$\uparrow$}} & { \textbf{LPIPS$\downarrow$}} & { \textbf{PSNR$\uparrow$}} & { \textbf{SSIM$\uparrow$}} & { \textbf{LPIPS$\downarrow$}} & { \textbf{PSNR$\uparrow$}} & { \textbf{SSIM$\uparrow$}} & { \textbf{LPIPS$\downarrow$}}  & { \textbf{PSNR$\uparrow$}} & { \textbf{SSIM$\uparrow$}} & { \textbf{LPIPS$\downarrow$}}  & { \textbf{PSNR$\uparrow$}} & { \textbf{SSIM$\uparrow$}} & { \textbf{LPIPS$\downarrow$}} &
            { \textbf{PSNR$\uparrow$}} & { \textbf{SSIM$\uparrow$}} & { \textbf{LPIPS$\downarrow$}} \\
			\hline
			{\Large \textbf{Spk2img+NeRF(200K)}} & {\Large \textbf{14.47}} & {\Large \textbf{.0876}} & {\Large \textbf{.2242}} & {\Large \textbf{17.21}} & {\Large \textbf{.0487}} & {\Large \textbf{.1964}} & {\Large \textbf{15.48}} & {\Large \textbf{.1702}} & {\Large \textbf{.2351}} & {\Large \textbf{14.73}} & {\Large \textbf{.1188}} & {\Large \textbf{.3057}} & {\Large \textbf{16.93}} & {\Large \textbf{.1386}} & {\Large \textbf{.2314}} & {\Large \textbf{17.78}} & {\Large \textbf{.0682}} & {\Large \textbf{.1826}} & {\Large \textbf{>3 hours}}\\
            \hline
			{\Large \textbf{Spk2img+GS(30K)}}  & {\Large \textbf{14.30}} & {\Large \textbf{.0870}} & {\Large \textbf{.4953}} & {\Large \textbf{16.85}} & {\Large \textbf{.0498}} & {\Large \textbf{.5345}} & {\Large \textbf{15.28}} & {\Large \textbf{.1636}} & {\Large \textbf{.5015}} & {\Large \textbf{14.56}} & {\Large \textbf{.1190}} & {\Large \textbf{.5319}} & {\Large \textbf{16.51}} & {\Large \textbf{.1281}} & {\Large \textbf{.5260}} & {\Large \textbf{17.40}} & {\Large \textbf{.0644}} & {\Large \textbf{.5031}} & {\Large \textbf{$\sim$~5 mins}} \\
			\hline
            {\Large \textbf{Spike-NeRF\cite{guo2024spike}(200K)}}  & {\Large \textbf{14.44}} & {\Large \textbf{.0865}} & {\Large \textbf{.2240}} & {\Large \textbf{17.21}} & {\Large \textbf{.0486}} & {\Large \textbf{.1975}} & {\Large \textbf{15.47}} & {\Large \textbf{.1701}} & {\Large \textbf{.2346}} & {\Large \textbf{14.72}} & {\Large \textbf{.1186}} & {\Large \textbf{.3043}} & {\Large \textbf{16.92}} & {\Large \textbf{.1386}} & {\Large \textbf{.2247}} & {\Large \textbf{17.79}} & {\Large \textbf{.0677}} & {\Large \textbf{.1780}} & {\Large \textbf{>3 hours}} \\
			\hline
			{\Large \textbf{SpikeNeRF\cite{zhu2024spikenerf}(200K)}}  & {\Large \cellcolor{orange!25}\textbf{20.06}} & {\Large \textbf{.1871}} & {\Large \cellcolor{orange!25}\textbf{.1271}} & {\Large \cellcolor{orange!25}\textbf{21.65}} & {\Large \textbf{.1081}} & {\Large \cellcolor{orange!25}\textbf{.1649}} & {\Large \cellcolor{orange!25}\textbf{19.94}} & {\Large \textbf{.2530}} & {\Large \cellcolor{red!25}\textbf{.1393}} & {\Large \cellcolor{orange!25}\textbf{18.62}} & {\Large \textbf{.2247}} & {\Large \cellcolor{red!25}\textbf{.1987}} & {\Large \cellcolor{orange!25}\textbf{21.84}} & {\Large \textbf{.2319}} & {\Large \cellcolor{orange!25}\textbf{.1396}} & {\Large \cellcolor{orange!25}\textbf{23.62}} & {\Large \textbf{.1299}} & {\Large \cellcolor{red!25}\textbf{.1235}} & {\Large \textbf{>10 hours}}\\
            \hline
            {\Large \textbf{Ours N = 128(30K)}} & {\Large \textbf{14.13}} & {\Large \textbf{.1526}} & {\Large \textbf{.3427}} & {\Large \textbf{14.22}} & {\Large \textbf{.0656}} & {\Large \textbf{.4605}} & {\Large \textbf{15.39}} & {\Large \textbf{.1962}} & {\Large \textbf{.3996}} & {\Large \textbf{14.78}} & {\Large \textbf{.1964}} & {\Large \textbf{.4003}} & {\Large \textbf{15.73}} & {\Large \textbf{.1576}} & {\Large \textbf{.4087}} & {\Large \textbf{16.56}} & {\Large \textbf{.0778}} & {\Large \textbf{.4157}} & {\Large \textbf{$\sim$~30 mins}}\\
            {\Large \textbf{Ours N = 512(30K)}} & {\Large \textbf{14.70}} & {\Large \cellcolor{red!25}\textbf{.5317}} & {\Large \textbf{.1692}} & {\Large \textbf{21.03}} & {\Large \cellcolor{red!25}\textbf{.5388}} & {\Large \cellcolor{red!25}\textbf{.1430}} & {\Large \textbf{15.71}} & {\Large \cellcolor{red!25}\textbf{.5403}} & {\Large \textbf{.1917}} & {\Large \textbf{15.08}} & {\Large \cellcolor{red!25}\textbf{.5134}} & {\Large \textbf{.2566}} & {\Large \textbf{17.77}} & {\Large \cellcolor{red!25}\textbf{.5258}} & {\Large \textbf{.1786}} & {\Large \textbf{21.88}} & {\Large \cellcolor{red!25}\textbf{.5576}} & {\Large \textbf{.1350}} & {\Large \textbf{>2 hours}}\\
			{\Large \textbf{Ours N = 256(30K)}}  & {\Large \cellcolor{red!25}\textbf{20.24}} & {\Large \cellcolor{orange!25}\textbf{.1984}} & {\Large \cellcolor{red!25}\textbf{.1213}} & {\Large \cellcolor{red!25}\textbf{21.86}} & {\Large \cellcolor{orange!25}\textbf{.1201}} & {\Large \textbf{.1820}} & {\Large \cellcolor{red!25}\textbf{20.17}} & {\Large \cellcolor{orange!25}\textbf{.2567}} & {\Large \cellcolor{orange!25}\textbf{.1612}} & {\Large \cellcolor{red!25}\textbf{18.63}} & {\Large \cellcolor{orange!25}\textbf{.2335}} & {\Large \cellcolor{orange!25}\textbf{.2470}} & {\Large \cellcolor{red!25}\textbf{22.21}} & {\Large \cellcolor{orange!25}\textbf{.2493}} & {\Large \cellcolor{red!25}\textbf{.1335}} & {\Large \cellcolor{red!25}\textbf{24.38}} & {\Large \cellcolor{orange!25}\textbf{.1406}} & {\Large \cellcolor{orange!25}\textbf{.1397}} & {\Large \textbf{$\sim$~40mins}}\\
			\hline
		\end{tabular}
	}
	\label{tab:spike_synthetic}
\end{table*}

\begin{figure*}[tb]
\centering
\begin{minipage}{0.8\textwidth}
\begin{tabular*}{0.8\textwidth}{cccc}
\includegraphics[width=0.25\textwidth, height=0.25\textwidth]{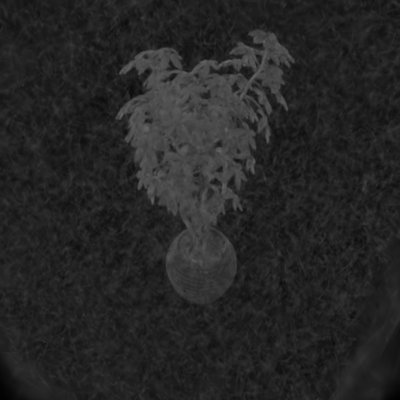} 
&\includegraphics[width=0.25\textwidth, height=0.25\textwidth]{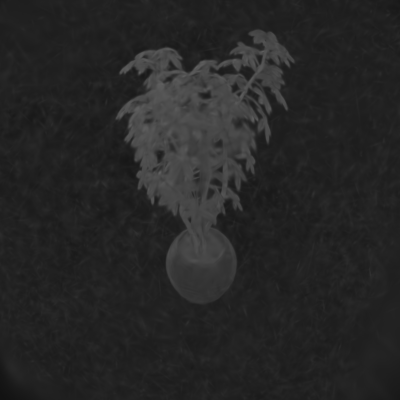} 
&\includegraphics[width=0.25\textwidth, height=0.25\textwidth]{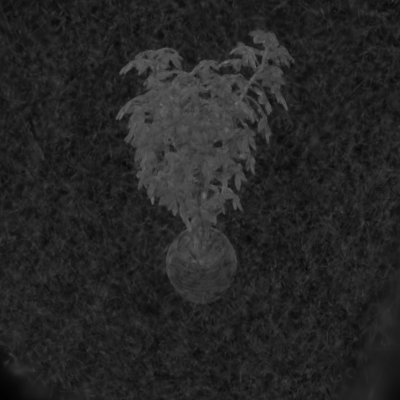} 
&\includegraphics[width=0.25\textwidth, height=0.25\textwidth]{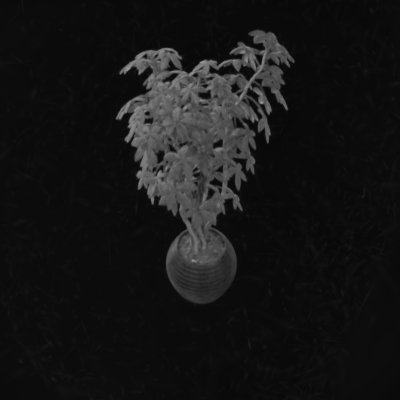}\\

\includegraphics[width=0.25\textwidth, height=0.25\textwidth]{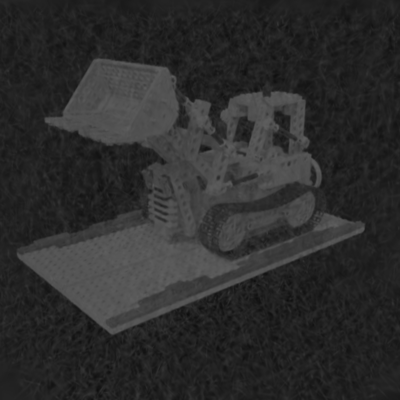} &
\includegraphics[width=0.25\textwidth, height=0.25\textwidth]{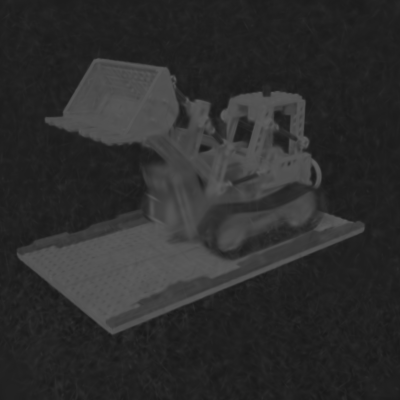} &
\includegraphics[width=0.25\textwidth, height=0.25\textwidth]{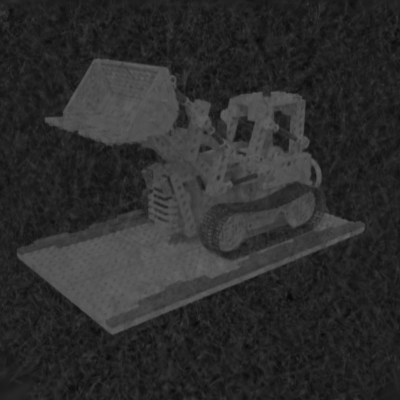} &
\includegraphics[width=0.25\textwidth, height=0.25\textwidth]{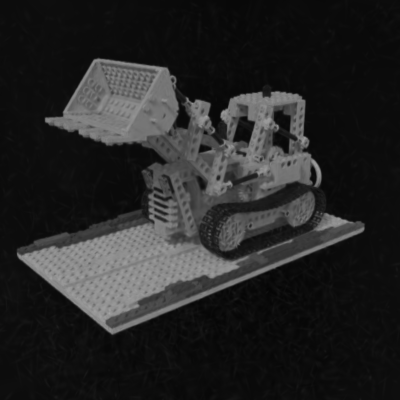}\\

\tiny {$L_{I_{in}}$} & \tiny {$L_{1}$} & \tiny {$D_{noise}$} & \tiny {$Full(L_{s})$}\\
\end{tabular*}
\end{minipage}
\caption{In the qualitative comparison of the ablation study (the full results are provided in the supplementary materials), it can be observed that the images rendered using $L_{I_{in}}$ and $D_{noise}$ contain significant noise, while the images rendered using $L_{1}$ are overly smooth with blurred details.}
\label{Ablation Study}
\end{figure*}

%


Qualitative results are demonstrated both on synthetic and real-world data. Fig.~\ref{fig:qualitative_synthetic} presents the rendered images with different methods based on synthetic data (The visualized spike images are simulated by point clouds). Our method shows significant results with noisy spike inputs. The rendered images with real data are shown in Fig.~\ref{fig:qualitative_real}. In the figure are four real scenes under different lighting conditions.
SpikeGS produces clearer images with fine textures than other methods.

\subsection{Ablation Study}
In this subsection, we perform ablation studies on each component of our framework. Specifically, we analyze the impact of our proposed 3D Gaussian-based spike generation pipeline and the spike rendering loss function. We denote the complete spike rendering loss function as $L_{s}$. $L_{I_{in}}$ represents directly using estimated light intensity and reconstructed images as supervision. $L_{1}$ denotes the supervision using only $L_{1}$ loss. We also investigate the effect of removing the noise embedding, represented by $D_{noise}$. $Full$ represents our complete model. 

\begin{table}[!htb]
	\centering
	\caption{Ablation Study for novel view synthesis on the synthetic dataset. }
	\resizebox{\linewidth}{!}{
 \renewcommand\arraystretch{2}
        \large
		\begin{tabular}{c|ccc|ccc|ccc|ccc|ccc|ccc|c|c}
            \hline
           \multicolumn{1}{c|}{\Large \textbf{\multirow{3}{*}{Method}}} & \multicolumn{18}{c|}{\Large \textbf{Synthetic dataset}} & \multicolumn{2}{c}{\Large \textbf{Real dataset}}\\
			\cline{2-21}
			& \multicolumn{3}{c|}{\Large \textbf{Chair}}& \multicolumn{3}{c|}{\Large \textbf{Ficus}} &  \multicolumn{3}{c|}{\Large \textbf{Hotdog}} & \multicolumn{3}{c|}{\Large \textbf{Lego}} & \multicolumn{3}{c|}{\Large \textbf{Materials}} & \multicolumn{3}{c|}{\Large \textbf{Mic}} 
             & \multicolumn{1}{c|}{\Large \textbf{\multirow{2}{*}{Brisque$\downarrow$}}} & \multicolumn{1}{c}{\Large \textbf{\multirow{2}{*}{NIQE$\downarrow$}}}\\
			& { \textbf{PSNR$\uparrow$}} & { \textbf{SSIM$\uparrow$}} & { \textbf{LPIPS$\downarrow$}} & { \textbf{PSNR$\uparrow$}} & { \textbf{SSIM$\uparrow$}} & { \textbf{LPIPS$\downarrow$}} & { \textbf{PSNR$\uparrow$}} & { \textbf{SSIM$\uparrow$}} & { \textbf{LPIPS$\downarrow$}}  & { \textbf{PSNR$\uparrow$}} & { \textbf{SSIM$\uparrow$}} & { \textbf{LPIPS$\downarrow$}}  & { \textbf{PSNR$\uparrow$}} & { \textbf{SSIM$\uparrow$}} & { \textbf{LPIPS$\downarrow$}} &
            { \textbf{PSNR$\uparrow$}} & { \textbf{SSIM$\uparrow$}} & { \textbf{LPIPS$\downarrow$}} &\multicolumn{1}{c|}{} & \multicolumn{1}{c}{}\\
			\hline
			{\Large \textbf{$\boldsymbol{L_{I_{in}}}$}} & {\Large \textbf{14.34}} & {\Large \cellcolor{orange!25}\textbf{.0887}} & {\Large \textbf{.4304}} & {\Large \textbf{16.90}} & {\Large \cellcolor{orange!25}\textbf{.0512}} & {\Large \textbf{.4779}} & {\Large \textbf{15.30}} & {\Large \textbf{.1668}} & {\Large \textbf{.4502}} & {\Large \textbf{14.58}} & {\Large \textbf{.1197}} & {\Large \textbf{.4850}} & {\Large \textbf{16.58}} & {\Large \textbf{.1327}} & {\Large \textbf{.4551}} & {\Large \textbf{17.43}} & {\Large \textbf{.0670}} & {\Large \textbf{.4238}} & {\Large \textbf{20.8}} & {\Large \cellcolor{orange!25}\textbf{16.72}}\\
            \hline
			{\Large \textbf{$\boldsymbol{L_{1}}$}}  & {\Large \cellcolor{orange!25}\textbf{14.37}} & {\Large \textbf{.0873}} & {\Large \cellcolor{orange!25}\textbf{.2657}} & {\Large \cellcolor{orange!25}\textbf{16.96}} & {\Large \textbf{.0498}} & {\Large \cellcolor{orange!25}\textbf{.2350}} & {\Large \cellcolor{orange!25}\textbf{15.36}} & {\Large \cellcolor{orange!25}\textbf{.1691}} & {\Large \cellcolor{orange!25}\textbf{.2874}} & {\Large \cellcolor{orange!25}\textbf{14.62}} & {\Large \textbf{.1101}} & {\Large \cellcolor{orange!25}\textbf{.3636}} & {\Large \cellcolor{orange!25}\textbf{16.71}} & {\Large \textbf{.1294}} & {\Large \cellcolor{orange!25}\textbf{.2913}} & {\Large \cellcolor{orange!25}\textbf{17.49}} & {\Large \cellcolor{orange!25}\textbf{.0674}} & {\Large \cellcolor{orange!25}\textbf{.2211}} & {\Large \textbf{68.73}} & {\Large \textbf{25.62}}\\
			\hline
			{\Large {$\boldsymbol{D_{noise}}$}}  & {\Large \textbf{14.31}} & {\Large \textbf{.0877}} & {\Large \textbf{.4903}} & {\Large \textbf{16.86}} & {\Large \textbf{.0504}} & {\Large \textbf{.5312}} & {\Large \textbf{15.28}} & {\Large \textbf{.1656}} & {\Large \textbf{.4894}} & {\Large \textbf{14.56}} & {\Large \cellcolor{orange!25}\textbf{.1198}} & {\Large \textbf{.5288}} & {\Large \textbf{16.55}} & {\Large \cellcolor{orange!25}\textbf{.1298}} & {\Large \textbf{.5161}} & {\Large \textbf{17.41}} & {\Large \textbf{.0650}} & {\Large \textbf{.4955}} & {\Large \cellcolor{orange!25}\textbf{20.63}} & {\Large \textbf{16.9}}\\
            \hline
            {\Large {$\boldsymbol{Full(L_{s})}$}} & {\Large \cellcolor{red!25}\textbf{20.24}} & {\Large \cellcolor{red!25}\textbf{.1984}} & {\Large \cellcolor{red!25}\textbf{.1213}} & {\Large \cellcolor{red!25}\textbf{21.86}} & {\Large \cellcolor{red!25}\textbf{.1201}} & {\Large \cellcolor{red!25}\textbf{.1820}} & {\Large \cellcolor{red!25}\textbf{20.17}} & {\Large \cellcolor{red!25}\textbf{.2567}} & {\Large \cellcolor{red!25}\textbf{.1612}} & {\Large \cellcolor{red!25}\textbf{18.63}} & {\Large \cellcolor{red!25}\textbf{.2335}} & {\Large \cellcolor{red!25}\textbf{.2470}} & {\Large \cellcolor{red!25}\textbf{22.21}} & {\Large \cellcolor{red!25}\textbf{.2493}} & {\Large \cellcolor{red!25}\textbf{.1335}} & {\Large \cellcolor{red!25}\textbf{24.38}} & {\Large \cellcolor{red!25}\textbf{.1406}} & {\Large \cellcolor{red!25}\textbf{.1397}} & {\Large \cellcolor{red!25}\textbf{13.11}} & {\Large \cellcolor{red!25}\textbf{16.15}}\\
			\hline
		\end{tabular}
  
	}
    
	\label{tab:Ablation Study blender}
\end{table}

\section{Conclusion}
This paper introduces SpikeGS, the work to learn 3D Gaussian fields solely from spike stream. we propose a novel rendering framework based on spike stream. We model the generation process of spike stream using 3DGS and embed the spike generation pipeline into the differentiable rasterization process of 3DGS, deriving the backpropagation accordingly. Additionally, we introduce a novel loss function for spike stream. Our model can recover clear novel views with fine details from extremely noisy spike stream under low-quality lighting conditions, using only the spike stream as supervision. We demonstrate the effectiveness of our approach on both synthetic and real datasets. 
\newpage
\clearpage

\bibliographystyle{splncs04}
\input{./supp}

\end{document}

%% file: supp.tex
\title{SpikeGS: Learning 3D Gaussian Fields from Continuous Spike Stream}
\subtitle{Supplementary Material}
\titlerunning{SpikeGS}
\authorrunning{Yu et al.}

\author{}


\institute{}

\maketitle

\section{Different lighting intensity experiments}
\begin{figure*}[!t]
\centering
\begin{minipage}{\textwidth}
\begin{tabular*}{\textwidth}{ccccc}
\includegraphics[width=0.19\textwidth, height=0.19\textwidth]{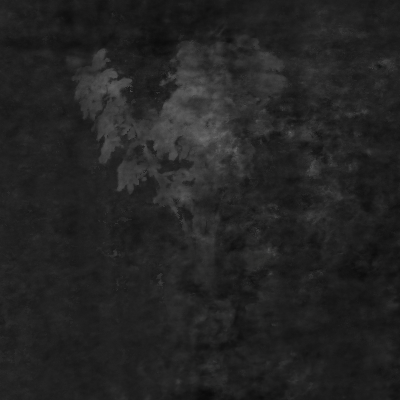} 
&\includegraphics[width=0.19\textwidth, height=0.19\textwidth]{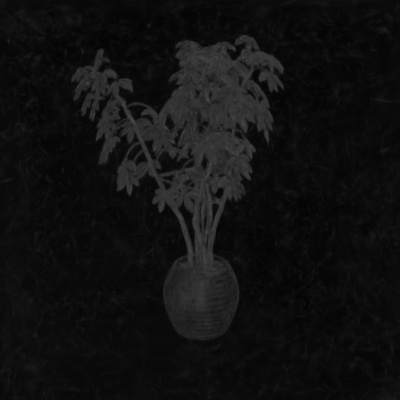} 
&\includegraphics[width=0.19\textwidth, height=0.19\textwidth]{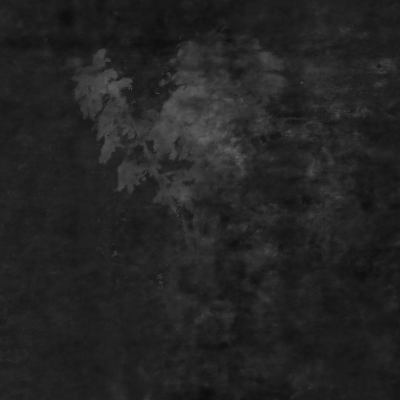} 
&\includegraphics[width=0.19\textwidth, height=0.19\textwidth]{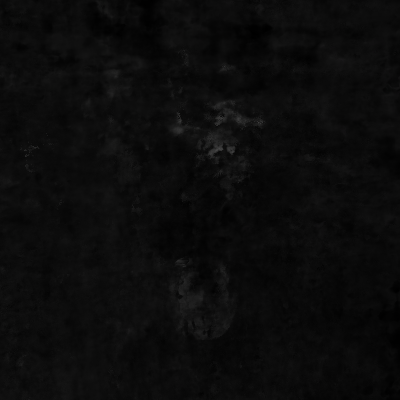}
&\includegraphics[width=0.19\textwidth, height=0.19\textwidth]{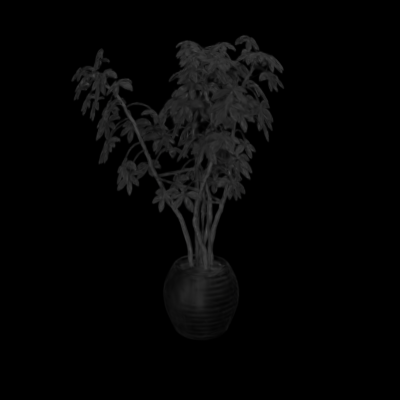}
\\

\includegraphics[width=0.19\textwidth, height=0.19\textwidth]{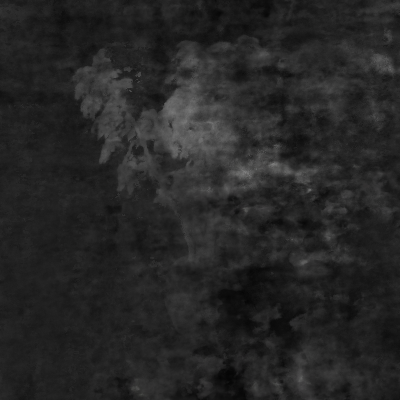} &
\includegraphics[width=0.19\textwidth, height=0.19\textwidth]{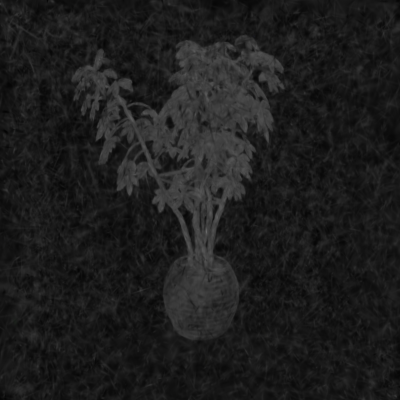} &
\includegraphics[width=0.19\textwidth, height=0.19\textwidth]{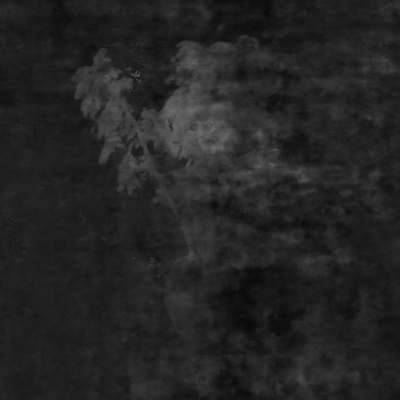} &
\includegraphics[width=0.19\textwidth, height=0.19\textwidth]{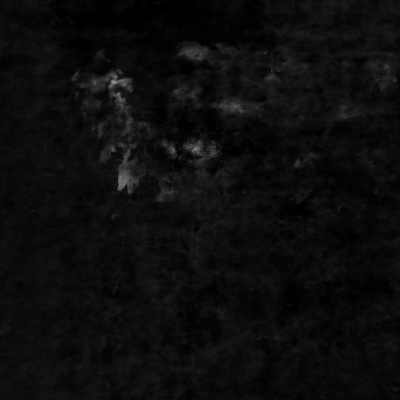}
&
\includegraphics[width=0.19\textwidth, height=0.19\textwidth]{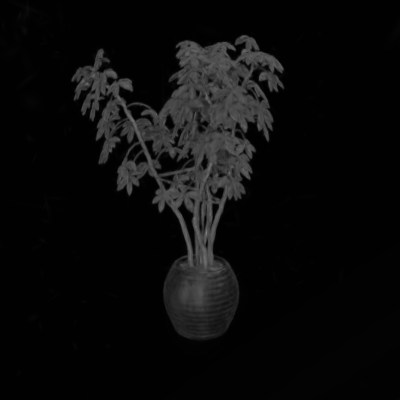}
\\

\includegraphics[width=0.19\textwidth, height=0.19\textwidth]{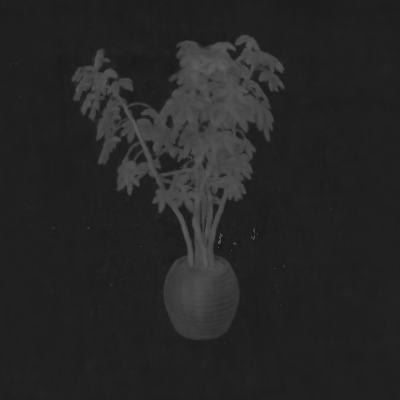} &
\includegraphics[width=0.19\textwidth, height=0.19\textwidth]{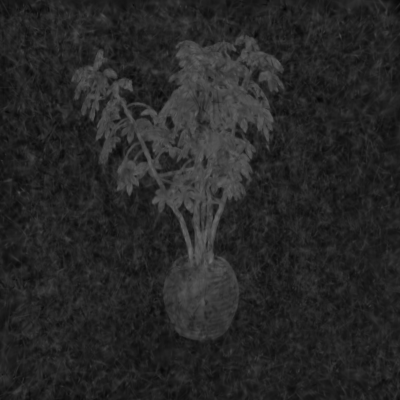} &
\includegraphics[width=0.19\textwidth, height=0.19\textwidth]{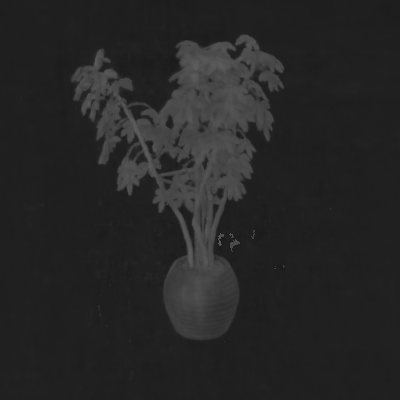} &
\includegraphics[width=0.19\textwidth, height=0.19\textwidth]{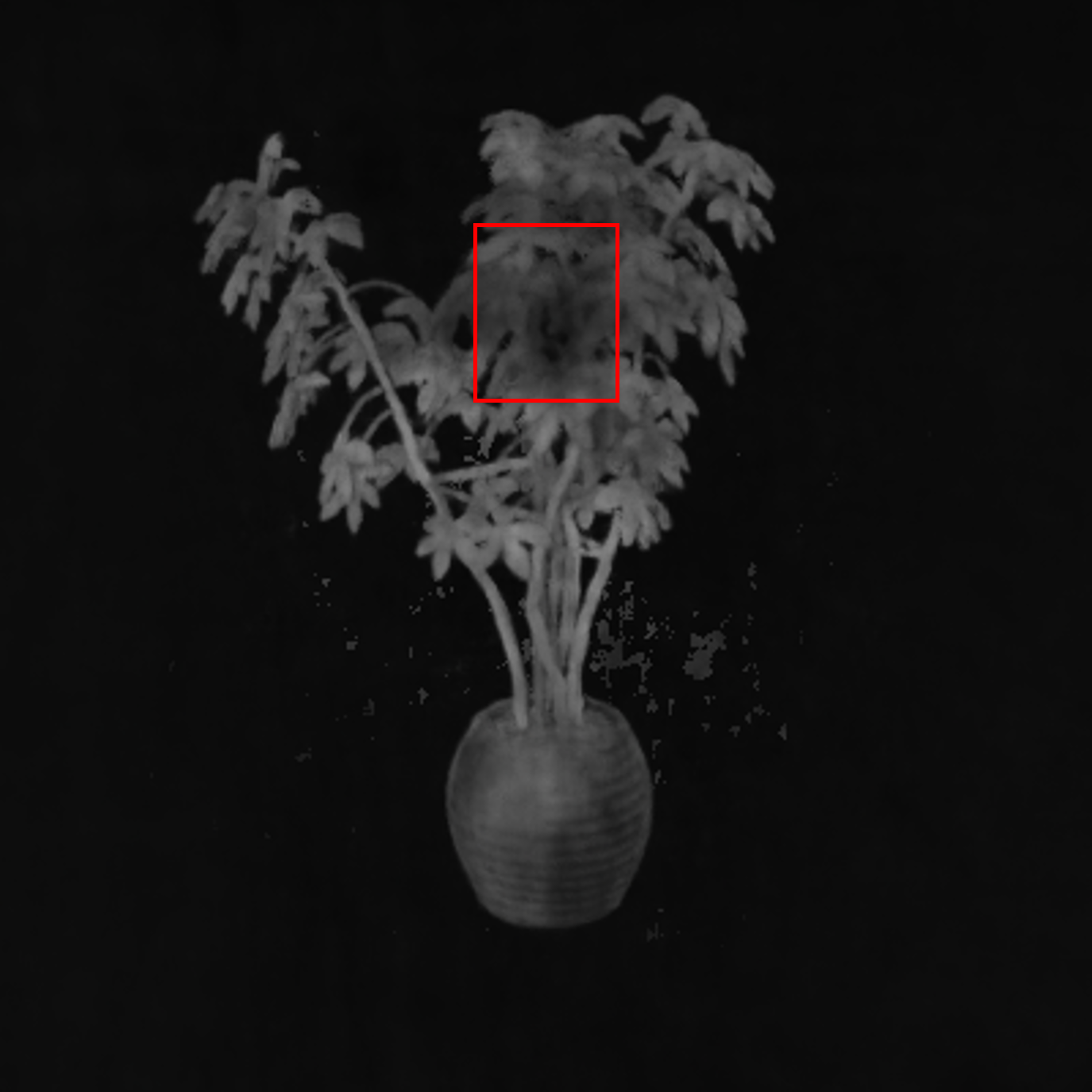}
&
\includegraphics[width=0.19\textwidth, height=0.19\textwidth]{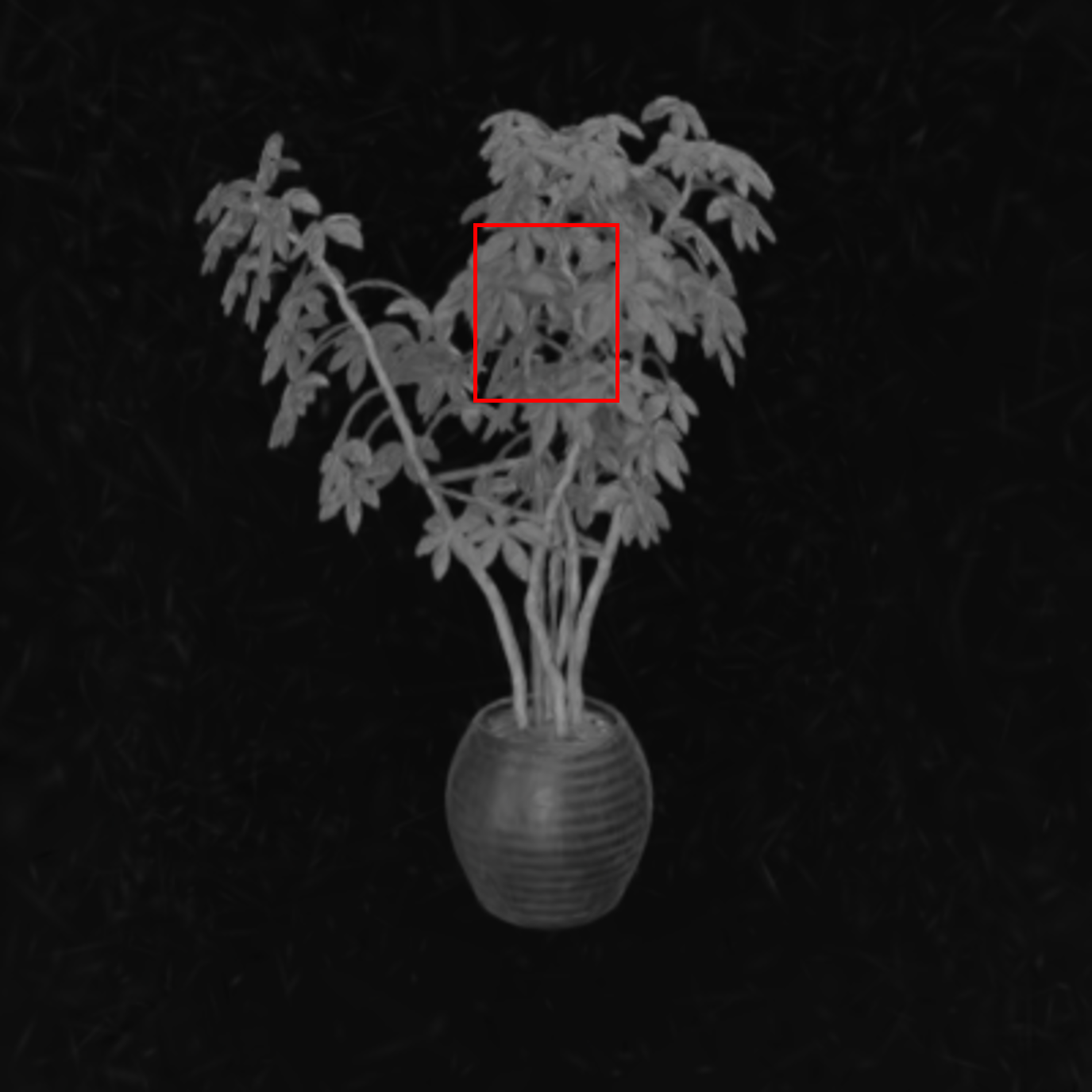}
\\

\includegraphics[width=0.19\textwidth, height=0.19\textwidth]{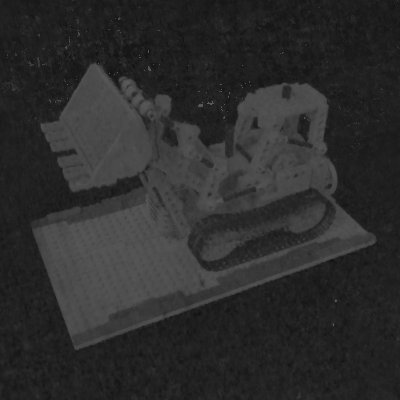} 
&\includegraphics[width=0.19\textwidth, height=0.19\textwidth]{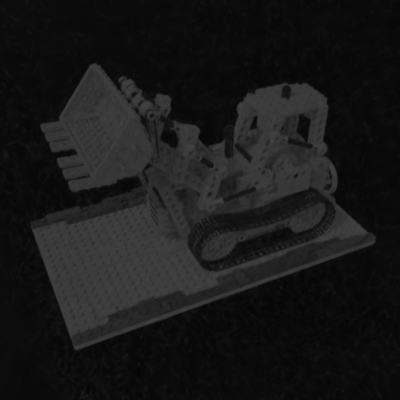} 
&\includegraphics[width=0.19\textwidth, height=0.19\textwidth]{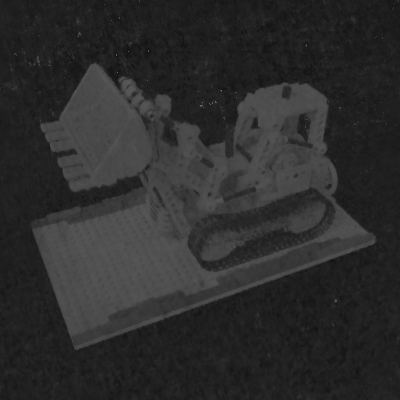} 
&\includegraphics[width=0.19\textwidth, height=0.19\textwidth]{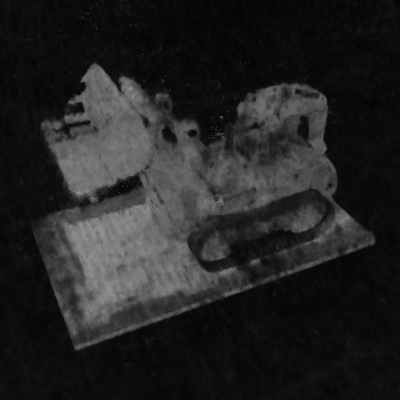}
&\includegraphics[width=0.19\textwidth, height=0.19\textwidth]{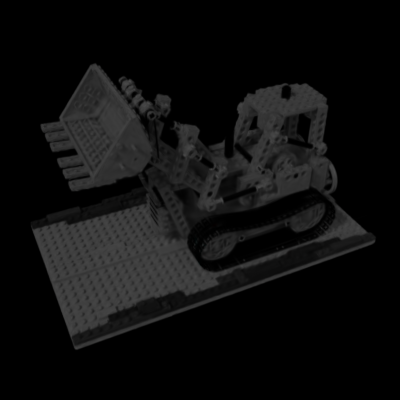}
\\

\includegraphics[width=0.19\textwidth, height=0.19\textwidth]{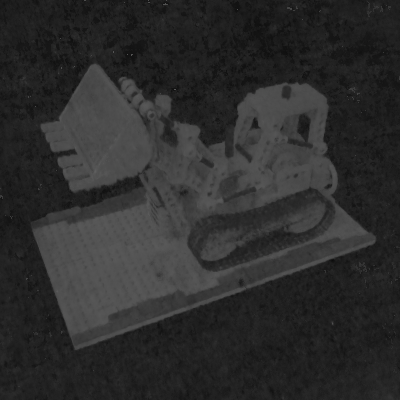} &
\includegraphics[width=0.19\textwidth, height=0.19\textwidth]{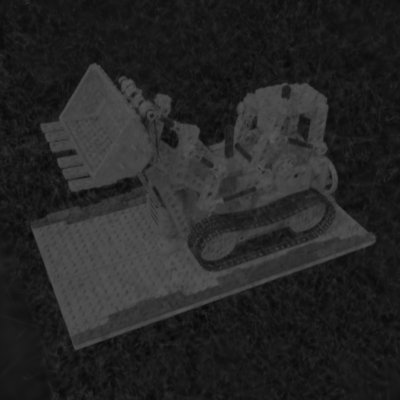} &
\includegraphics[width=0.19\textwidth, height=0.19\textwidth]{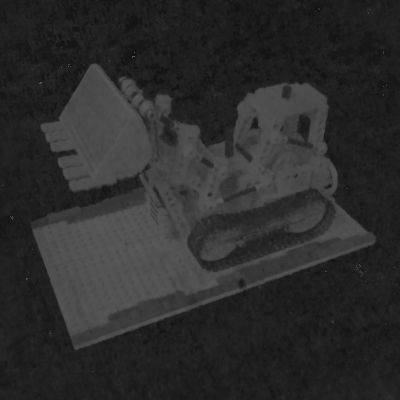} &
\includegraphics[width=0.19\textwidth, height=0.19\textwidth]{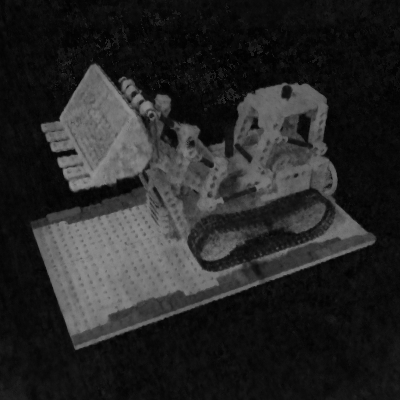}
&
\includegraphics[width=0.19\textwidth, height=0.19\textwidth]{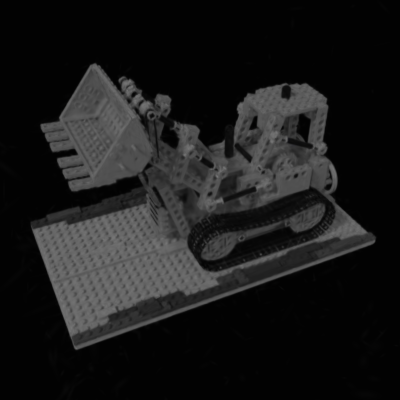}
\\

\includegraphics[width=0.19\textwidth, height=0.19\textwidth]{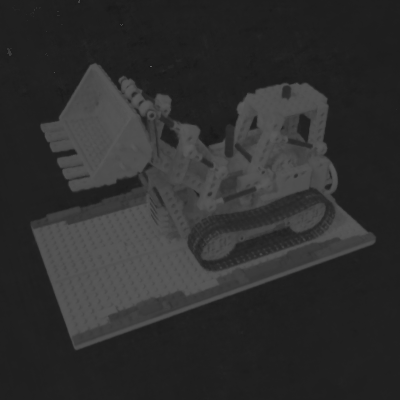} &
\includegraphics[width=0.19\textwidth, height=0.19\textwidth]{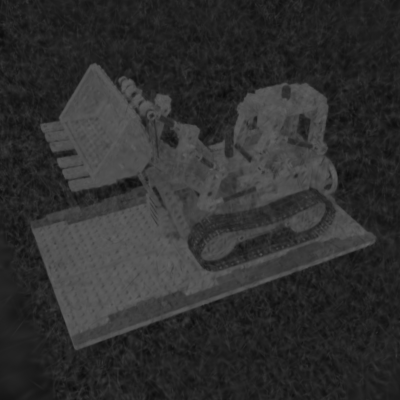} &
\includegraphics[width=0.19\textwidth, height=0.19\textwidth]{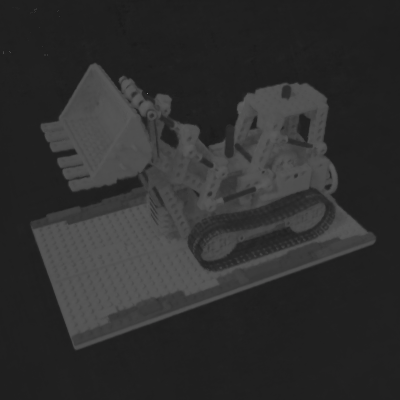} &
\includegraphics[width=0.19\textwidth, height=0.19\textwidth]{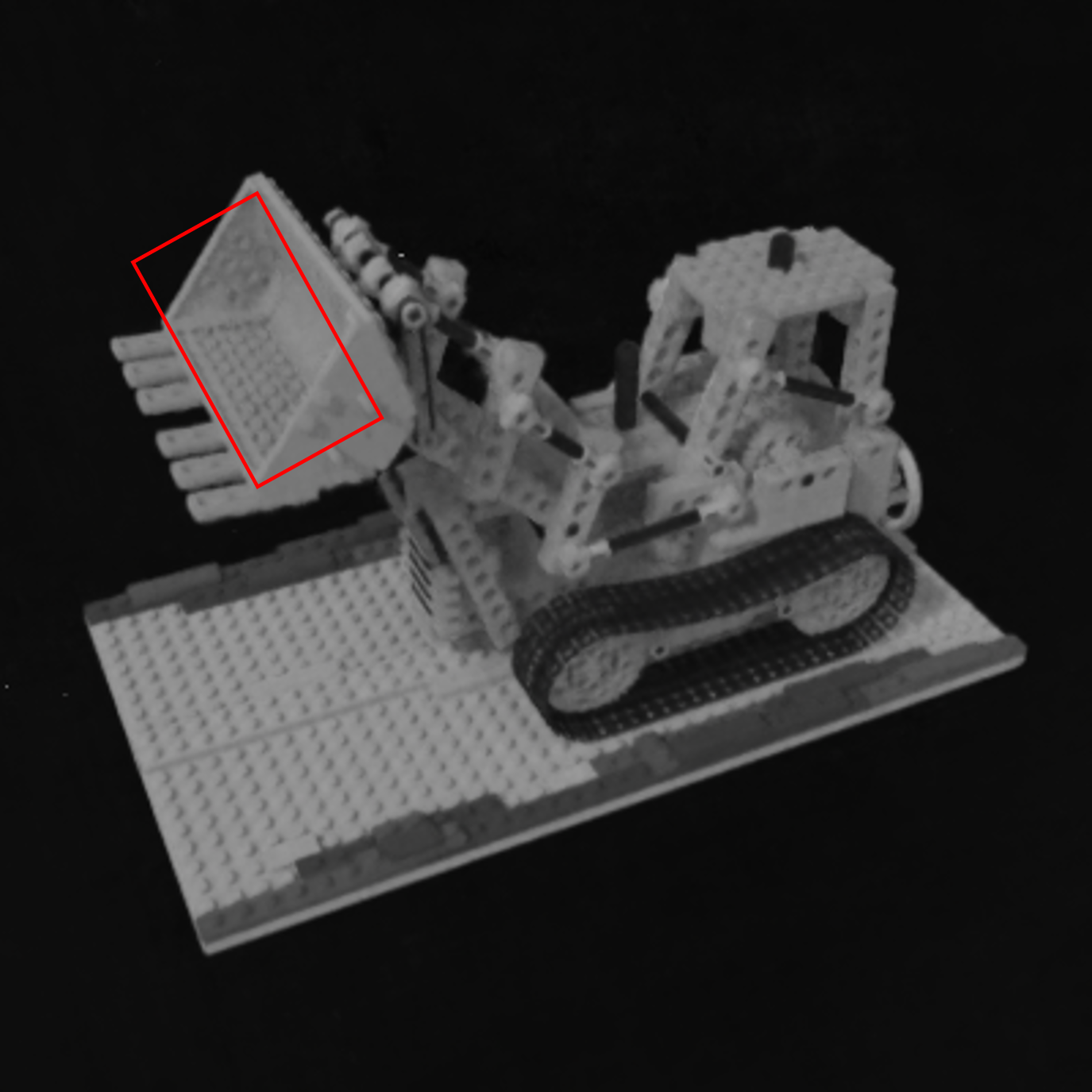}
&
\includegraphics[width=0.19\textwidth, height=0.19\textwidth]{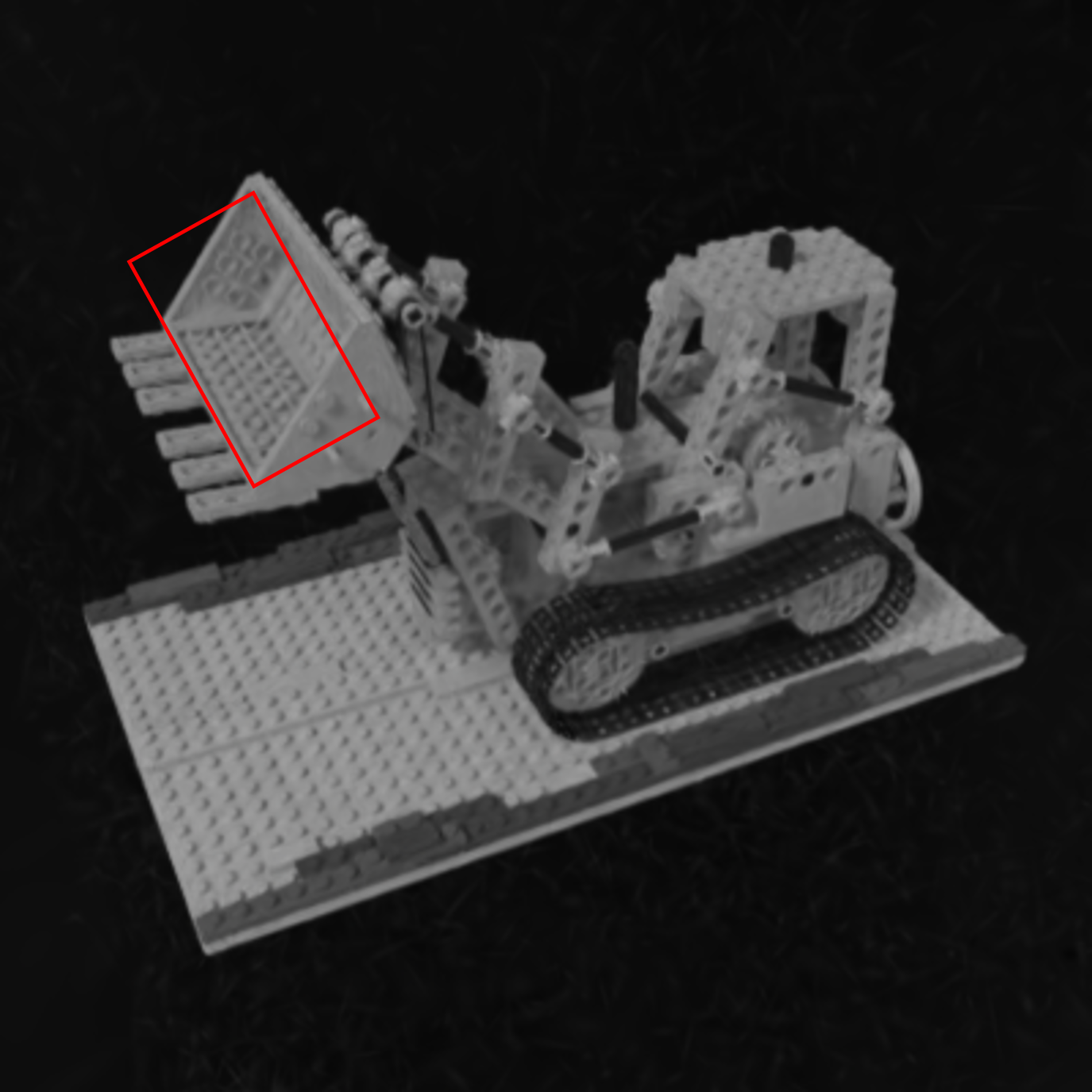}
\\
\tiny {Spk2img+NeRF} & \tiny {Spk2img+GS} & \tiny {Spike-NeRF} & \tiny {SpikeNeRF} & \tiny {Ours}\\
\end{tabular*}
\end{minipage}
\caption{Qualitative results on different light intensities. In the figure, every three rows represent one scene (The names of the two scenes are "ficus" and "lego"), with the first, second, and third rows corresponding to extreme low light intensity, medium low light intensity, and original light intensity, respectively. It is evident from the figure that our model consistently reconstructs the complete scene structure and fine details under all lighting conditions. In contrast, other methods often fail to reconstruct accurately and struggle to recover fine scene details under low lighting conditions, and they also produce significant noise at relatively higher lighting intensities.}
\label{Different Lighting Intensity01}
\end{figure*}

\begin{figure*}[t]
\centering
\begin{minipage}{\textwidth}
\begin{tabular*}{\textwidth}{ccccc}
\includegraphics[width=0.19\textwidth, height=0.19\textwidth]{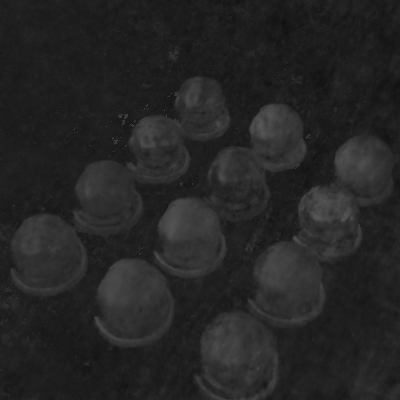} 
&\includegraphics[width=0.19\textwidth, height=0.19\textwidth]{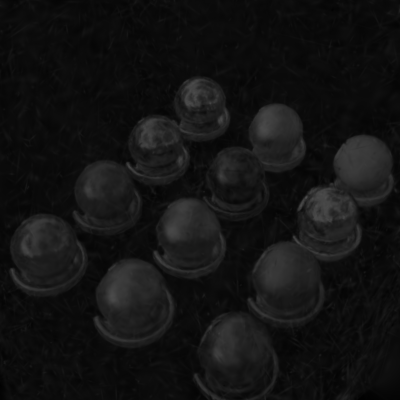} 
&\includegraphics[width=0.19\textwidth, height=0.19\textwidth]{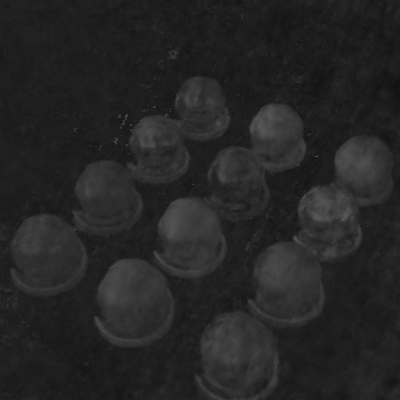} 
&\includegraphics[width=0.19\textwidth, height=0.19\textwidth]{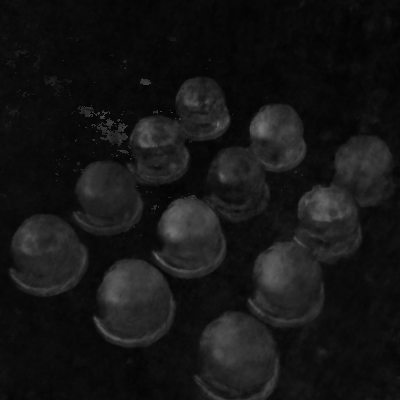}
&\includegraphics[width=0.19\textwidth, height=0.19\textwidth]{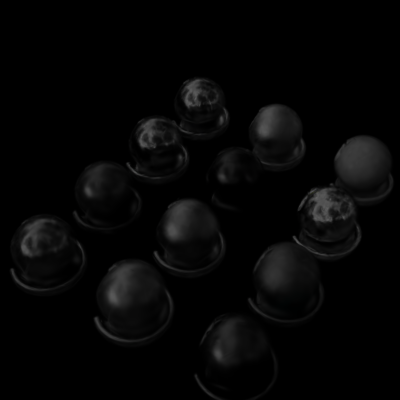}
\\

\includegraphics[width=0.19\textwidth, height=0.19\textwidth]{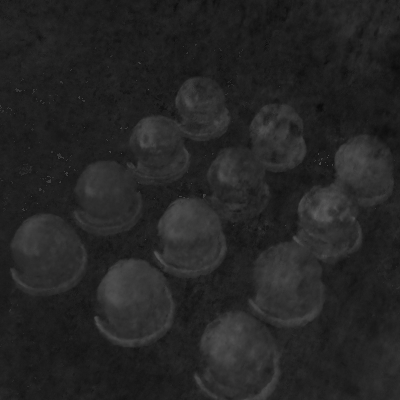} &
\includegraphics[width=0.19\textwidth, height=0.19\textwidth]{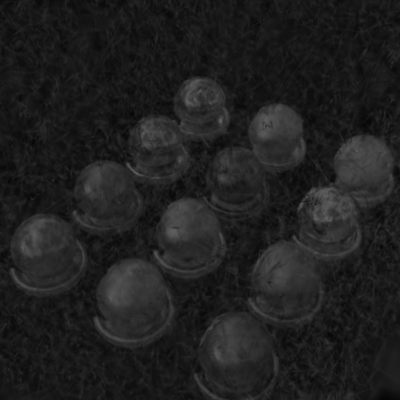} &
\includegraphics[width=0.19\textwidth, height=0.19\textwidth]{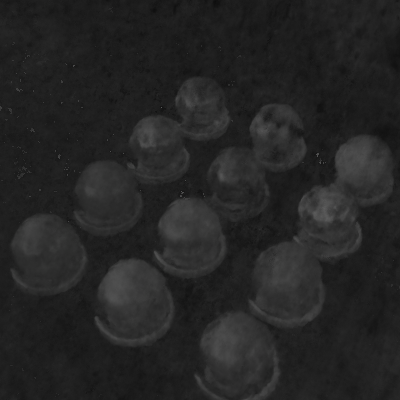} &
\includegraphics[width=0.19\textwidth, height=0.19\textwidth]{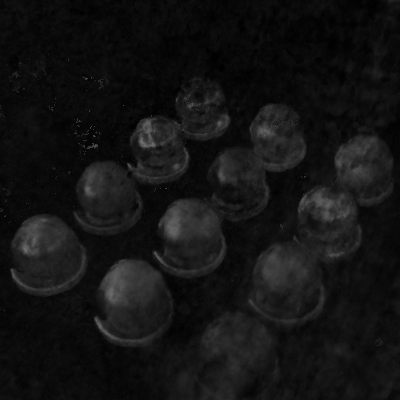}
&
\includegraphics[width=0.19\textwidth, height=0.19\textwidth]{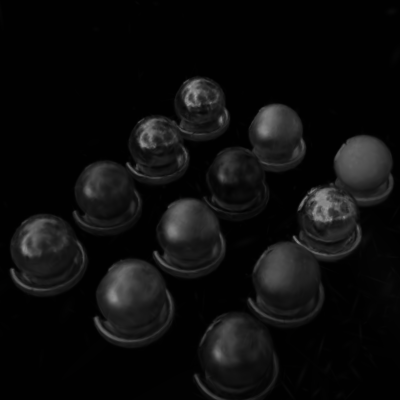}
\\

\includegraphics[width=0.19\textwidth, height=0.19\textwidth]{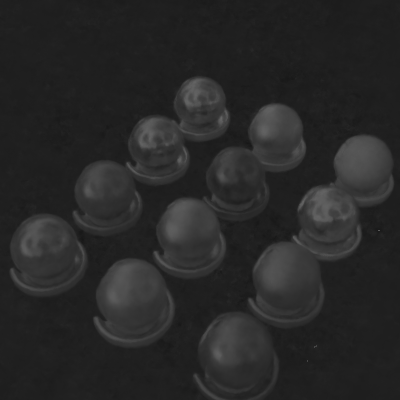} &
\includegraphics[width=0.19\textwidth, height=0.19\textwidth]{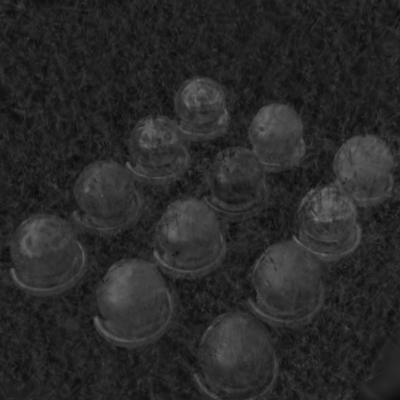} &
\includegraphics[width=0.19\textwidth, height=0.19\textwidth]{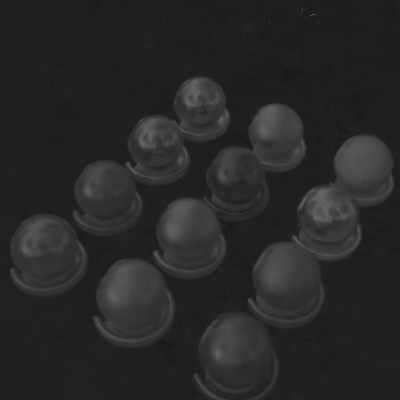} &
\includegraphics[width=0.19\textwidth, height=0.19\textwidth]{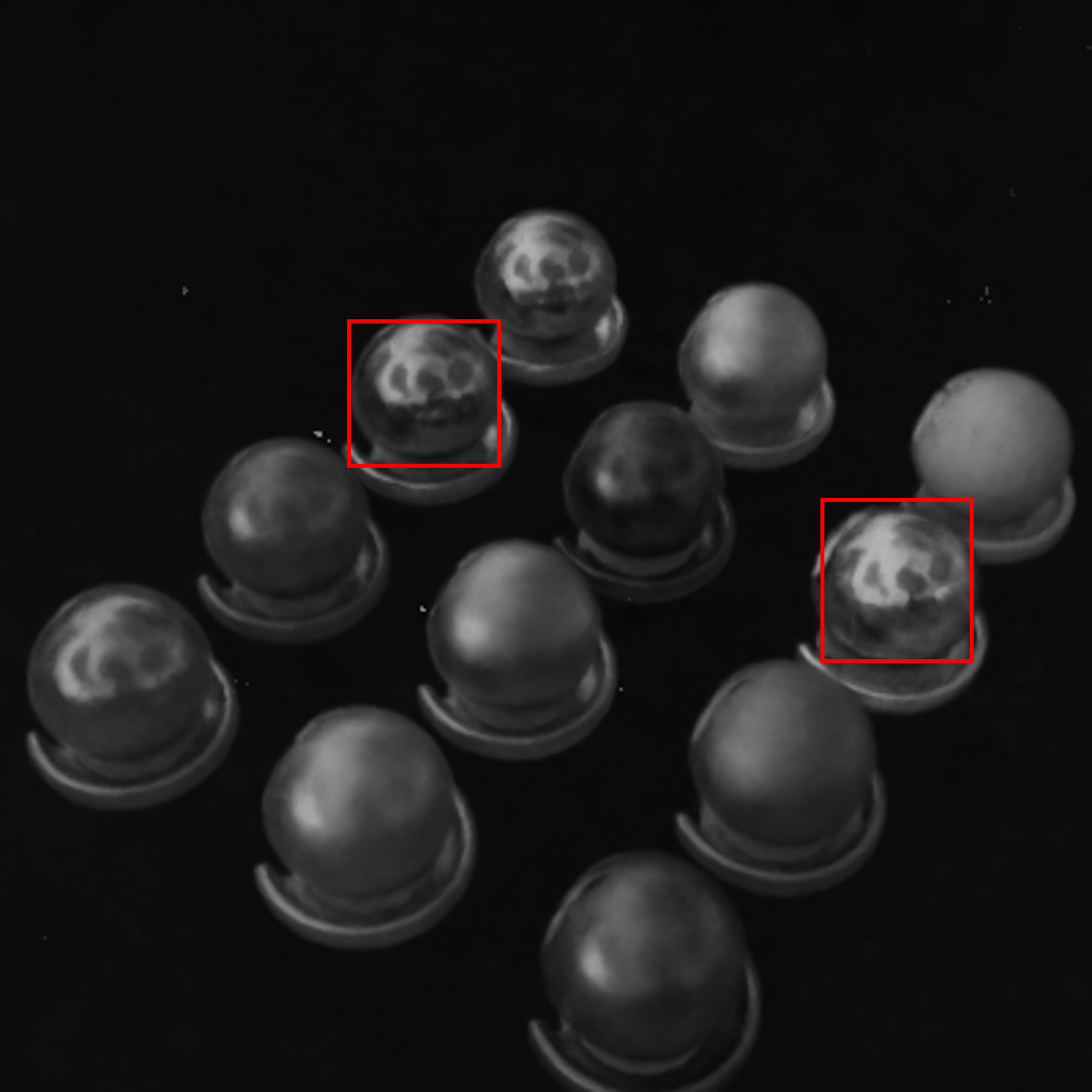}
&
\includegraphics[width=0.19\textwidth, height=0.19\textwidth]{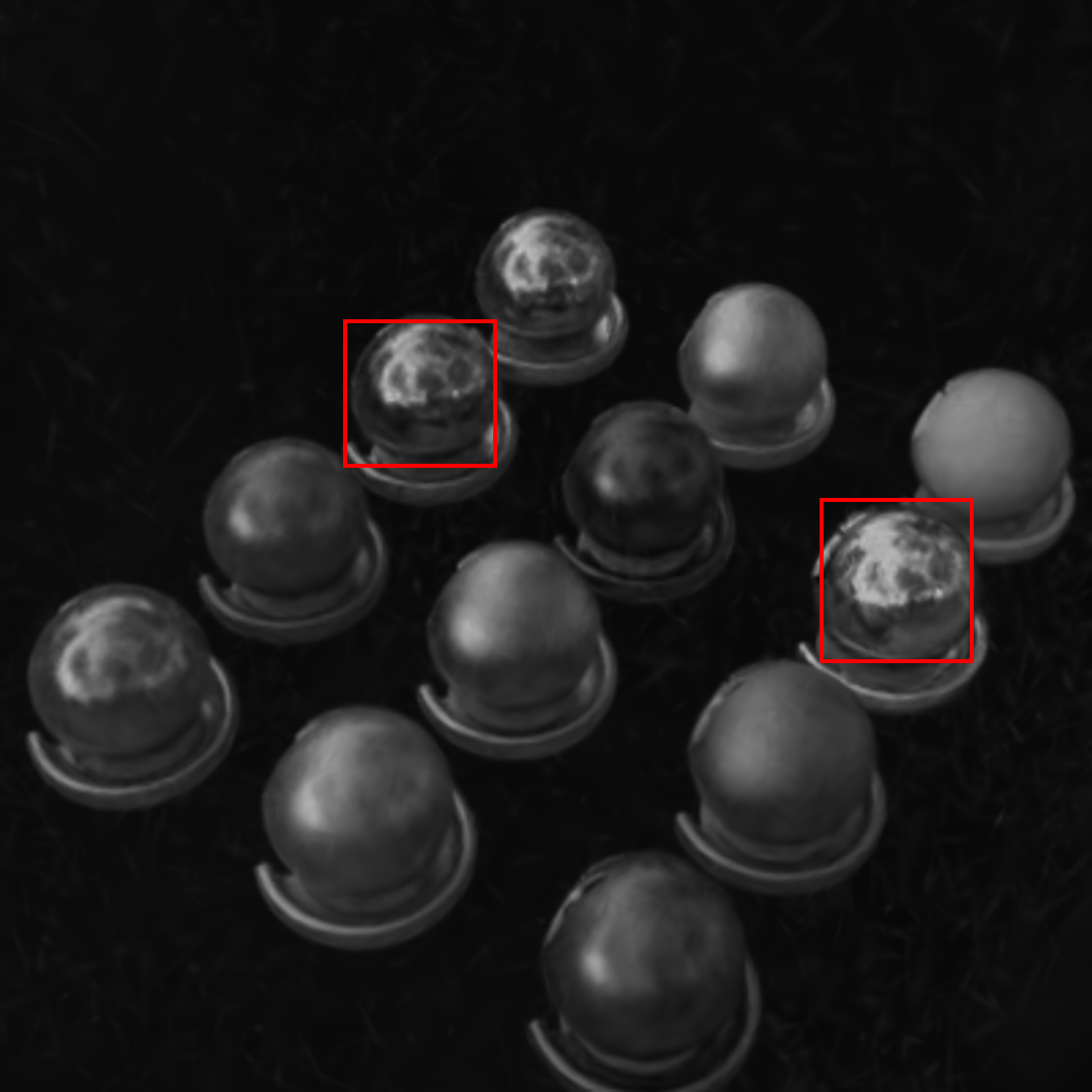}
\\
\tiny {Spk2img+NeRF} & \tiny {Spk2img+GS} & \tiny {Spike-NeRF} & \tiny {SpikeNeRF} & \tiny {Ours}\\
\end{tabular*}
\end{minipage}
\caption{Qualitative results on different light intensities. This figure is a continuation of Fig.~\ref{Different Lighting Intensity01}, with the tested scene named "materials." The light intensity settings are consistent with Fig.~\ref{Different Lighting Intensity01}. It can be observed that our model consistently recovers fine reflective details under different lighting intensities. In contrast, other methods struggle to recover fine details under lower lighting intensities and produce significant noise under relatively higher lighting intensities.}

\label{Different Lighting Intensity02}
\end{figure*}

We conducted experiments on three synthetic datasets (ficus, lego and materials) with varying lighting intensities to further demonstrate the generalization of our method under different illumination conditions (We set our model N to 256 for the experiment). For each scene, we performed experiments under three lighting intensity conditions: extremely low lighting, moderately low lighting, and original lighting intensity. The qualitative results are shown in Figure.~\ref{Different Lighting Intensity01} and Figure.~\ref{Different Lighting Intensity02}. As can be observed, our method outperforms existing methods under all lighting conditions. Even in extremely low-light scenarios, our method is able to reconstruct complete scene structures and render fine texture details. In contrast, other methods either fail to reconstruct the complete scene structure or produce very blurred texture details, and even generate significant noise at relatively higher lighting intensities.

The quantitative comparison results are shown in Table~\ref{different table}. For each lighting condition in each scene, we calculated three metrics: PSNR, SSIM, and LPIPS (using the ground truth RGB images as reference, consistent with the full paper). As shown, our method outperforms existing methods on nearly all metrics.


\section{Complete visualization results of the ablation experiments}
\begin{figure*}[!t]
\centering
\begin{minipage}{\textwidth}
\begin{tabular*}{\textwidth}{cccc}
\includegraphics[width=0.24\textwidth, height=0.24\textwidth]{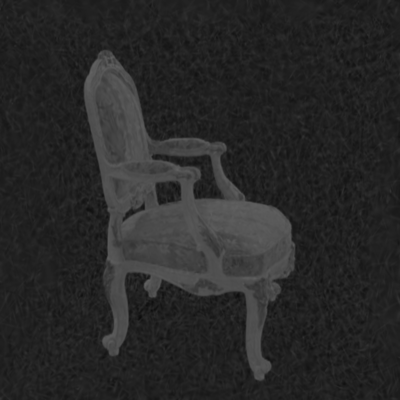} 
&\includegraphics[width=0.24\textwidth, height=0.24\textwidth]{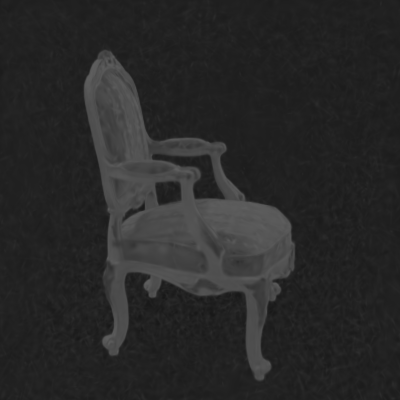} 
&\includegraphics[width=0.24\textwidth, height=0.24\textwidth]{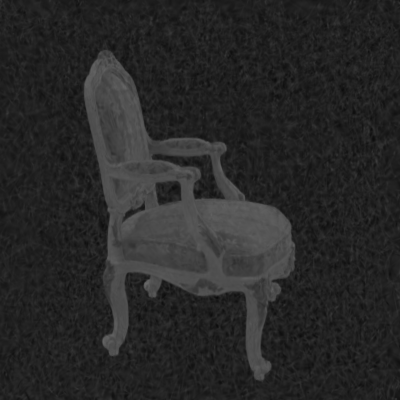} 
&\includegraphics[width=0.24\textwidth, height=0.24\textwidth]{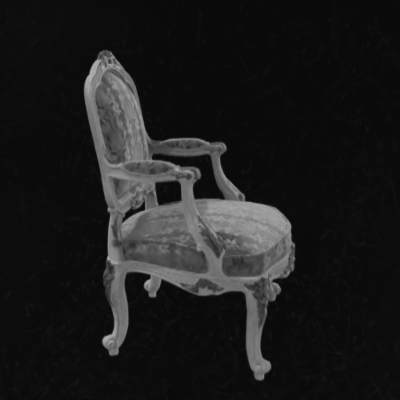}\\

\includegraphics[width=0.24\textwidth, height=0.24\textwidth]{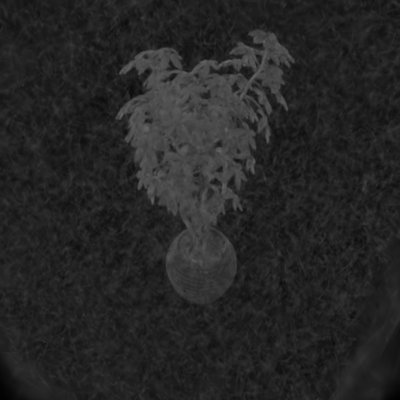} &
\includegraphics[width=0.24\textwidth, height=0.24\textwidth]{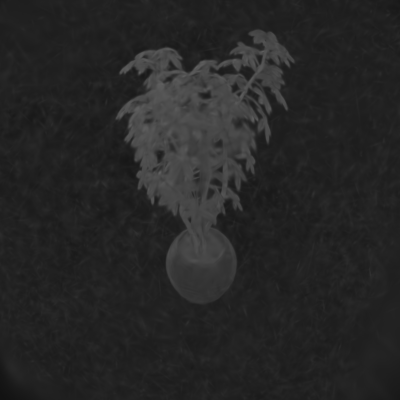} &
\includegraphics[width=0.24\textwidth, height=0.24\textwidth]{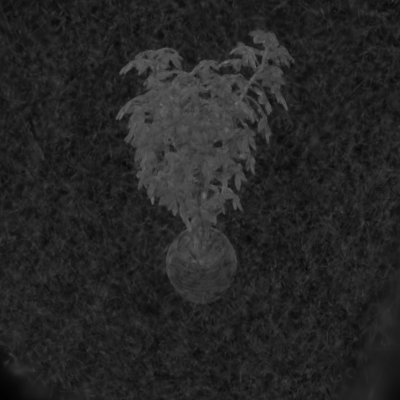} &
\includegraphics[width=0.24\textwidth, height=0.24\textwidth]{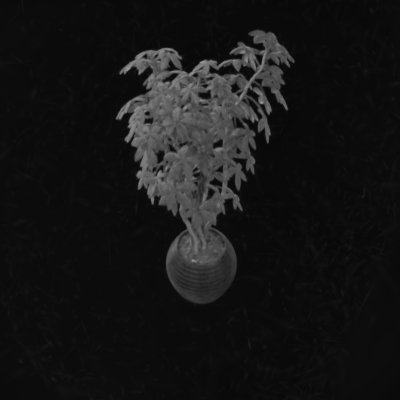}\\

\includegraphics[width=0.24\textwidth, height=0.24\textwidth]{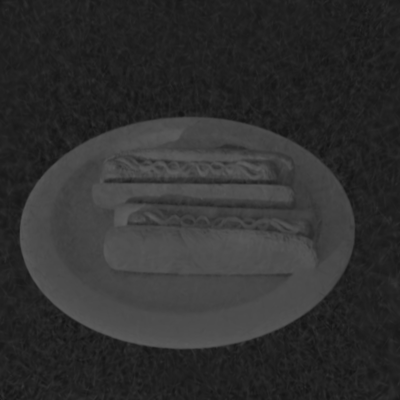} &
\includegraphics[width=0.24\textwidth, height=0.24\textwidth]{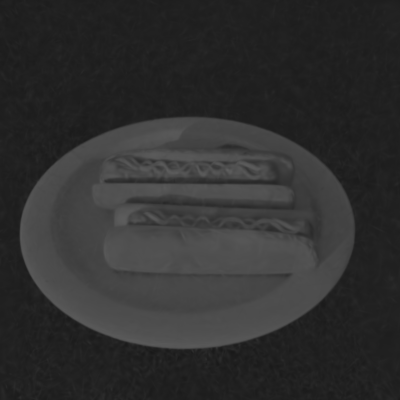} &
\includegraphics[width=0.24\textwidth, height=0.24\textwidth]{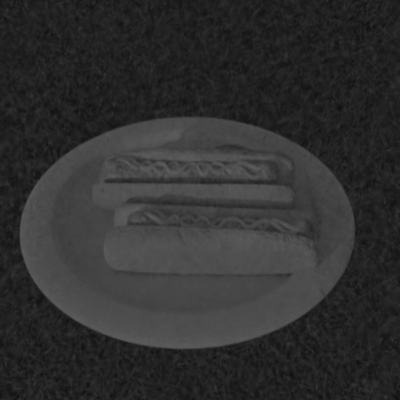} &
\includegraphics[width=0.24\textwidth, height=0.24\textwidth]{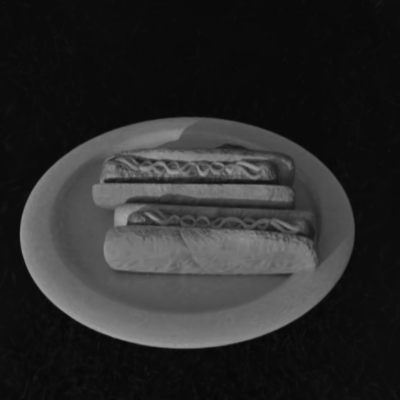}\\

\includegraphics[width=0.24\textwidth, height=0.24\textwidth]{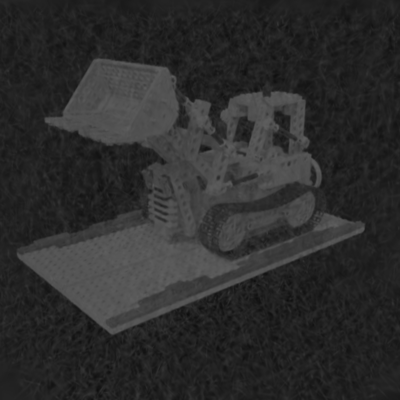} &
\includegraphics[width=0.24\textwidth, height=0.24\textwidth]{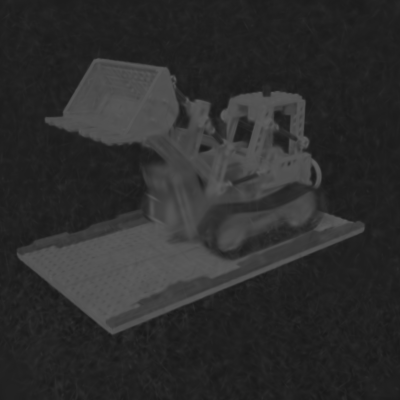} &
\includegraphics[width=0.24\textwidth, height=0.24\textwidth]{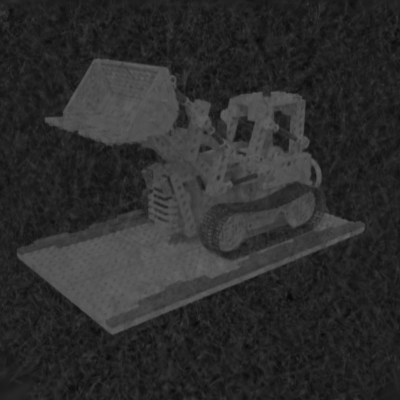} &
\includegraphics[width=0.24\textwidth, height=0.24\textwidth]{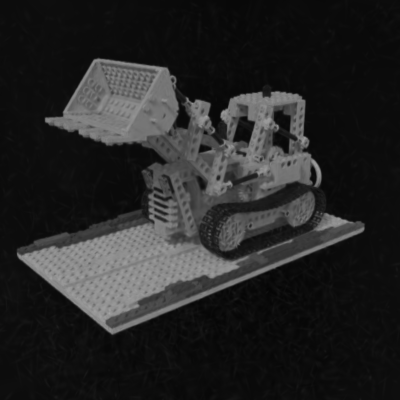}\\

\includegraphics[width=0.24\textwidth, height=0.24\textwidth]{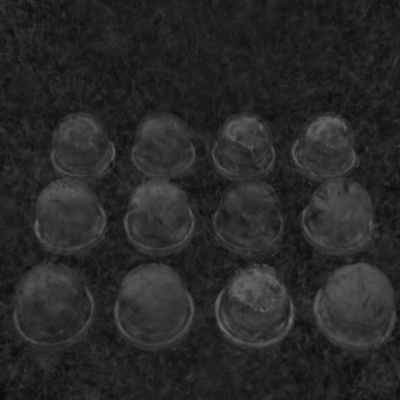} &
\includegraphics[width=0.24\textwidth, height=0.24\textwidth]{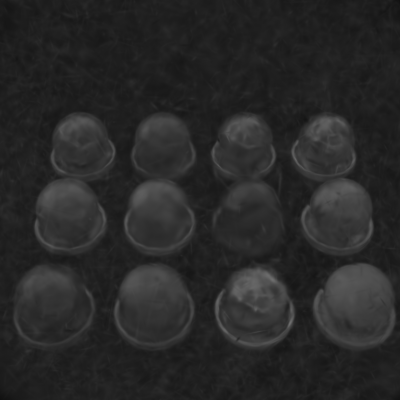} &
\includegraphics[width=0.24\textwidth, height=0.24\textwidth]{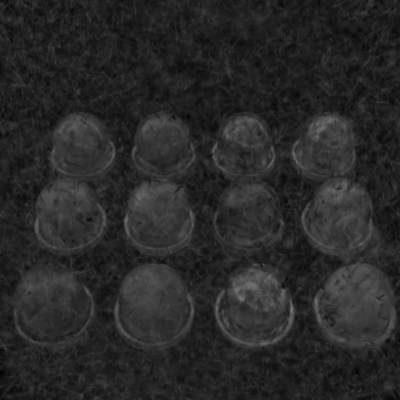} &
\includegraphics[width=0.24\textwidth, height=0.24\textwidth]{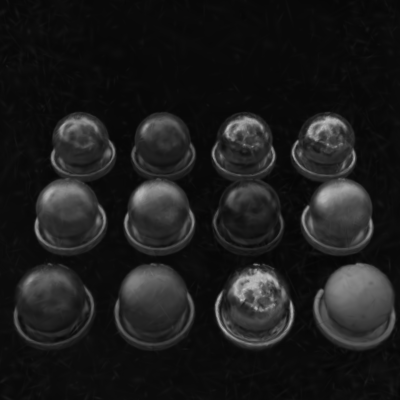}\\

\includegraphics[width=0.24\textwidth, height=0.24\textwidth]{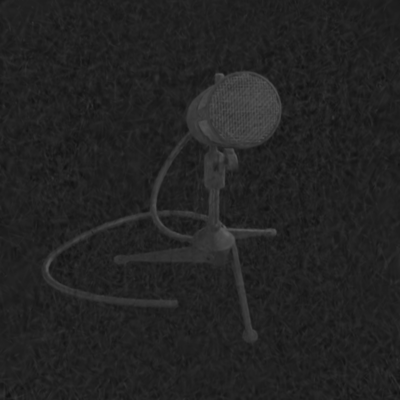} &
\includegraphics[width=0.24\textwidth, height=0.24\textwidth]{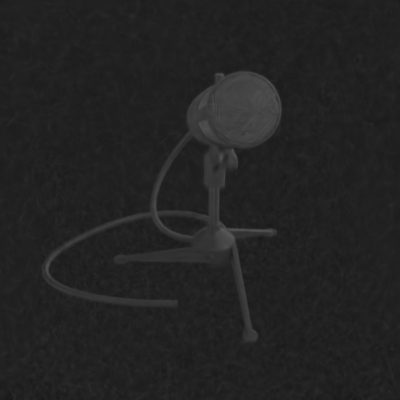} &
\includegraphics[width=0.24\textwidth, height=0.24\textwidth]{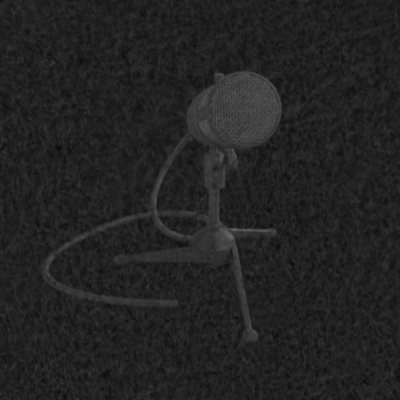} &
\includegraphics[width=0.24\textwidth, height=0.24\textwidth]{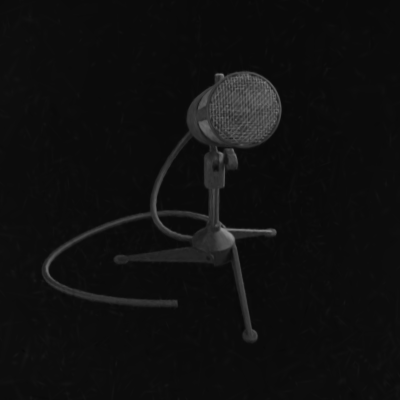} \\

\tiny {$L_{I_{in}}$} & \tiny {$L_{1}$} & \tiny {$D_{noise}$} & \tiny {$Full(L_{s})$}\\
\end{tabular*}
\end{minipage}
\caption{Qualitative comparison of ablation experiments on the synthetic dataset. As shown in the figure, images rendered with $L_{I_{in}}$ and $D_{noise}$ contain noticeable noise, while images rendered with $L_{1}$ are overly smooth and lack detail.}
\label{Ablation Study Synthetic}
\end{figure*}

\begin{figure*}[t]
\vspace{-28em}
\begin{minipage}{0.9\textwidth}
\begin{tabular*}{0.9\textwidth}{cccc}
\includegraphics[width=0.275\textwidth]{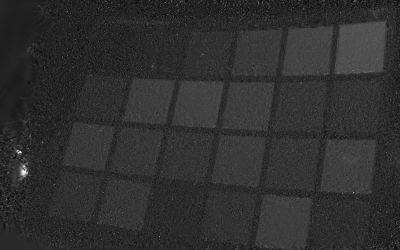} 
&\includegraphics[width=0.275\textwidth]{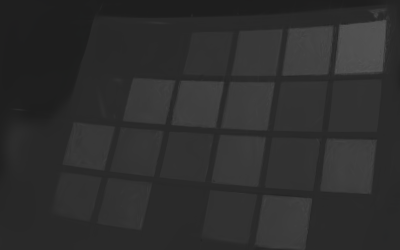} 
&\includegraphics[width=0.275\textwidth]{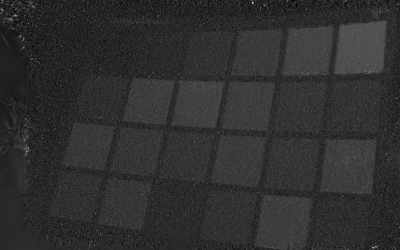} 
&\includegraphics[width=0.275\textwidth]{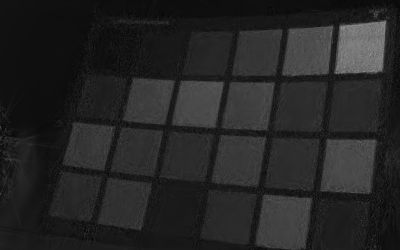}\\

\includegraphics[width=0.275\textwidth]{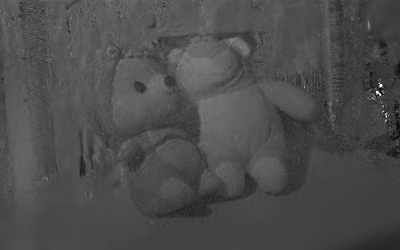} &
\includegraphics[width=0.275\textwidth]{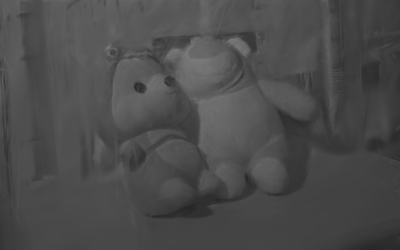} &
\includegraphics[width=0.275\textwidth]{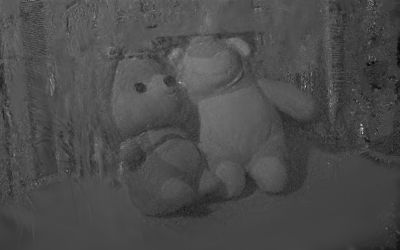} &
\includegraphics[width=0.275\textwidth]{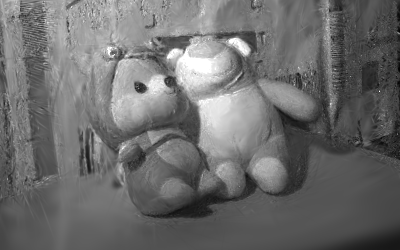}\\

\includegraphics[width=0.275\textwidth]{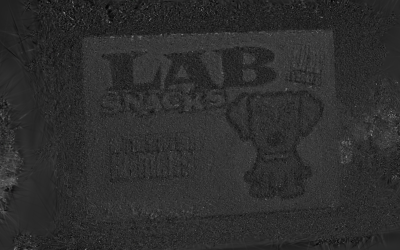} &
\includegraphics[width=0.275\textwidth]{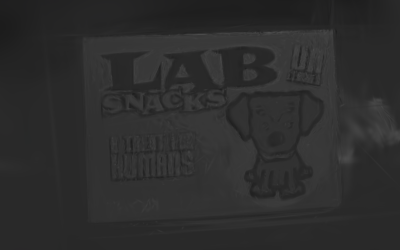} &
\includegraphics[width=0.275\textwidth]{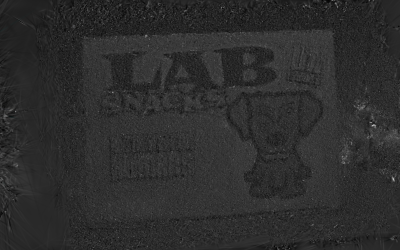} &
\includegraphics[width=0.275\textwidth]{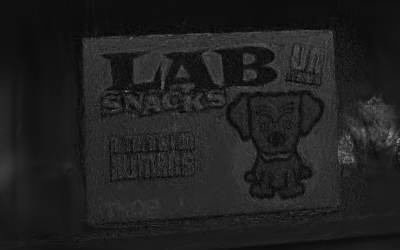}\\

\includegraphics[width=0.275\textwidth]{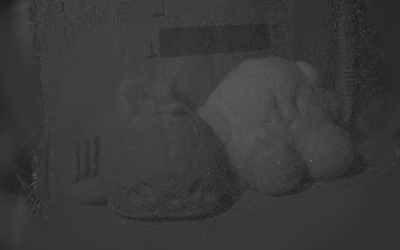} &
\includegraphics[width=0.275\textwidth]{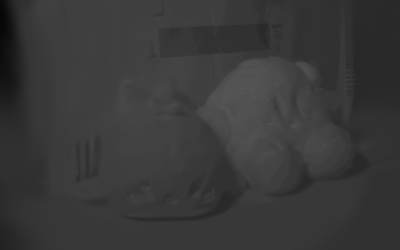} &
\includegraphics[width=0.275\textwidth]{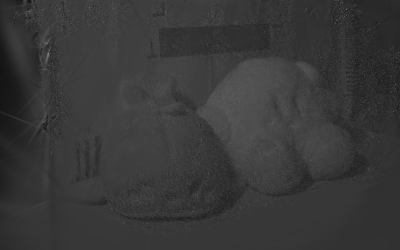} &
\includegraphics[width=0.275\textwidth]{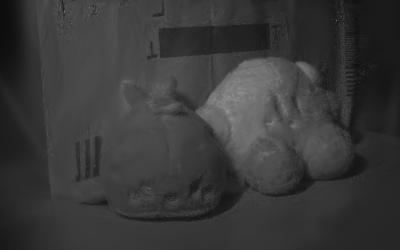}\\

\tiny {$L_{I_{in}}$} & \tiny {$L_{1}$} & \tiny {$D_{noise}$} & \tiny {$Full(L_{s})$}\\
\end{tabular*}
\end{minipage}
\caption{Qualitative comparison of ablation experiments on the real dataset. As shown in the figure, images rendered with $L_{I_{in}}$ and $D_{noise}$ contain noticeable noise, while images rendered with $L_{1}$ are overly smooth and lack detail.}
\label{Ablation Study Real}
\end{figure*}

According to the experimental setup described in the paper, we have provided the complete visual results for both the synthetic and real datasets in the supplementary materials. We conducted ablation experiments on 6 scenes from the synthetic dataset and 4 scenes from the real dataset. As shown in Fig.~\ref{Ablation Study Synthetic} and Fig.~\ref{Ablation Study Real}, when only using $L_{1}$ as the loss function, the texture details of the scenes appear overly smooth. Additionally, when the noise embedding pipeline is removed or when using estimated light intensity directly (bypassing the spike generation pipeline) as input, the synthesized views exhibit a significant amount of noise. In contrast, the complete framework demonstrates high robustness to noise and is capable of recovering fine texture details(The quantitative comparison of ablation experiments is presented in Table 3 of the full paper. here, we only showcase the complete visual results(qualitative comparison)).


\begin{table}[!ht]
\footnotesize
\centering
\caption{Different Lighting Intensity Experiments. The terms "Light intensity (Low)," "Light intensity (Med)," and "Light intensity (Orig)" in the table correspond to extremely low lighting, moderately low lighting, and original lighting intensity, respectively. We calculated the average metrics for the three scenes (ficus, lego, and materials) under each lighting condition, and the results are shown below. Each color shading indicates the \colorbox{red!25}{best} and \colorbox{orange!25}{second-best} result.}
\resizebox{\linewidth}{!}{
\vspace{-4mm}
\begin{tabular}{ccccccccccccc|c}
\toprule[1pt]
\multirow{2}{*}{Method}&\multicolumn{3}{c}{Light intensity(Low)} &\multicolumn{3}{c}{Light intensity(Med)} &\multicolumn{3}{c}{Light intensity(Orig)} &\multirow{2}{*}{Time$\downarrow$}\\
\cmidrule(r){2-4} \cmidrule(r){5-7} \cmidrule(r){8-10}

& \multicolumn{1}{c}{PSNR$\uparrow$} & \multicolumn{1}{c}{SSIM $\uparrow$} & \multicolumn{1}{c}{LPIPS $\downarrow$} & \multicolumn{1}{c}{PSNR $\uparrow$} & \multicolumn{1}{c}{SSIM $\uparrow$} & \multicolumn{1}{c}{LPIPS $\downarrow$}& \multicolumn{1}{c}{PSNR $\uparrow$} & \multicolumn{1}{c}{SSIM $\uparrow$} & \multicolumn{1}{c}{LPIPS $\downarrow$}\\
\hline
Spk2img+NeRF(200K)&\multicolumn{1}{c}{15.73}& \multicolumn{1}{c}{.0786}& \multicolumn{1}{c}{.4162}& \multicolumn{1}{c}{15.83}& \multicolumn{1}{c}{.0788}& \multicolumn{1}{c}{.4582}& \multicolumn{1}{c}{16.92}& \multicolumn{1}{c}{.1020}& \multicolumn{1}{c}{.2445} & >3 hours \\
Spk2img+GS(30K)& \cellcolor{orange!25}17.72 & \cellcolor{orange!25}.1342 & \cellcolor{orange!25}.2921 & 16.93&.1038 & .4920& 15.97& .0989& .5308 & $\sim$5 mins \\ 
Spike-NeRF(200K)& 15.76& .0791& .4011& 15.83& .0792& .4433& 16.28& .1019& .2422 & >3 hours\\
SpikeNeRF(200K)& 17.33& .1126& .3578& \cellcolor{orange!25}18.11& \cellcolor{orange!25}.1352& \cellcolor{orange!25}.3782& \cellcolor{orange!25}20.7& \cellcolor{orange!25}.1882& \cellcolor{red!25}.1677 & >10 hours\\ \hline
Ours(30K)& \cellcolor{red!25}18.25& \cellcolor{red!25}.8021& \cellcolor{red!25}.1977& \cellcolor{red!25}19.89& \cellcolor{red!25}.7035& \cellcolor{red!25}.1574& \cellcolor{red!25}21.0& \cellcolor{red!25}.2010& \cellcolor{orange!25}.1875 & $\sim$40 mins\\ \toprule[1pt]
\end{tabular}
}
\label{different table}

\end{table}

\newpage
